\documentclass[authoryear,10pt,oneside,onecolumn]{elsarticle}

\usepackage{algorithm}
\usepackage[noend]{algpseudocode}
\usepackage{amsfonts}

\usepackage{amsmath}

\usepackage{amssymb}
\usepackage{amsthm}
\usepackage{capt-of}

\usepackage{commath}
\usepackage[noadjust]{cite}
\usepackage{float}
\usepackage{graphicx}
\usepackage[utf8]{inputenc}

\usepackage{hyperref}

\usepackage{marginnote}

\usepackage{mathtools}
\usepackage{multirow}
\usepackage{prletters}

\usepackage{siunitx}
 \usepackage{setspace}
\usepackage{soul}

\usepackage{subcaption}

\usepackage[colorinlistoftodos,prependcaption]{todonotes}
\usepackage[normalem]{ulem}

\usepackage{url}

\usepackage{verbatim}
\usepackage{xparse}
\usepackage{xcolor}

\newcommand{\mbf}{{\mathbf{f}}}

\newcommand{\mbI}{{\mathbf{I}}}

\newcommand{\mbK}{{\mathbf{K}}}

\newcommand{\mbp}{{\mathbf{p}}}

\newcommand{\mbq}{{\mathbf{q}}}

\newcommand{\mbx}{{\mathbf{x}}}
\newcommand{\mbX}{{\mathbf{X}}}

\newcommand{\mby}{{\mathbf{y}}}

\newcommand{\mbz}{{\mathbf{z}}}

\newcommand{\mbomega}{{\boldsymbol{\omega}}}

\newcommand{\mbphi}{{\boldsymbol{\phi}}}

\definecolor{red}{rgb}{0.00,0.00,0.00}
\newcommand{\cred}[1] {\textcolor{red}{#1}}

\date{}

\begin{document}

  \setcounter{page}{1}

\begin{frontmatter}

\title{
Learning Inter-Annual Flood Loss Risk Models From Historical Flood Insurance Claims and Extreme Rainfall Data}


\author[1,2]{Joaquin \snm{Salas}\corref{cor1}}
\ead{jsalasr@ipn.mx}

\author[2]{Anamitra  \snm{Saha} }
\author[2]{Sai   \snm{Ravela} }

\cortext[cor1]{Corresponding author:
  Joaquín Salas. CICATA Querétaro. Instituto Politécnico Nacional. Cerro Blanco 141, Colinas del Cimatario. Querétaro. México. 76090. Tel.: +52-55-5729-6000 $\times$ 81015}



\address[1]{CICATA Querétaro. Instituto Politécnico Nacional.
\\Cerro Blanco 141, Colinas del Cimatario, Querétaro, Querétaro. México. 76090.}
\address[2]{Earth Signals and Systems Group, Earth, Atmospheric and Planetary Sciences, Massachusetts Institute of Technology,
\\77 Massachusetts Avenue,
Cambridge, MA 02139-4307}



\begin{abstract}

Flooding is one of the most disastrous natural hazards responsible for substantial economic losses. A predictive model for flood-induced financial damages is helpful for many applications, such as climate change adaptation planning and insurance underwriting. This research assesses the predictive capability of regressors constructed on the National Flood Insurance Program (NFIP) dataset
using neural networks (Conditional Generative Adversarial Networks), decision trees (Extreme Gradient Boosting), and kernel-based regressors (Gaussian Process).  The assessment highlights the most informative predictors for regression. A Burr distribution with a bias correction scheme to increase the regressor's predictive capability effectively enables inference for claim amount distributions. Aiming to study the impact of physical variables,  we incorporate Daymet rainfall estimation to NFIP as an additional predictor. A study of the coastal counties in the eight US South-West states resulted in an $R^2=0.807$. Further analysis of 11 counties with a significant number of claims in the NFIP dataset reveals that Extreme Gradient Boosting provides the best results, that bias correction significantly improves the similarity with the reference distribution, and that adding the rainfall predictor strengthens the regressor performance.
\end{abstract}

\end{frontmatter}
\begin{singlespace}



\section{Introduction}
Between 1998 and 2017, climate-related disasters caused economic losses of  USD 2,245 trillion, affecting an estimated $4.4$ billion people~\citep{CRED2018}. While 43.4\% of the disaster events corresponded to flooding, making them the most significant natural hazard to people's lives and properties~\citep{hallegatte2016unbreakable}, the expectation is that the cost associated with them will grow. In its latest assessment, the United Nations (UN) Intergovernmental Panel on Climate Change (IPCC)~\citeyearpar{IPCC2021climate} concludes that the responsibility for climate change resides in human activity. Furthermore, it notices that nothing is arresting the planet from warming 1.5 $^\circ$C above pre-industrial levels in the next two decades, not even stopping greenhouse gas emissions right now. As a result, we anticipate more significant weather events, including extreme droughts,  heatwaves, and floods~\citep{frame2020climate}. Therein lies the need to improve flood loss risk modeling to enable rapid, up-to-date, and reliable projections for adaptation and mitigation strategies, reduce human vulnerability, and advance sustainable decision-making.

Even at a shorter time scale, for example, underwriting insurance at interannual timescales, it has been argued that it is essential to account for climate change~\footnote{Kerry Emanuel, personal communication} that has already occurred. This paper's approach to the problem involves learning an adaptive flood loss model from historical claims and meteorological data. As time evolves, so does the model; thus, a static model does not forecast the distant future. Results indicate that the dynamic data-driven modeling strategy is quite effective at short interannual horizons. At longer time horizons, {\it e.g.}, decadal time scales, using numerical climate model projections is arguably essential for modeling flood loss risk.

In September 1965, Hurricane Betsy hit the US on the coast of Louisiana after traveling around the Florida peninsula, where it caused catastrophic damage in the Florida Keys~\citep{perkins1968hurricane}. The \ USD 8.5 billion (in the year 2000 dollars)  in estimated costs~\citep{emanuel2005divine} prompted the US Congress to create the National Flood Insurance Program (NFIP) in 1968~\citep{elliott2021underwater}. The program, run by the US government, permits homeowners to buy flood insurance at reduced premiums. The insurance is required for all the buildings receiving a loan and located in a Special Flood Hazard Area (SFHA), an area with a one percent chance of flooding in a given year\cred{~\citep{lea2022appeal}}. Recently, NFIP released a censored dataset containing  2,546,311 claims and 4,034,086 policies~\citep{fema_nfips}, providing a rich source of valuable information. The claims may be geospatially aggregated by latitude and longitude up to 0.1$^{\cred{\circ}}$ or politically by county. The dataset offers a glimpse into a vast market to learn about the financial impacts of flooding events. In particular, this research studies the capacity of the dataset to support the construction of flood loss risk models based on machine learning (ML)  to predict the amount paid on flooding claims using the NFIP dataset.

\begin{figure}[t]
    \centering
    \includegraphics[width=6in]{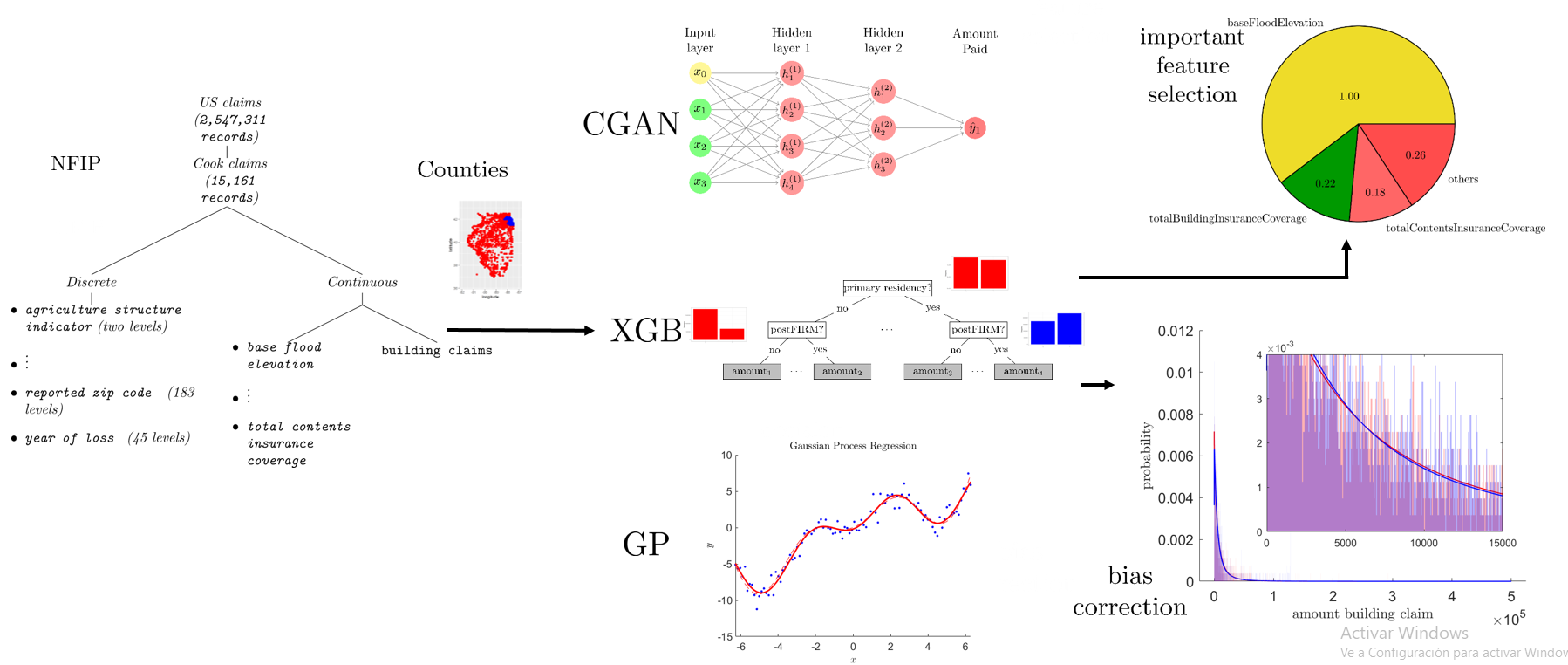}
    \caption[Amount Paid on Building Claims]{Predicting flood loss. Using the NFIP dataset, we construct and compare regressors based on decision trees, kernels, and neural networks to predict the flooding loss and select the most critical variables for estimation.  }
    \label{fig:abstract}
\end{figure}

In this study (see Figure~\ref{fig:abstract}), we utilize the NFIP dataset to evaluate different regressors in counties with many claims. Among the regressors, we include neural networks ( Conditional Generative Adversarial Networks), decision trees (Extreme Gradient Boosting), and kernel-based regressors (Gaussian Processes). Also, we identify the most significant features for estimating claims' cost and introduce a scheme to improve the predictability power of the ML regressors through bias correction in the resulting distributions. The contributions of this research include the following:
\begin{itemize}
    \item Evaluation of the predictive skill of ML regressors to infer the amount paid on building flooding claims using the NFIP dataset. That includes a benchmark with a  battery of regressors, including neural network, kernel, and decision tree-based methods.
    \item A distribution bias correction scheme is introduced, which boosts the resulting determination coefficient inference on the amount paid on building flooding claims.
    \item A study is conducted on the benefit of including rainfall as a meteorological variable. That includes information on the expected improvements in inference from an extension of NFIP to external variables, including hydrological, geological,  and social.
    \item For interannual flood loss modeling, assess the performance of using the entire historical record vis a vis the record at only a fixed lag relative to the present time. Does forgetting the past improve effectiveness in the face of the already-changed climate?
\end{itemize}
The rest of this document develops as follows:
We provide an overview of the literature related to flood cost assessment.
After describing the  NFIP dataset, we detail the benchmarked ML algorithms. After that, we explain how to post-process the results of the regression prediction to correct for estimation bias, incorporate the rainfall and evaluate the performance.
\cred{Finally, we}  present and discuss the approach's experimental outcomes  \cred{and} summarize the contributions and delineate future lines of research.

\section{Related Literature}

This section reviews related literature on flood-risk assessment, NFIP data analysis, characterizing flooding using ML, and established approaches to predicting flood loss.

\subsection{Statistical-Physical Flood Risk Assessment}

In contrast to event-wise flood forecasting~\citep{basha10} using real-time data, long-term flood risk assessment is a relatively well-developed approach in certain areas. Our prior work, for example, downscales reanalysis and climate models simulated under various scenarios~\citep{IPCC2021climate} to produce a synthetic catalog of extreme events. This includes tropical cyclones~\citep{emanuel_ravela_2006,ravela_emanuel_2010,emanuel_ravela_2012} and, more recently precipitation extremes\cred{~\citep{saha2022downscaling}}. The winds and rain from synthetic events then drive storm surge and flood inundation models that predict water depths (heights). The vulnerability of exposed regions to the flood hazard projections calculates damage and loss distributions using characteristic damage and loss curves~\citep{hazus,scawthorn2006hazus}. The methodology has been used to estimate economic impacts and adaptation strategies~\citep{neumann2015joint,neumann2015risks}. Although flood risk projections are rapidly maturing through modeling, there is a substantial gap in carrying forward to projecting loss distributions. This \cred{gap} is largely due to the difficulty of scaling damages and losses, and direct data at scale appears essential. The role of \cred{artificial intelligence (}AI\cred{)} has been highlighted in addressing problems relating to geosciences~\citep{gil_pierce_19,karpatne19}, even in data paucity regimes. Here we use  \cred{ML} approaches with NFIP to construct a loss model. Although NFIP is a redacted and granular data set, it still provides a vast trove of data for constructing time-dependent dynamic data-driven loss models~\citep{dddasbook}. When paired with short-term or long-term climate projections and downscaling\cred{~\citep{saha2022downscaling}}, the corresponding short- and long-term risks can be estimated for various applications beyond insurance.

\subsection{Analysis of the NFIP Dataset }
Sun {\it et al.}~\citeyearpar{sun2020applications} present an overview of AI applications in disaster management, particularly during the phases of mitigation, preparedness, response, and recovery. They highlight the availability of datasets for analysis, including the NFIP dataset~\citep{fema_nfips}. Dombrowski {\it et al.}~\citeyearpar{dombrowski2021fima} explore this dataset and integrate it with Zillow and American Community Survey data to research factors leading to flooding insurance take-up.

Some of the NFIP dataset's current uses include assessing flood probability. For instance, Mobley {\it et al.}~\citeyearpar{mobley2021quantification} introduce a Random Forest (RF) classifier that predicts flood probability along the Texas Gulf Coast. They employ the NFIP dataset to obtain insured flood loss and high-resolution geospatial topography and hydrology data at 10 m resolution. Similarly, Zarekarizi {\it et al.}~\citeyearpar{zarekarizi2021flood} produces continuous probabilities for flooding improving over the FEMA discrete assessments.

Yang {\it et al.}~\citeyearpar{yang2021predicting} present an RF regressor that predicts the number of claims in 589 flooding events. Besides the NFIP dataset, they employ satellite-based flood extent, precipitation, coastal weather, building location, land use, topography, geomorphology, and policy count. Similarly, Lin and Cha~\citeyearpar{lin2021hurricane} develop a neural network and gradient boosting-based statistical model to predict hurricane freshwater flood loss. Their model predicts flood depth, which they input to stage damage curves to estimate the loss by aggregating over the census tracts for the hurricane event and the residential categories.

\subsection{Flooding Characterization through ML}
Some research effort has focused on predicting flood susceptible areas. For instance, employing Landsat satellite images, Wang {\it et al.}~\citeyearpar{wang2021flood} assess hybridized Multi-Linear Perceptrons (MLP) to construct a flood susceptibility map.
They identify the \cred{elevation}, slope angle, and soil as \cred{important} features.
Saha {\it et al.}~\citeyearpar{saha2021flood} propose an ensemble of ``hyper-pipes" and support vector regression (SVR) to infer flood susceptibility from satellite images, from which they extract topography and hydro-climatological predictors. Also, Janizadeh {\it et al.}~\citeyearpar{janizadeh2021novel}  identify areas prone to flooding by comparing  Bayesian Additive Regression \cred{Trees}, Naive Bayes, and RF. Siam {\it et al.}~\citeyearpar{siam2021study} perform a comparison between MIKE~\citep{patro2009flood}, a hydrodynamic model for flood forecasting, and hybridized SVR optimized with Genetic Algorithms, extracting important features with RF.

Lin and Billa~\citeyearpar{lin2021spatial} implement a predictor of flood-prone areas employing Geographically Weighted Regression (GWR). Feature selection results in stream order, drainage texture, relief ratio, bifurcation ratio, and topographic wetness position indices. Lee and Kim~\citeyearpar{lee2021scenario} predict in real-time the flood extent by estimating the amount of runoff caused by rainfall. Using past precipitation and rainfall-runoff inundation data, they employ the return period, duration, and time distribution and predict whether each grid point will flood for the simulated or observed runoff amount based on logistic regression. Finally, Persiano {\it et al.}~\citeyearpar{persiano2021comparison} compare generalized least squares (GLS) regression and top-kriging (TK) to predict at-site flood quantiles. They use the maximum annual flood in 20 catchments.

Ramasamy {\it et al.}~\citeyearpar{ramasamy2022case} compare the Linear Log Regression (LLRM) and Gumbel's Analytical (GAM) methods to assess the flood magnitude at a given return period. Their data correspond to 24 years of annual daily peak flood flow value at the Vaigai reservoir gauging station from 1995 to 2018. Parizi {\it et al.}~\citeyearpar{parizi2021linkage}  study the driver factors of discharges (FPD) at different return periods. Their analysis of more than 30 years of data from 206 gauging stations indicates that FPD increases with drainage area, heavy precipitation, and slope while decreasing with elevation and NDVI. Costache {\it et al.}~\citeyearpar{costache2021flash} identify slope surfaces with potential for flash flooding. Their approach ensemble techniques include deep learning neural networks, naive Bayes, \cred{gradient boosting}, and classification and regression trees.

Choi~\citeyearpar{choi2021development} study the construction of flood damage risk assessment using the
rainfall for 6 and 24 hours as the input for four types of nonlinear functions. Choi  \cred{regresses} over two rainfall datasets, including district size, topographic features, and urbanization rate. Meanwhile, Siam {\it et al.}~\citeyearpar{siam2021effects} study the effect of noise in the labels employed for regression with ML algorithms to map the spatial flood susceptibility. They examine different kinds of noise and explore the optimal hyper-parameter values for different hybridized ML algorithms relevant to model spatial flood susceptibility.

\subsection{Predicting Flood Loss}
The relationship between flooding and climate change is of  interest. For instance, Liu {\it et al.}~\citeyearpar{liu2021half} combined future climate scenarios with a quantitative assessment model of natural disaster risk to obtain the response to flooding events in China to 1.5 and 2$^{\circ}$ C of global warming assessing socio-economic risks of the floods, and determining the integrated risk levels. Hu {\it et al.}~\citeyearpar{hu2021using} identifies the influence of precipitation on economic flood risk by developing a linear regression model to predict precipitation patterns. They estimate financial loss risk several months in advance.

Economic loss analysis further segregates into urban and rural sectors. For instance,
Basnayake {\it et al.}~\citeyearpar{basnayake2021assessing} assess flood loss and damage in agricultural fields by evaluating the farmers' crop production outcome. Their results are robust across crop types and flood severity and independent of household characteristics. Mohammadi {\it et al.}~\citeyearpar{mohammadi2021flood} studied water loss in agricultural products after damage by flooding by employing water footprint and farming statistics. In urban settings, Schoppa {\it et al.}~\citeyearpar{schoppa2021developing} use a dataset corresponding to company loss data for 545 buildings to construct multivariate flood loss models. They tried RF, Bayesian networks, and Bayesian regression, with the first one outperforming the others. Mohor {\it et al.}~\citeyearpar{mohor2021residential} introduce a Bayesian multilevel model to estimate residential flood loss based on a dataset of 1,812 observations. Also, Nofal {\it et al.}~\citeyearpar{nofal2021modeling} model the impact of early warning systems on the reduction of flood loss. They evaluate building-level flood mitigation measures by constructing fragility functions, {\it, i.e.}, relationships that allow the propagation of uncertainty over the damage modeling process. Finally, Chen {\it et al.}~\citeyearpar{chen2021applying} fit diverse actuarial models to insurance claims data on flood damage in Taiwan transportation construction projects.

Chen {\it et al.}~\citeyearpar{chen2021does} present a systematic analysis of whether risk analysis translated to loss in flood cost assessment in southern China. Maiwald {\it et al.}~\citeyearpar{maiwald2021new} extend the Earthquake Damage Analysis Center (EDAC) flood damage model refining the description and analysis over inundation level, flow velocities, building type, and the number of building floors. To assess flood risk, Scawthorn {\it et al.}~\citeyearpar{scawthorn2006hazus} proposed to employ the flood loss rate function, with the inundated depth and loss rate as the independent and dependent variables, respectively. For places for which there is not enough data, Lv {\it et al.}~\citeyearpar{lv2021construction} introduce a method to transfer the flood loss rate data from cities with models constructed with enough information to others lacking data.

\section{NFIP Dataset}
The NFIP dataset~\citep{fema_nfips} consists of 2,547,311 records, from which we utilize 22 discrete and nine continuous variables (see  Table~\ref{tb:NFIPDataset}). As a response variable, we use the amount paid on building claims.

\subsection{Missing Data in the NFIP Dataset}
\begin{figure}
    \centering
    \includegraphics[width=6.5in]{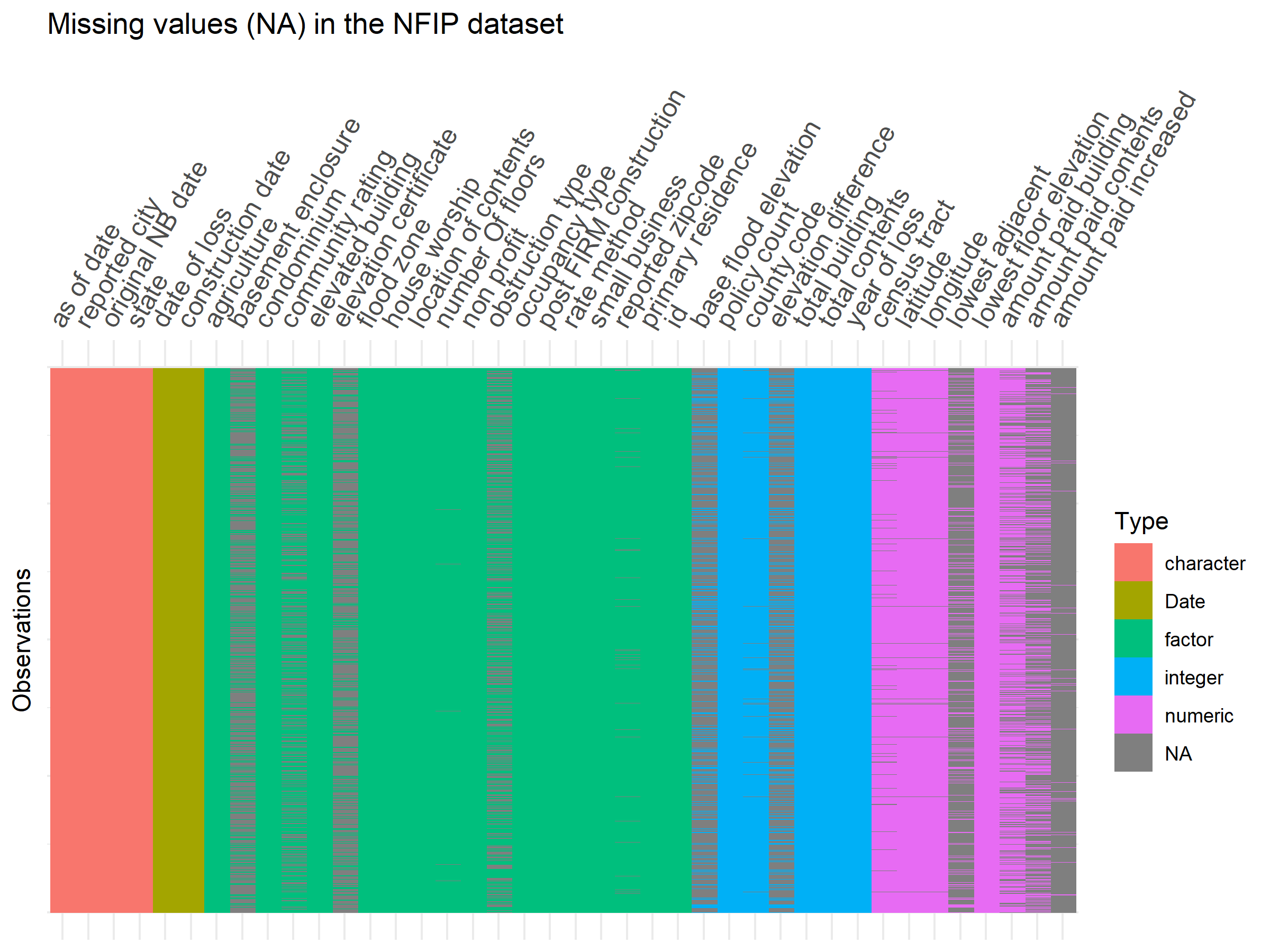}
    \caption[Missing Values Representation]{Visual representation of the missing values in the NFIP dataset. Columns and rows represent features and records, respectively. \cred{Gray} spaces represent missing values.  }
    \label{fig:missing}
\end{figure}

The NFIP database has missing data in some of its records. For instance, out of its original 40 features,  three predictors have between 15\% and 55\% of their entries with missing values, while seven have more than 55\%. The field about the amount paid on the increased cost of compliance has 97.42\% of missing values across the dataset. Only seven of the variables have no missing values. Overall, each record in the data set has an average number of missing fields of about 6.7, with one standard deviation of about 2.53. Figure~\ref{fig:missing} illustrates the distribution of missing data across features. For this research, we separately imputed the dataset for each county before using it as input for the regression models. We employed a different strategy depending on whether the variable was continuous, categorical, or a date-type. For continuous variables, we used Expectation-Maximization (EM) to fill the missing values because it has proved to be an effective method in different tasks~\citep{malan2020missing,taghavi2020concurrent}. The EM algorithm~\citep{moon1996expectation} obtains maximum likelihood parameter estimators in probabilistic models using hidden variables. Incorporating them as observable in the EM algorithm's expectation stage (step E) generates the likelihood expectation. The maximization step (step M) maximizes the likelihood parameters obtained in the E-step. The parameters from the M-step start a new E-step, and the procedure repeats itself until convergence or for an arbitrary maximum number of iterations. For categorical variables, we defined a new category for the missing values. We impute date fields by splitting them into year and day of the year, a numeric value ranging from 1 to 365 (or 366 in leap-years). Afterward, we filled in the missing values with the \cred{records} median  \cred{for} the year and day of the year and converted the completed values  to a date format.

\subsection{Preprocessing}
We adjust for inflation the amounts in dollars for the monetary features considering the Consumer Price Index from the US Bureau of Labor Statistics~\citep{CPI2021} and express them in 2020 dollars. Similarly, we create a time index variable that counts the months since January 1, 1960. Using the date of loss and original construction date to estimate the house age at the event may be interesting. However, NFIP contains some erroneous entries where the date of loss occurs before the date of construction. In this study, we subtracted 100 years from the construction date whenever this case occurred. Before constructing the regressors, we expand categorical variables using a one-hot representation. That is,  discrete values transform into binary variables where the value one indicates its presence and a zero of absence. The incorporation of dummy variables results in a dataset of 332 predictors. The dataset splits into learning (70\%) and testing (30\%) subsets. Continuous predictors are normalized using the learning dataset to have zero mean and unitary variance.

\section{Inferring Flood Loss}
We investigate representative regressors by studying the inference capability embedded in the NFIP dataset. Ensuing, we introduce a scheme for bias correction of the resulting inference distribution.

\subsection{ML Regressors}
To assess the amenability of the NFIP data set for predicting the amount paid on building claims, we tried kernel, decision trees, and neural networks-based methods. Here, we describe the ML regression algorithms employed in the benchmarking.

\noindent
{\it Gaussian Processes } \cred{(GP)} is a type of stochastic process relying on the assumption that every subset of the collection of random variables follows a multivariate normal distribution, starting from a dataset of observations $\mbX_{n\times m}\cred{ = [\mbx_1, \dots, \mbx_n]^T}$\cred{,  $\mbx_i \in {\cal{R}}^m$}, containing  $n$ observations of $m$ variables, and the \cred{corresponding responses $\mby =[y_1, \dots, y_n]^T$}. The problem is to compute the parameters $\mbphi\cred{ = \{\phi_1,\dots, \phi_{\kappa}\}}$\cred{, including the model's variance $\sigma^2$}, which best map the observations with the response variable. GP is a Bayesian approach that estimates \cred{the} parameters \cred{$\mbphi$} as
\begin{equation}
P(\mbphi\mid \mbX, \mby) =
\frac{P(\mby\mid \mbX, \mbphi) P(\mbphi)}
{P(\mby\mid \mbX)}.
\end{equation}
Once the parameters are estimated, for each new observation $\mbx^{*}$, we could estimate the response variable $y^{*}$ marginalizing over the parameters as
\begin{equation}
P(y^{*}\mid \mbx^*, \mbX, \mby)
=
\int_{\cred{\mbphi}} P(y^{*}\mid \mbx^*, \cred{\mbphi}) P(\phi\mid \mbX, \mby) d\cred{\mbphi}.%
 \label{eq:newObservation}
\end{equation}
The assumption about a Gaussian nature of the individual components in (\ref{eq:newObservation}) permits obtaining the result in closed form. Furthermore, the inverse in the mean value and the variance can be expressed in a computationally amenable form employing the Sherman-Morrison-Woodbury relationship~\citep{guttman1946enlargement}. In addition, once we use a nonlinear transformation $\mbz = \mbf (\mbx)$ on the input data, we may express the result with kernels as~\citep{prince2012computer}
\begin{equation}
\!\begin{aligned}
P(y^{*}\mid \mbx^{*}, \mbX, \mby) =
{\cal{N}}
\left[
\frac{\sigma_p^2}{\sigma^2}
\mbK[\mbx^*, \mbX]\mby -
\frac{\sigma_p^2}{\sigma^2}
\mbK[\mbx^*, \mbX]
\left(
\mbK[\mbX,\mbX]
+
\frac{\sigma^2}{\sigma_p^2} \mbI
\right)^{-1}
\mbK[\mbX, \mbX] \mby,\right.\\
\left.
\vphantom{dummy}
\sigma_p^2
\mbK[\mbx^*, \mbx^*] -
\sigma_p^2
\mbK[\mbx^*, \mbX]
\left(
\mbK[\mbX,\mbX]
+
\frac{\sigma^2}{\sigma_p^2} \mbI
\right)^{-1}
\mbK[\mbX, \mbx^*] + \sigma^2
 \right],
\end{aligned}
\end{equation}
where \cred{$\sigma_p^2$ is the prior covariance, $\mbI$ is the identity matrix, and} $\mbK[\mbx, \mbx]$\cred{, an entry in the kernel matrix,} computes implicitly the product $\mbf(\mbx)^T \mbf(\mbx)$, and can be chosen among the expressions meeting the Mercer criteria~\citep{ghojogh2021reproducing}.

\noindent
{\it Extreme Gradient Boosting }\cred{(XGB)} is constructed as a forest of decision trees. Given a loss function, such as~\citep{chen2016xgboost}
\begin{equation}
\cred{
{\cal{L}}\cred{(F)} =  \sum_{i=1}^n (y_i - F(\mbx_i))^2 +  \sum_{m=1}^M\Omega(F_m), \mbox{ where }\Omega(F_m) = \gamma T_m + \frac{1}{2} \lambda \mid\!\mid \mbomega_m \mid\!\mid^2.}
\end{equation}
Here $y_i$ is \cred{one of the $n$ } reference value\cred{s}, $F(\mbx_i)= \sum_{m=1}^M F_m(\mbx_i)$ represents the outcome of the tree ensemble model\cred{, for $F_m(\mbx_i)$ being one of the $M$ regression trees with $T_m$ leaves and \cred{$\mbomega_m$} leave weights,}   and $\Omega\cred{(F_m)}$ is a regularization term that penalizes the complexity of the model \cred{ with corresponding weights $\gamma$ and $\lambda$}. Gradient boosting creates incremental models based on previous iterations and focuses on the most challenging examples. That is, given a model prior, $F_m$, the new approximation to $y$ can be achieved by incorporating a new function $h_m$ such that
\begin{equation}
F_{m+1}(\mbx) = F_m(\mbx) + h_m(\mbx) = y,
\end{equation}
is employed, {\it i.e.}, $h_m$ can be expressed as
\begin{equation}
h_m(\mbx) = y  - F_m(\mbx).
\end{equation}
Residuals  $h_m(\mbx)$ keep some resemblance to the negative of the gradient
\begin{equation}
h_m(\mbx) \propto -\frac{\partial {\cal{L}}}{\partial F} = \frac{2}{n}( y  - F\cred{_m}(\mbx)) + \frac{\partial}{\partial F} \Omega(F\cred{_m}).
\end{equation}
\cred{Furthermore,} XGB computes a measure of importance for each feature by estimating its contribution to the overall performance. The importance measure considers how each attribute split point improves each tree's performance. The final estimation of importance is the average over all the decision trees in the model.


\noindent
{\it Conditional Generative Adversarial Networks \textnormal{(CGAN)}} extend the capability of generative adversarial network models. CGAN typically  produce data with similar characteristics to the training data by utilizing the response variable $\mby$ as input, along with a random variable $\mbz$, for both the generator $G(\mbz\mid \mby)$ and the discriminator $D(\mbx\mid \mby)$. For regression, the predictor  $\mbX$ is employed instead of the response variable. During training, the discriminator updates its parameters using the gradient of the cross-entropy. In CGAN regression, the generator and discriminator functions are optimized by gradient descent using expressions to modify the parameters as~\citep{goodfellow2020generative}
\begin{equation}
\theta_d^{+} \leftarrow \theta_d^{-} - \rho_d \nabla_{\theta_d} \left\{ \frac{1}{m}
              \sum_{i=1}^m \left(
                     \log D\left( \mbx_i\mid \mby_i \right) +
                     \log\left[ 1 - D\left(G
                             \left(\mbz_i\mid \mby_i\right) \mid \mby_i
                                     \right)
                        \right]
              \right) \right\},
\end{equation}
while the generator employs just the portion affected by the random variable \cred{$\mbz_i$}, as~\citep{goodfellow2020generative}
\begin{equation}
\theta_g^{+} \leftarrow \theta_g^{-} -
\rho_g\nabla_{\theta_g} \left\{ \frac{1}{m}
                \sum_{i=1}^m
                     \log\left[ 1 - D\left( G
                                  \left(
                                       \mbz_i\mid \mby_i
                                  \right) \mid \mby_i
                                      \right)
                         \right]
           \right\},
\end{equation}
for suitable learning rates $\rho_d$ \cred{and} $\rho_g$.

\subsection{Bias Correction}
Once observations infer response values, a parametric model fit to the distribution extracts the properties embedded in the predictions, which may be critical for assessing flood loss. As it is common in econometrics and risk statistics analysis, we approximate the distribution of the resulting predictions of the regression models using Burr (otherwise known as Burr Type XII, Burr, or Singh-Maddala)~\citep{chen2021applying}. The Burr distribution has a probability density function (pdf) given as
\begin{equation}
f(y; c, k, \cred{\lambda}) =
\frac{ck}{\cred{\lambda}}
\left(
\frac{y}{\cred{\lambda}}
\right)^{c-1}
\left[
1 + \left(\frac{y}{\cred{\lambda}}\right)^{c}
\right]^{-k -1},
\end{equation}
with parameters $k, c$ for shape and  $\cred{\lambda}$ for scale. Given the
Burr's cumulative probability function (CDF) $F$ for the predicted values
\begin{equation}
F(y_p\cred{; c_p, k_p, \lambda_p}) = 1 -
\left(
1 + \frac{y^{c_p}}{\cred{\lambda}_p}
\right)^{-k_p},
\end{equation}
one could approximate a reference distribution $F\cred{(y_g;  c_g, k_g, \lambda_g)}$ by assuming equality for corresponding response values $y_p$ and $y_g$ as $F(y_g\cred{; c_g, k_g, \lambda_g}) = F(y_p\cred{; c_p, k_p, \lambda_p})$. One may correct for bias using the  expression
\begin{equation}
y_g = \cred{\lambda}_g
\left\{
   \left(
      1 - F(y_p\cred{; c_p, k_p, \lambda_p})
   \right)^{-1/k_g} - 1
\right\}^{1/c_g},
\end{equation}
for predicted and reference values, $y_p$ and $y_g$, respectively. The parameters for the Burr distribution may be fit using Maximum-Likelihood. When Maximum-Likelihood fails to converge, a Weibull distribution is fit instead. The definition of Weibull distribution is
\begin{equation}
g(y; k, \cred{\lambda}) =
\frac{k}{\cred{\lambda}}
\left(
\frac{y}{\cred{\lambda}}
\right)^{k-1}
e^{-(y/\alpha)^k},
\end{equation}
with parameters $k$ for shape and  $\cred{\lambda}$ for scale. Given the  Weibull's CDF $G_p$ for the predicted values
\begin{equation}
G(y_p\cred{; k_p, \lambda_p}) = 1 - e^{-(y_p/\cred{\lambda_p})^{k_p}},
\end{equation}
one can approximate a reference distribution $G\cred{(y_g; k_g, \lambda_g)}$ by assuming equality for corresponding response values $y_p$ and $y_g$ as $G(y_g\cred{; k_g, \lambda_g}) = G(y_p\cred{; k_p, \lambda_p})$. Using that assumption, one may correct for bias using the following closed-form expression
\begin{equation}
y_g = \alpha_g
\left\{ -
\log
   \left(
      1 - G(y_p\cred{; k_p, \lambda_p})
   \right) \right\}^{1/k_g},
\end{equation}
for predicted and reference values, $y_p$ and $y_g$, respectively.

\section{Experimental Setup}
We analyze the NFIP by shifting and expanding periods to assess the regressors developed. We also study how significant physical variables could enrich NFIP. Finally, we detail the performance metric employed for assessment.
\subsection{Analysis Periods}
In many applications of \cred{ML} regression, we start with a dataset split into training and testing partitions. The assumption is that the elements in the test split constitute a faithful representation of the data distribution where the learning has to occur, {\it i.e.}, the task of the ML method is to perform well within the distribution defined by the dataset.  This approach is not practical for predicting losses in general. In this problem, we want to learn from the past and assess events in the future. Thus, simulating flooding and claims can generate inferences about these possible scenarios.

To build a regressor for a county,  we train with the claims covering the period from years ${b+\delta}$ to ${b+o+k-1}$, inclusive, and test with claims from year ${b+o+k}$, the immediate next one. Here, $b$ represents a baseline year, $o$ is an offset, and $k$ is an iterator for integer variables where $k\geq 0$ and $o\geq1$. The variable $\delta$ has two possible values. For shifting periods of analysis, $\delta = k$, meaning that the period of $o$ years employed for training shifts with new intervals. For expanding analysis periods, $\delta = 0$, the claims considered start on year $b$ and end on year $b+o+k$, for the $k$-th iteration considered.

\subsection{Incorporating Physical Variables}
\begin{figure}
    \centering
    \begin{tabular}{cc}
    \includegraphics[width=3in]{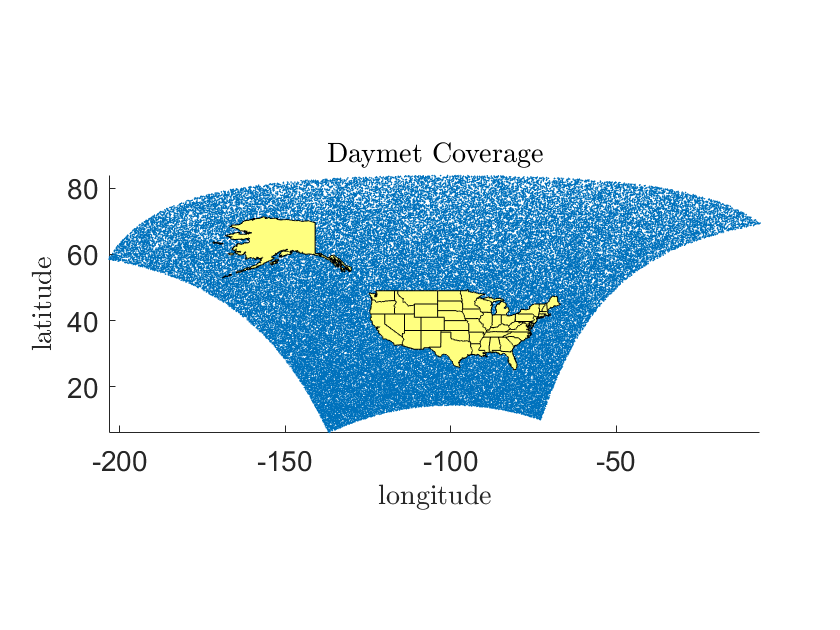} & \includegraphics[width=3in]{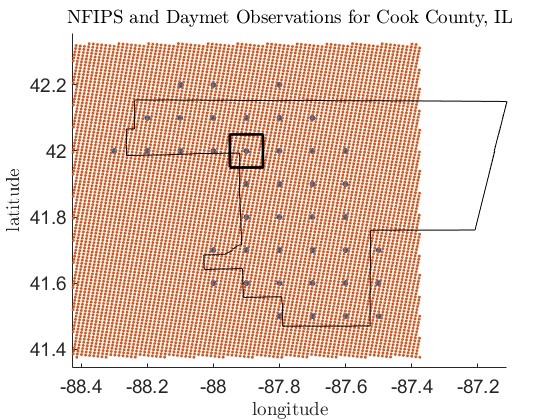}\\
    \begin{minipage}{3in}
    \centering
    (a) Daymet geospatial coverage
    \end{minipage}&
    \begin{minipage}{3in}
    \centering
    (b) NFIP (blue) and Daymet (red) geolocations for Cook county, IL (black polygonal shape).
    \end{minipage}\\
    \end{tabular}
    \caption[Daymet and NFIP]{Daymet is a dataset with daily weather parameters for North America, Hawaii (not shown), and Puerto Rico (not shown) (a). The resolution for Daymet is 1 km $\times$ 1 km, while NFIP has a resolution of $0.1^{\circ}$ (b). We incorporate Daymet precipitation by aggregating within non-overlapping squares around the NFIP location.  }
    \label{fig:daymet_nfip}
\end{figure}

Rainfall may strongly influence economic loss~\citep{hu2021using}, which we aim to quantify within the context of NFIP. To that end, we incorporated the Daymet~\citep{daymet2020} dataset, a daily estimation from 1980 to 2021 of weather and climatology variables. The variables are available in a 1 km $\times$ 1 km regular grid (see Figure~\ref{fig:daymet_nfip}). They include the maximum and minimum temperature, precipitation, shortwave radiation, vapor pressure, snow water, and day length.

Since NFIP has a lower resolution, we incorporate the Daymet precipitation by aggregating the estimates around the geolocation of the claim at $\pm 0.05^{\circ}$ of longitude and $\pm 0.05^{\circ}$ of latitude. We tried six aggregation schemes, including $\Sigma_3$, $\Sigma_5$, $\Sigma_7$, $\max_3$, $\max_5$, and $\max_7$, corresponding to the sum of rain two, four, and six days before,  and the maximum for the two, four and six days earlier, in all cases including the day of the event. Once added to the predictors, the precipitation is normalized to have zero mean and unitary variance, just as the other continuous variables.

\subsection{Performance Evaluation}
To assess the performance of the different ML regressors, we utilize metrics reflecting the pointwise evaluation of inferences and the results we obtain by modeling the distributions of predictions.

\subsubsection{Pointwise Performance Evaluation}
To measure the pointwise similarity between the predictions for the response variable and the corresponding reference values, we employ indicators including the root-mean-squared error (RMSE), the RMSE divided by the standard deviation $\sigma$ of the response value, and the coefficient of determination. Given a data set $\{\mbX, \mby\}$, the unbiased estimation for the variance  of the ground truth response variable is given by
\begin{equation}
    \sigma^2 = \frac{1}{n-1}  (\mby - \mu)^T (\mby - \mu),
\end{equation}
where $\mu$ is its mean value and $n$ the number of elements in $\mby$. In regression, a common metric is RMSE, which measures the difference between the predicted values $\mby_p$ and the reference values $\mby_g$ using
\begin{equation}
\mbox{RMSE} =  \sqrt{\frac{1}{n} (\mby_p - \mby_g)^{T} (\mby_p - \mby_g)}.
\end{equation}
 Since the RMSE corresponding to different counties may result in a wide range of values, it is convenient to RMSE relative to the standard deviation as RMSE/$\sigma$. If this ratio is more than one, using the mean as the resulting inference would result in better performance than the outcome of the ML algorithm, while if it is below one, it is better to use the ML method.

\subsubsection{Comparison between Distributions}
Once a parametric continuous distribution approximates the data, there is a need to verify the goodness of fit. Among the several options available~\citep{rayner2009smooth}, we select the  Kolmogorov-Smirnov (K-S) test because it can be applied to continuous distributions and works best when the number of observations is in the order of thousands. For two CDFs $P_r$ and $Q_p$, corresponding to the reference and prediction CDFs, the test employs the supreme of the difference to construct its test statistic, {\it i.e.},
\begin{equation*}
D_n = \max_x \mid P_r(x) - Q_p (x) \mid.
\end{equation*}
In the K-S test, the null hypothesis $H_0$ claims that the observations under consideration come from the same underlying distribution. One rejects the null hypothesis whenever  $\sqrt{n} D_n$ is larger than a certain critical value  $K_{\alpha}$, where $n$ is the number of samples, and $\alpha$ reflects a level of confidence, {\it i.e.}, the area of the Kolmogorov pdf $p_K$ beyond the threshold $K_{\alpha}$ is smaller than $\alpha$. Otherwise, we are assuming that  $p_K\left(\sqrt{n} D_n \leq K_{\alpha}\right) = 1 - \alpha$.  In practice, one rejects the null hypothesis when the $p$-value of the test is smaller than the significance level $\alpha$.

After testing the goodness of fit of the parametric pdf, we assess the difference between the reference and prediction distributions using the  Kullback-Leibler \cred{(KL)} divergence as
\begin{equation}
D(\mbp\mid\mid \mbq) =
\int_x p_r(x) \log
\left(
\frac{p_r(x)}{q_p(x)}
\right) dx,
\end{equation}
where $p_r(x)$ and $q_p(x)$ correspond to the reference and inferred pdfs.  Another performance measure consists of constructing a classifier to distinguish the distributions as considered different classes. Let the classification space ${\cal{C}} = \{{\cal{P}}, {\cal{Q}}\}$ for the reference and the prediction classes, respectively, with corresponding characterization  $p_r(x)$ and $q_p(x)$. In the classification problem, given the observation of the predictor $x$, the objective is to assign it to the right class. In general, the classifier has to define whether an observation belongs to class ${\cal{P}}$ with probability \cred{$p$} or to class ${\cal{Q}}$ with probability $1-p$. To indicate similarity between the distributions, we rely on regression gradient boosting trees~\citep{friedman2001greedy}.

Another insightful measure of performance is the determination coefficient $R^2$, representing the proportion of the variability of the independent variable that is explainable in terms of the dependent ones. After bias correction, when there is a parametric representation of the reference $p_r(x)$ and prediction $q_p(x)$ distributions,  we express $R^2$ as
\begin{equation}
 R^2 = 1 - \frac{S_r}{S_v} = 1 - \frac{\displaystyle{\int_x \left(p_r(x) - q_p(x) \right)^2 dx } }{\displaystyle{\int_x \left(p_r(x) - \mu_r \right)^2 dx}},
\end{equation}
where $S_r$ is the difference between the reference and prediction distributions, and $S_v$ is the deviation of the reference distribution from its mean value $\mu_r$. Chicco {\it et al.}~\citeyearpar{chicco2021coefficient} argue that since the interval $[0,1]$ bounds the range of $R^2$ and because its value reflects the number of correctly-predicted elements, its use should be preferred based on its informative capability.

\begin{figure}
    \centering
    \includegraphics[width=6in]{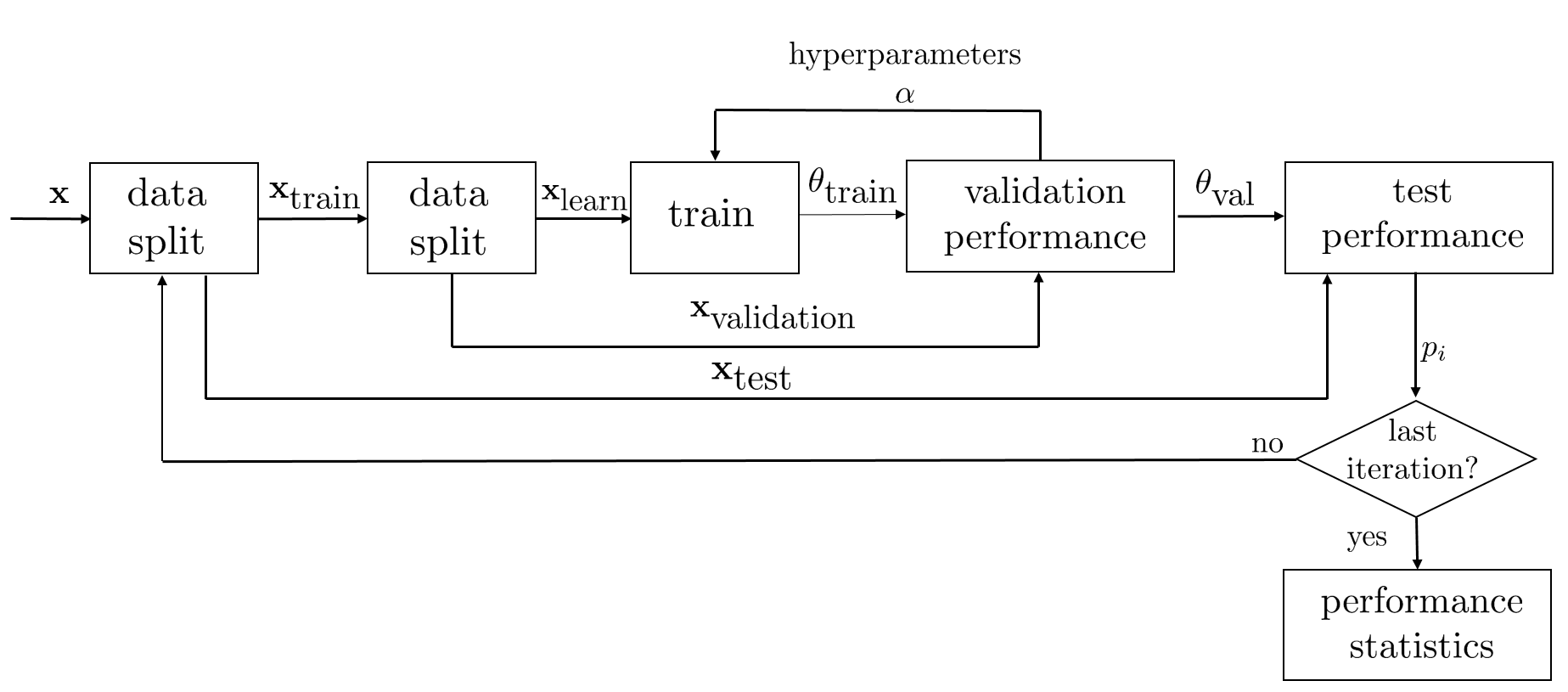}
    \caption[Dataset protocol]{Dataset protocol. The dataset partitions into training (70\%) and testing (30\%) partitions. The training dataset further has a learning (70\%) and validation partition (30\%). The first split is employed to fine-tune the hyperparameters. After training, performance results $p_i$ are obtained using the testing partition. Repeating this exercise 30 times estimates performance statistics.
   }
    \label{fig:performance_protocol}
\end{figure}

\section{Results}
To assess the predictive capabilities of the NFIP dataset, we implemented the method described in the previous sections and tested them on several scenarios. We standardized the continuous predictors using the training dataset, dividing the difference from its mean by the standard deviation, and created one-hot representations of the categorical variables. To assess the different regressors,  we selected 11 US counties with many claims in the NFIP  dataset for comparison  (see Table~\ref{tb:counties}).

\begin{table}[]
 \caption[Counties with many claims employed in this study]{Counties employed in assessing ML regressors because of their large number of claims.}
    \label{tb:counties}
    \centering
    \begin{tabular}{rccr||rccr}
    \multicolumn{1}{c}{{\bf n}}    &  {\bf code}& {\bf name} & \multicolumn{1}{c}{{\bf claims}} &  \multicolumn{1}{c}{{\bf n}}    &  {\bf code}& {\bf name} & \multicolumn{1}{c}{{\bf claims}}   \\
1 &            12011 &Broward, FL & 31,059& 7 &  34029 &Ocean, NJ  & 52,436 \\
2 &  12086 & Miami-Dade, FL & 61,197& 8 &   36059 &Nassau, NY & 50,067  \\
3 &   17031 &Cook, IL & 15,180& 9 &   36103 &Suffolk, NY  & 33,130  \\
4 &  22051 &Jefferson Parish, LA & 133,162 & 10 &   48167 &Galveston, TX  & 60,224  \\
5 &   22071 &Orlean Parish, LA &  126,405 & 11 &  48201 &Harris, TX  & 171,202\\
6 &   22103 &St Tammany Parish, LA & 37,514 \\

    \end{tabular}

\end{table}

\subsection{Setting Up, Fine Tuning  and Comparing the Regressors}

Initially, we compared the regressors using the claims dataset corresponding to each of the counties under analysis. In this stage (see Figure~\ref{fig:performance_protocol}), we took 70\% of the data for training and 30\% for testing. Then, we split the training dataset into learning (70\%) and validation (30\%) subsets. After evaluating the testing split, we obtained a performance assessment $\mbp_i$ for this partition. After repeating this procedure 30 times, we estimated the performance statistics for each regressor in each county.

\paragraph{\textnormal{\bf XGB}}  For this regressor, we used the XGBoost version 1.4.1.1 in R. The parameters to fine-tune include
\begin{enumerate}
\item $\eta\sim{\cal{U}}(0.0001, 1)$, the information from a new tree employed during boosting.
\item  $c_s\sim{\cal{U}}(0.1, 1)$, the fraction of the variables considered during branch splitting.
\item $d_m\sim{\cal{U}}(2, 10)$, the maximum depth of the trees;
\item $s_s\sim{\cal{U}}(0.1, 1)$, the percentage of the data employed to grow the tree, in what is known as stochastic boosting; and
\item $\gamma\sim{\cal{U}}(0.01, 100)$, the minimum reduction in the loss function that is needed to create a tree.
\end{enumerate}

We find these hyperparameters via a uniform random search over a range. Using a particular set of parameters, we trained an XGB regressor using the first split of the training/testing datasets taking 70\% of the training samples for learning and validating with the remaining 30\%. We kept the parameters for which, after 50 rounds, the root-mean-square error was relatively insignificant. A random search for the best parameters during 1,000 cycles for each county. Afterward, we trained for 100 iterations with the hyperparameters providing the best performance.

\paragraph{\textnormal{\bf GP}} In this case, we optimized the hyperparameter using the Matlab {\tt fitrgp} method, selecting the basis function among a constant, linear, and quadratic. We sought the optimal kernel scale between $r \cdot (10^{-3}, 1)$, where $r$ is the maximum predictor range. The value of sigma was \cred{sought} in the range $(10^{-4}, \max(10^{-3}, 10 \cdot \sigma_y))$, for $\sigma_y$ being the standard deviation of the response.  Also, we determined whether to standardize or not. In the former case, we divided the difference between the predictor and its mean by standard deviation.    For the optimization, we employed five-fold cross-validation on the training dataset.

\paragraph{\textnormal{\bf CGAN}} We defined a fixed learning rate for the discriminator and generator of 0.001, trained with Adam the latter and Stochastic Gradient Descend the former. We employed the Exponential Linear Unit  (ELU)  as the activation function, training for 20 epochs with a batch size of 128. The stochastic input has a length of one. Our discriminator receives as input the predictors and is fully connected to a layer with 100 units, as does the stochastic value. Then, there are four fully connected layers, each with 50 units, ending with a single output with a sigmoid activation function. The generator and discriminator use a cross-validation loss function. The generator has the same architecture except for a linear output layer.

\begin{figure}
    \centering
    \includegraphics[width=4in]{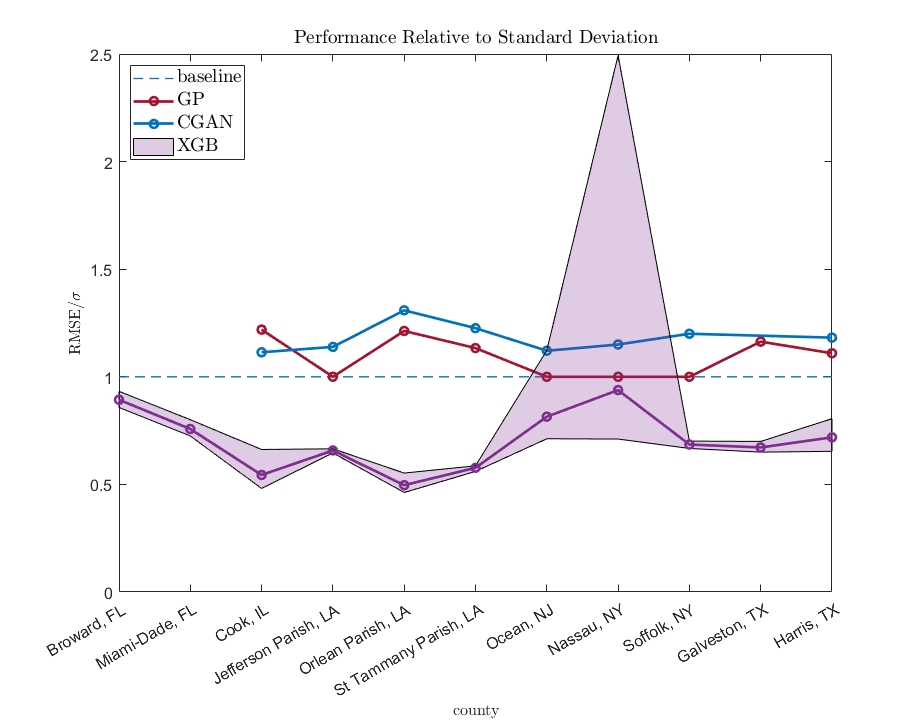}

    \caption[Regressors' Performance]{Comparison of regressor performance. XGB generated regressors with a ration RMSE/$\sigma$ below 1.0, albeit with at least one partition above it for Ocean and Nassau counties. The shaded area delimits the best and worst performance observed during the 30 dataset splits. GP and CGAN produced regressors with RMSE/$\sigma$ ratios above 1.0, except for Broward and Miami-Dale, for which counties the methods did not converge to create a regressor.

    }
    \label{fig:regressors_performance}
\end{figure}

After training the regressors, we computed the prediction for the elements in the training set and assessed the performance using the testing partition (see Figure~\ref{fig:regressors_performance}). We notice that the XGB was the only regressor consistently providing results below one for the ratio RMSE/$\sigma$ for almost all the counties under consideration. The exceptions were Ocean (NJ) and Nassau (NY) counties, for which XGB obtained a mean value of 0.8147 and 0.9382 but a maximum of 1.1227 and 2.4937. On the contrary, neither CGAN nor GP converged to a value smaller than one for the ratio RMSE/$\sigma$. Furthermore, they could not converge on a solution for Broward and Miami-Dade. For instance, the best results for GP were Jefferson Parish, Ocean, Nassau,  and Suffolk, with an RMSE/$\sigma$ ratio of 1.0. Based on these results, for the rest of the article, we will focus on the description of the predictive capabilities of the XGB regressor.

\subsection{Bias Correction}
\begin{figure}
    \centering
    \begin{tabular}{cccc}
    \includegraphics[width=1.6in]{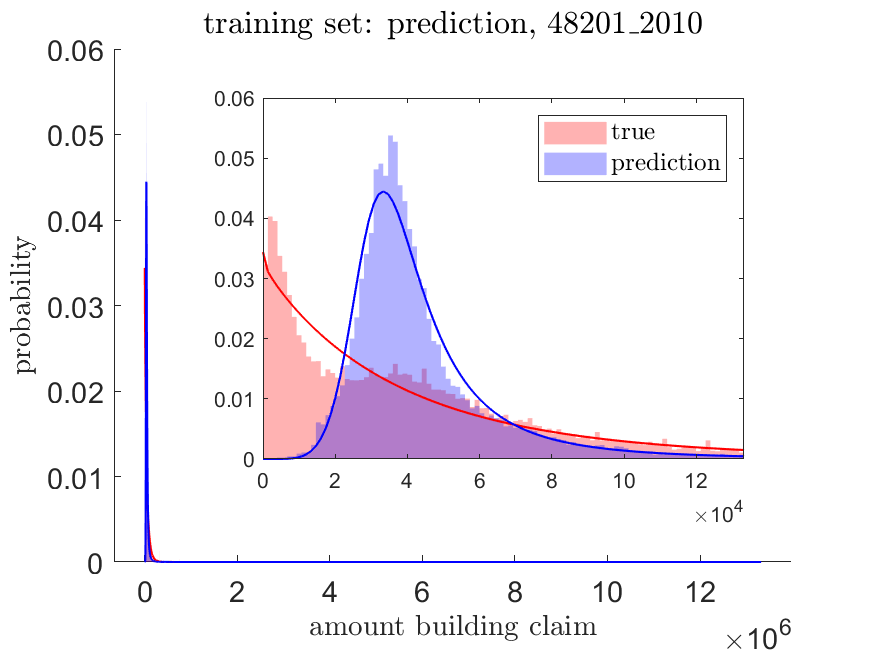} & \includegraphics[width=1.6in]{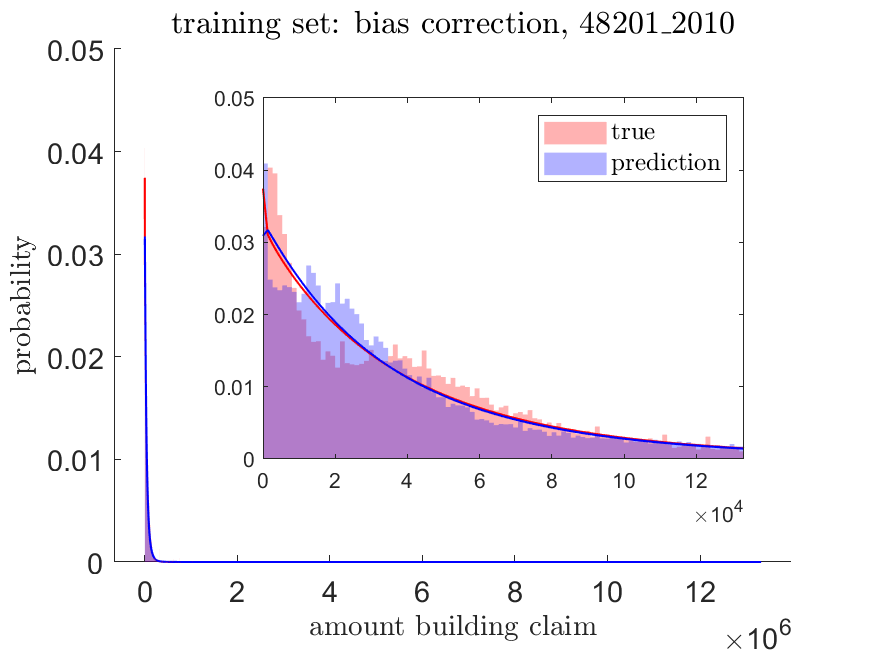} &
    \multicolumn{2}{c}{\includegraphics[width=1.6in]{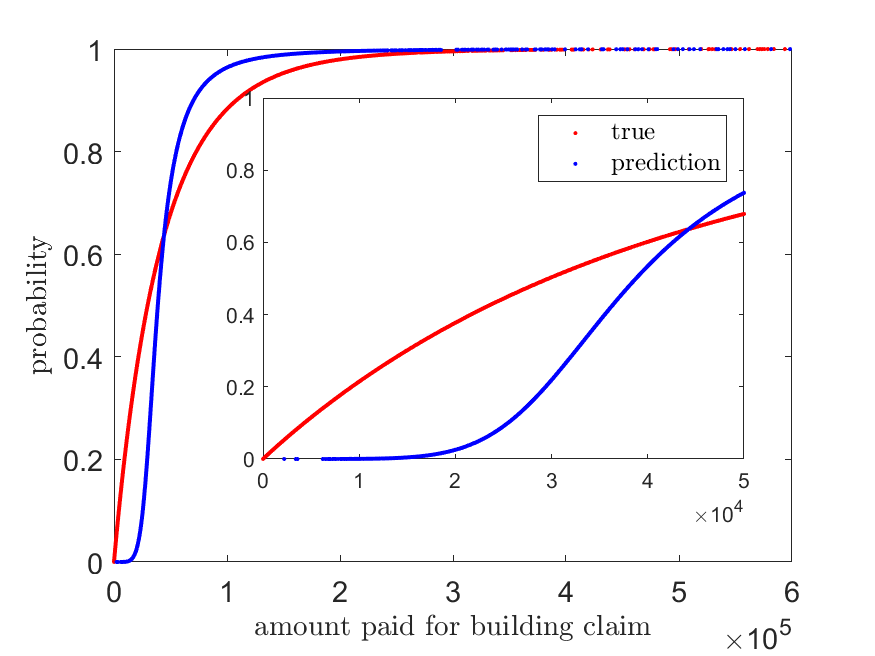}} \\

     (a)  XGB initial & (b) XGB final &
     \multicolumn{2}{c}{(c)  CDF matching}   \\

        \includegraphics[width=1.6in]{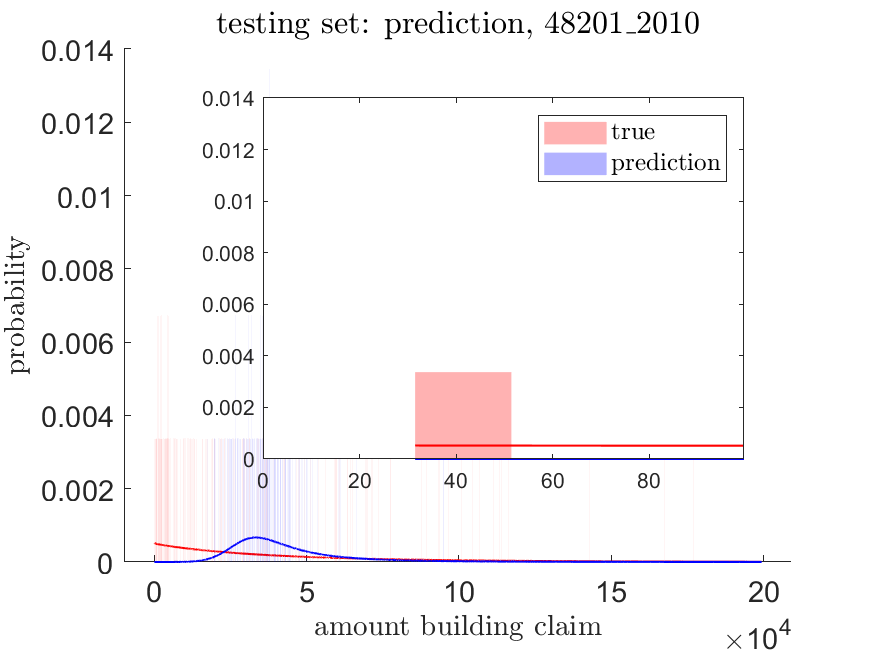}  &
         \includegraphics[width=1.6in]{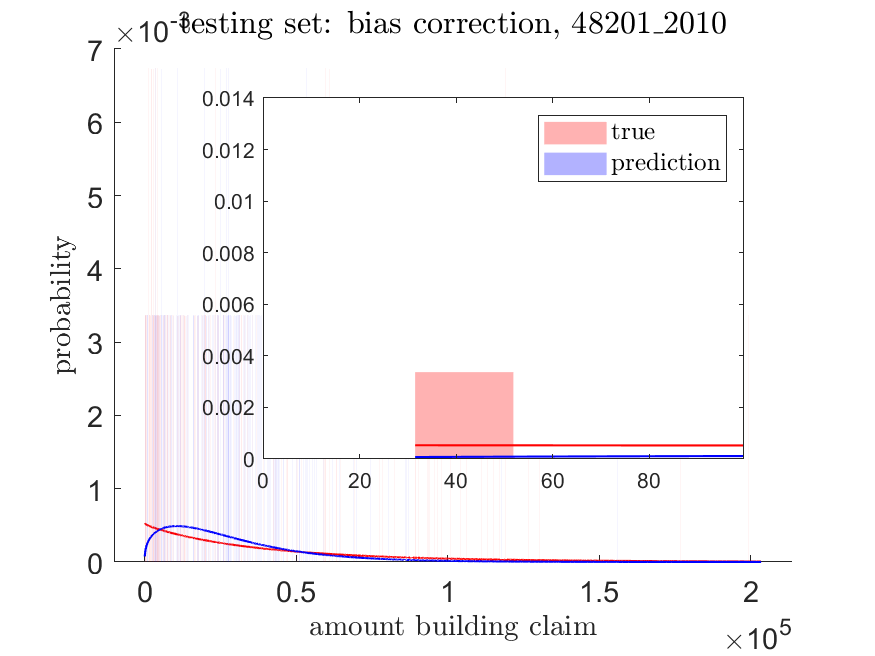}
         &
         \includegraphics[width=1.6in]{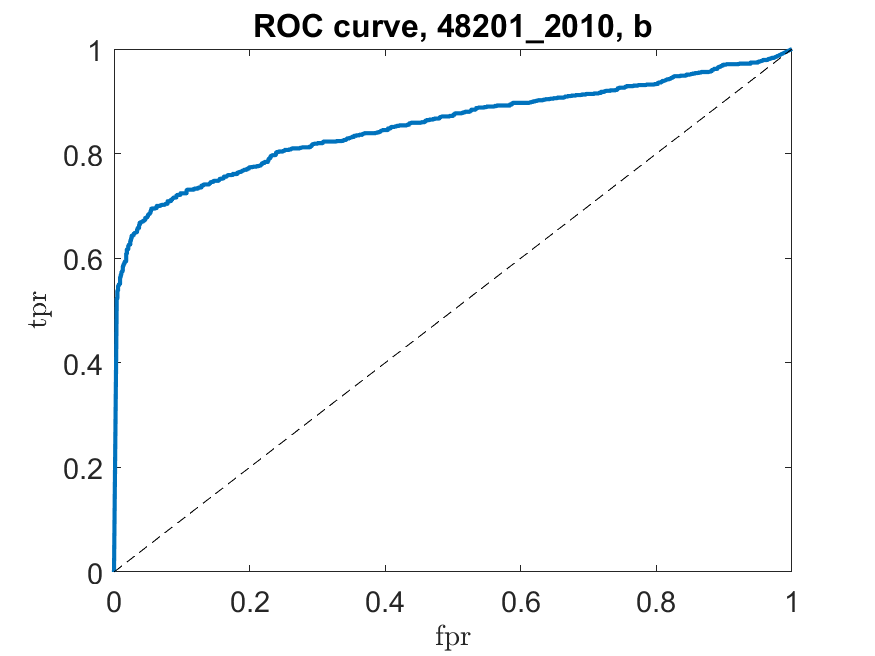} & \includegraphics[width=1.6in]{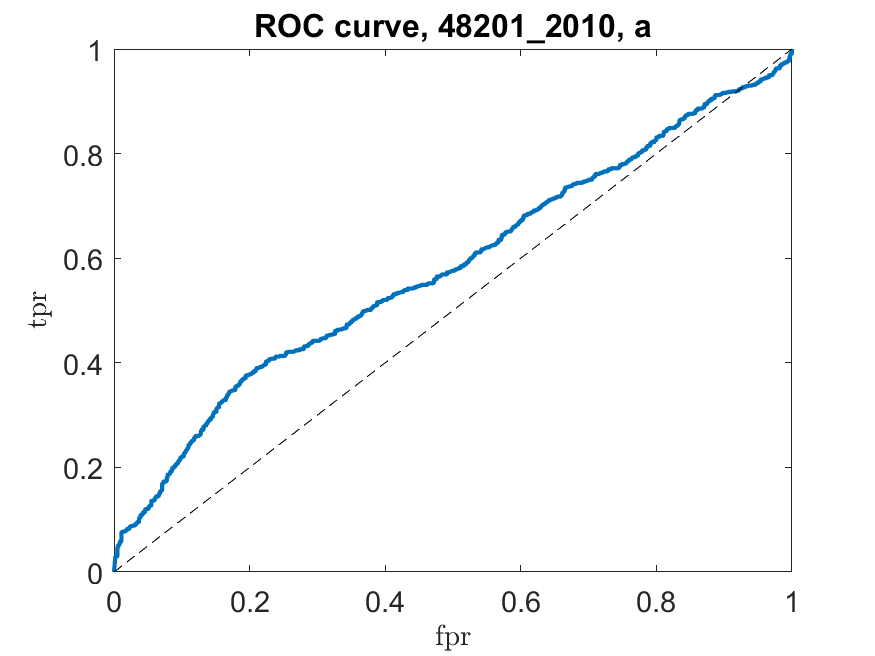}  \\

      (d) XGB test initial &

     (e) XGB test final & (f) ROC initial & (g) ROC final \\

    \end{tabular}
    \caption[Bias Correction]{Bias Correction. Once the regressor trains (a), we obtain the distribution of its outputs and fit a parametric probability density function (pdf). We adjust the distribution to match the reference and training output prediction (b). The alignment relies in the inverse cumulative density function method (c). The method is then applied to the test split of the dataset (d)-(e). A possible form to assess the resulting match is by constructing a classifier to distinguish between the populations before (f) and after (g) bias correction.}
    \label{fig:bias_correction}
\end{figure}

During bias correction, we aim to fit a Burr distribution to the set of predictions, trying a Weibull when the minimization procedure fails to converge. In Figure~\ref{fig:bias_correction}, the method is illustrated for Harris county (48201) using the shifting period between 2000 and 2009, where the test dataset corresponded to 2010. For the predictions resulting from the evaluation of the training dataset, containing 47,160 claims, we fit a Burr distribution to the reference and predicted resulting in parameters $\alpha=  444,293$, $c = 0.99$, and $k =  10.42$ and  $\alpha =  31,832.9$, $c =  6.18$, and $k = 0.47$. With a $p$-value of $6.5\times 10^{-35}$ and $1.37\times 10^{-19}$, we reject the null hypothesis between the parametric model and the data distribution using the \cred{K-S} test. We estimated the transformation between the reference and prediction-adjusted parametric distributions using the CDF and transformed the predicted values. Again, we fit a Burr distribution to these values, resulting in the parameters  $\alpha =  of9,294$, $c = 1.00$, and $k = 6.38$. In this case, the K-S test resulted in a $p$-value of $5.43\times 10^{-15}$.

  During testing, with a dataset of 297 claims, we modeled the reference and prediction distribution, finding the parameters $A =  28,570.6$, $B = 0.78$ and $\alpha = 36,950.7$, $c = 6.87$, and $k = 1.04$. Note that for the reference distribution, a Weibull distribution was fit. The function learned during training transforms observations corresponding to the predictions. The resulting distribution was fit using a Burr distribution with parameters $\alpha =  131,763$, $c = 1.37$, and $k = 7.68$. This time the K-S test resulted in a $p$-value of 0.83. Hence, we did not reject the null hypothesis. The initial distribution had a KL divergence of 0.0123, while the final distribution had a KL divergence of 0.0015. Similarly, the initial ROC AUC had a value of 0.85, while the final ROC AUC had a value of 0.58.

\subsection{Analysis Periods}
We studied the behavior of the XGB regressor under shifting and expanding periods. In both cases, 2000 was the baseline employed, while 2020 was the last analysis year. For {\it expanding} periods, we initially took ten years, {\it i.e.}, used the period from 2000 to 2009 for training and 2010 for testing. Then, we selected the period from 2000 to 2010 for training and 2011 for testing. We kept doing this until we ultimately chose the period from 2000 to 2019 for training and tested with the claims in 2020. We repeated this exercise for the 11 counties under study. For shifting periods, we initially took the period from 2000 to 2009 for training and tested it in 2010. Contrasting with expanding periods, in the next iteration, we took from 2001 to 2010 for training and tested it in 2011. In the last iteration, we trained with data from 2010 to 2019 and tested with 2020. To assess the $R^2$ performance of the XGB regressor, we weighted the value of the performance indicator by the number of claims employed for testing.

\paragraph{Shifting and Expanding Periods}
For shifting periods (see Figure~\ref{fig:shifting_window} in the additional material section), the combination of XGB and bias correction produced regressors most of the time except in 2012 for Orlan Parish (22071). For five counties, they fail to generate a regressor for one year, different in each case. $R^2$ varies within a county. It can be as low as 0.15 for Cook county in  2014 or as high as 0.98 in 2015 for Nassan (36059). Correspondingly, the parametric fit was, in general adequate. For instance, the $p$-value did not fall below $\alpha= 0.05$ for Jefferson Parish (22051), Orlean Parish (22071), and St Tammany Parish (22103), but it fell below $\alpha=0.05$ for Harris county (48201). The performance for expanding periods was lower (see Figure~\ref{fig:expanding_window} in the additional material section). For instance, Jefferson Parish (22051) had only five useful regressors in the analysis period. The resulting values are uncorrelated with a linear correlation coefficient of -0.064  between the corresponding $R^2$ values of shifting and expanding periods.   However, weighting by the number of claims per county in a given year, the $R^{2}$ value for the shifting and expanding window was  0.61 and 0.7, respectively.

\paragraph{Shifting and Expanding Periods with Rainfall}
To study the incorporation of Daymet, we obtained the rain for two, four, and six days before, including the date of loss for Cook county (17031), and characterized it with their sum and the maximum over the period. To aggregate rainfall, we incorporated Daymet rainfall for the area, including $\pm 0.05^{\circ}$ in latitude and longitude around the NFIP claim location. Then, we constructed six regressors, incorporating rain's corresponding characterization as a predictor. For the sums, we obtain an $R^2 = 0.516$, and for the maximums, $R^2 = 0.515$. Thus, we arbitrarily decided to use the sum of the millimeters of rain falling around the claim's location for the two days before, including the date of loss.

Then, we decided to study the implications of using the rainfall information for the other counties under study. Figures~\ref{fig:shifting_window}~\ref{fig:expanding_window} illustrate the results. For shifting periods, XGB could fit a regressor with a ration RMSE$/\sigma$ for all the test years in Orlean Parish county. Although it fit ten regressors for three counties, it fit nine regressors for nine counties. Using the weighted with the claims $R^2$ values, we obtain a linear correlation coefficient of 0.096 between the shifting and expanded periods of analysis with Daymet information included; a correlation of 0.056 between the shifting without Daymet and expanding with it; and a correlation of -0.215 between the shifting periods with Daymet and expanding periods without it. The $p$-value was generally good, with few cases per county rejecting the null hypothesis of the goodness of fit with the parametric Burr distribution. In the case of the shifting period plus Daymet, the only exception was Jefferson Parish (22051), where the $p$-value was larger than 0.05 through the analysis period. In the case of expanding periods plus Daymet, we observed $p$-values below 0.05 for Nassau (36059), Suffolk (36103), Galveston (48167), and Harris (48201) counties. Weighting by the number of claims for each county in a given year, the $R^{2}$ value for the shifting and expanding period with Daymet information included was  0.61 and 0.7, respectively.

\begin{figure}[t]
    \centering
    \begin{tabular}{cc}
    \includegraphics[width=3in]{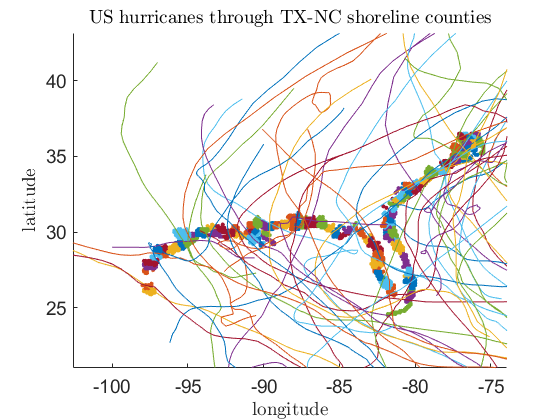} & \includegraphics[width=3in]{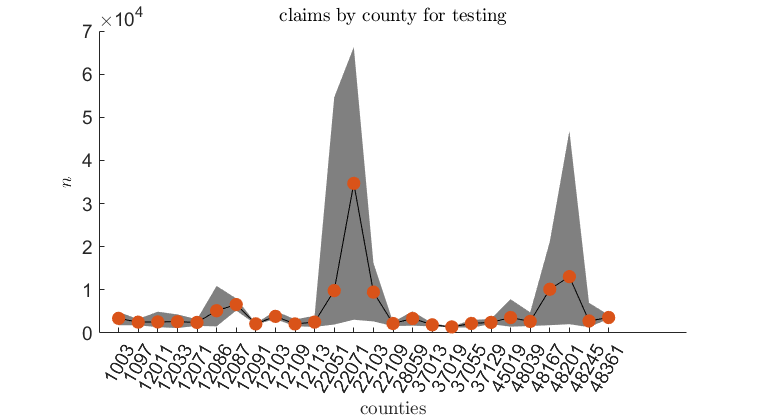}\\
    \begin{minipage}{3in}
    \centering
    \begin{minipage}{3in}
    \centering
    (a) Hurricane trajectories from 1980 (Allen) to 2017 (Maria).
    \end{minipage}
    \end{minipage}&
    \begin{minipage}{3in}
    \centering
    (b) Maximum, minimum and average number of claims in the LOO-CV construction of regressors.
    \end{minipage}\\
        \includegraphics[width=3in]{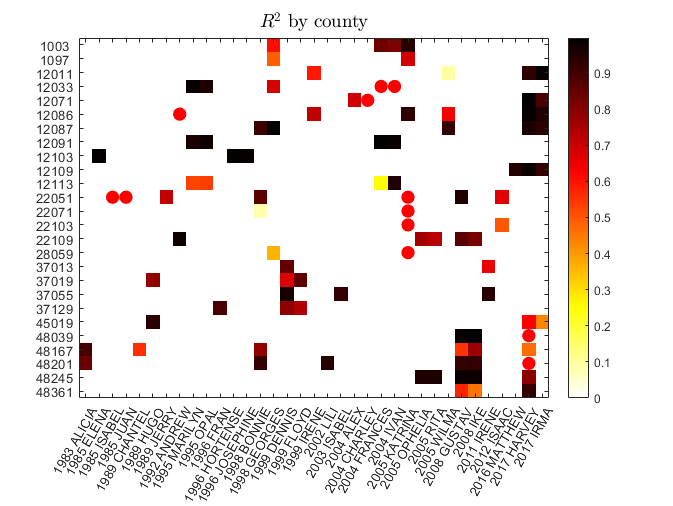} &
         \includegraphics[width=3in]{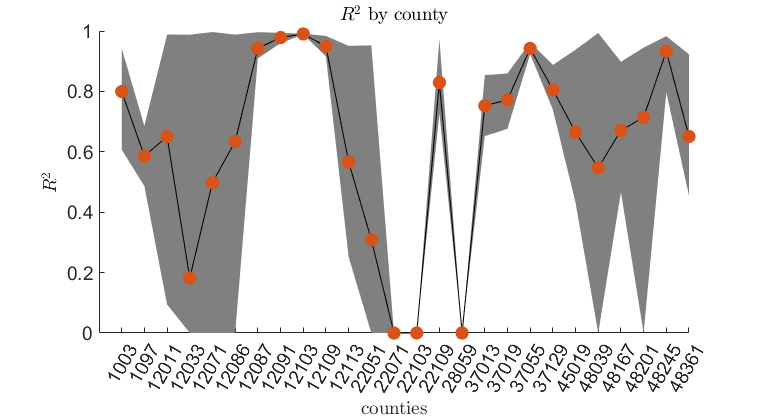}\\
    \begin{minipage}{3in}
    \centering
    (c) Maximum, minimum and average $R^2$ in the LOO-CV construction of regressors
    \end{minipage}&
    \begin{minipage}{3in}
    \centering
    (d) $R^2$ by county and hurricane. The circled cells signal places that lack a useful regressor.
    \end{minipage}\\

    \end{tabular}
    \caption[Comparison]{\cred{C}onstruction of regressors by the county during the analysis of significant flooding events due to hurricanes. Hurricanes affect shoreline counties (a) and generate significant flooding events (b). Using Leave-One-Out (LOO) Cross-Validation (CV), we selected one hurricane affecting a county and the remaining ones for testing (c). The resulting mean $R^2$ value is 0.807 (d).    }
    \label{fig:county-hurricanes}
\end{figure}

\subsection{Comparing with Alternatives}
Research on the NFIP dataset to predict loss has been scarce. An exception is a work by Lin and Cha~\citep{lin2021hurricane}. In their approach, Lin and Cha start simulating hurricane storm tracks. Then, they use a rainfall prediction model to simulate rainfall intensity, predicting the stream stage increment and the flood depth at the centroid of census tracts. Finally, they aggregate flood loss using stage-damage curves, obtaining an $R^2$ value of 0.3382. To assess, they employ FEMA-classified significant flooding events, those generating at least 1,500 claims, in the eight coastal states in the US South-West, from Texas to North Carolina for the period from 1980 (hurricane Allen) to 2017 (hurricane Maria) (see Figure~\ref{fig:county-hurricanes} (a)). These eight states include 121 counties on the shoreline generating 1,246,888 claims on NFIP.

Tracing the trajectories of hurricanes, we detected the counties affected by significant flooding events. Using a leave-one-out (LOO) training strategy with cross-validation (CV), we took one storm for testing and the remaining for training. Thus, the number of claims for testing varies for each county regressor, depending on the hurricane employed. Figure~\ref{fig:county-hurricanes}(b)  shows the maximum, minimum and average number of claims used for testing. We observe a surge of claims in Orlean Parish (22071) and  Harris (48201) counties originating from Katrina (2005) and Harvey (2017) (see Figure~\ref{fig:county-hurricanes}(c)). Considering the number of claims, the average $R^2$ after bias correction is 0.807 (see Figure~\ref{fig:county-hurricanes}(d)).

\begin{table}[t]
    \centering
    \caption[Feature Selection]{Ten most important features by county, as estimated by XGB. The value represents the fractional gain of a feature concerning the total gain of the feature's split~\citep{chen2016xgboost}. Acronyms stand for  	BFE (baseFloodElevation), 	
	ED	(elevationDifference),	
	PC	(policyCount),	
	TCI	(totalContentsInsuranceCoverage),	
	LAG	(lowestAdjacentGrade),	
	LFE	(lowestFloorElevation),	
	LAT	(latitude),	
	MO	(months),	
	LON	(longitude),	
	CI	(condominiumIndicator),	
	RZ	(reportedZipcode),	
	RM	(rateMethod),	
	OT	(occupancyType),	
	BE	(basementEnclosureCrawlspace),	
	ECI	(elevationCertificateIndicator),	
	EBI	(elevatedBuildingIndicator),	
	CRS	(communityRatingSystemDiscount),	
	NFB	(numberOfFloorsInTheInsuredBuilding),	
	OT	(obstructionType).	
 }
    \label{tb:importance}
\begin{footnotesize}
\begin{tabular}{rccccccccccc}
  \hline
 & {\bf county } &  {\bf 1} & {\bf 2} & {\bf 3} & {\bf 4} & {\bf 5} & {\bf 6} & {\bf 7} & {\bf 8} & {\bf 9} & {\bf 10} \\
  \hline
1 & 12011 & BFE & TBI & ED & PC & LAG & MO & TCI & RZ.33009 & LFE & PR.1 \\
   &   & 1.00 & 0.90 & 0.49 & 0.41 & 0.34 & 0.29 & 0.21 & 0.19 & 0.16 & 0.14 \\

  2 & 12086 & TBI & BFE & PC & MO & LFE & ED & LON & LAG & TCI & FZ.A14 \\
  &   & 1.00 & 0.92 & 0.59 & 0.39 & 0.37 & 0.29 & 0.16 & 0.16 & 0.13 & 0.13 \\

  3 & 17031 & BFE & TCI & MO & CI.A & TBI & OT.4 & LAT & LAG & ECI.1 & ED \\

   &   & 1.00 & 0.13 & 0.10 & 0.08 & 0.07 & 0.05 & 0.03 & 0.03 & 0.02 & 0.02 \\
  4 & 22051 & TBI & BFE & MO & LFE & TCI & RZ.70006 & LAG & ED & PC & LAT \\
   &   & 1.00 & 0.94 & 0.68 & 0.28 & 0.17 & 0.14 & 0.11 & 0.10 & 0.10 & 0.09 \\


  5 & 22071 & BFE & TBI & MO & ED & PC & TCI & LAG & LFE & BE.4 & LON \\

   &   & 1.00 & 0.89 & 0.57 & 0.27 & 0.27 & 0.25 & 0.21 & 0.14 & 0.03 & 0.03 \\
  6 & 22103 & MO & TBI & BFE & LAT & LAG & TCI & LFE & ED & LON & RZ.70461 \\
   &   & 1.00 & 0.93 & 0.82 & 0.40 & 0.37 & 0.35 & 0.31 & 0.31 & 0.07 & 0.07 \\


  7 & 34029 & BFE & LAG & MO & TBI & LFE & ED & TCI & LAT & PC & CRS.5 \\

   &   & 1.00 & 0.40 & 0.38 & 0.37 & 0.36 & 0.34 & 0.27 & 0.20 & 0.13 & 0.07 \\
  8 & 36059 & PC & RM.5 & BFE & TBI & LAG & MO & ED & TCI & LFE & NFB.6 \\
   &   & 1.00 & 0.76 & 0.48 & 0.39 & 0.21 & 0.16 & 0.10 & 0.04 & 0.03 & 0.03 \\

  9 & 36103 & BFE & MO & LFE & TBI & TCI & ED & LAG & LAT & LON & RM.1 \\

   &   & 1.00 & 0.57 & 0.43 & 0.37 & 0.20 & 0.19 & 0.18 & 0.10 & 0.09 & 0.06 \\
  10 & 48167 & TBI & BFE & TCI & MO & PC & LAG & ED & LFE & EBI.1 & LON \\
   &   & 1.00 & 0.44 & 0.33 & 0.31 & 0.25 & 0.22 & 0.20 & 0.19 & 0.14 & 0.12 \\

  11 & 48201 & BFE & TBI & MO & LAG & PC & TCI & ED & LFE & LAT & LON \\

   &   & 1.00 & 0.55 & 0.35 & 0.23 & 0.18 & 0.15 & 0.06 & 0.05 & 0.04 & 0.03 \\
   \hline
\end{tabular}
\end{footnotesize}
\end{table}

\subsection{Feature Importance}

An important by-product of regressors such as XGB is their ability to rank importance for the employed variables. Table~\ref{tb:importance} depicts the results for the counties under analysis. Most counties' predictors, such as BFE (base flood elevation), rank high, but it is only sometimes the most important. It stands out that each county has a particular set of essential features with different relative importance. Also, most of the highest-ranked features are continuous, but some discrete features, such as RM (rate method) with value 5, are the second most important feature for the county with code 36103 (Suffolk, NY).

\section{Discussion}
As the effects of climate change spread across the globe~\citep{liu2021half,hu2021using}, the occurrence of more frequent and extreme flood events is anticipated, which will cause a significant economic~\citep{basnayake2021assessing} set back in urban~\citep{mohor2021residential,nofal2021modeling} and rural~\citep{mohammadi2021flood} settings. Thus, there is a need for novel models to estimate flood risk~\citep{chen2021applying} and estimate loss~\citep{maiwald2021new}.
This research sheds light on a data-driven estimation of flood loss in the US, where a considerable data-gathering infrastructure is readily available, as opposed to other nations where this effort still needs to take place~\citep{lv2021construction}.

This research on the NFIP dataset~\citep{fema_nfips} shows that it is possible to construct regressors to infer the amount paid on building claims. This work adds to the current body of research on NFIP, enriching it to produce continuous probability for flooding~\citep{zarekarizi2021flood}, predict the number of claims~\citep{yang2021predicting} and in general assess risk~\citep{lin2021hurricane}. The regression results show significant progress over the employment of flood depth and damage curves~\citep{lin2021hurricane}, while the bias correction stage improves the overall estimation of the amount paid distribution, extending Chen {\it et al.}\cred{~\citeyearpar{chen2021applying}} analysis on risk models.

The assessment of classic decision tree, kernel, and neural networks-based nonlinear regression methods highlights the potential of \cred{ML} techniques and is similar to current efforts aimed to assess susceptibility~\citep{wang2021flood,saha2021flood,janizadeh2021novel,siam2021study,liu2021half}, estimate the spatial extent~\citep{lin2021spatial,lee2021scenario}, compare flood regions~\citep{persiano2021comparison}.
NFIP is a challenging dataset offering ample opportunities for ML techniques, {\it e.g.}, consider the uncertainty quantification in the labeling process for predictors~\citep{siam2021effects}.

An emerging argument is that the already-changed climate renders the extended historical record ineffective for modeling future risk. To make matters worse, since extremes are rare, a short history is also insufficient. Thus, the argument goes, we may no longer rely on the historical record to assess risk, even in the near future! Figures~\ref{fig:shifting_window}~\ref{fig:expanding_window} show models with considerable skill at projecting interannual flood loss risk using the historical record. We posit that this is primarily because they adapt in a dynamic data-driven manner. The model adapts faster as the horizon of future interest shrinks. Further, performance is similar between the expanding and shifting window modes of adaptation. This empirical evidence implies that it is neither true that the record is too short for the present flood loss modeling problem nor that using the entire history per se makes risk assessment less skillful. We posit that accelerating climate change at longer horizons will require better models through physics, but the issue remains unsettled at short horizons.

\section*{Conclusion}
The evaluation of flood risk is complex due to its multidimensionality characteristics that touch social, economic,  political, territorial, and scientific dimensions~\citep{elliott2021underwater}. Nevertheless, the quantitative assessment of the amount paid on insurance claims offers a glimpse of the challenges ahead and could serve as a solid guide for decision-making. This research explores the censored NFIP dataset by studying diverse classic ML techniques covering kernel, neural network, and decision tree-based approach. For a sample of some counties with the most significant number of claims, we demonstrate that it is possible to construct regressors that offer a critical predictive ability beyond the conventional approach of water depth and damage risk maps. Furthermore, we showed that incorporating meteorological variables can enhance performance. We also demonstrate that feature importance varies by county, which justifies the need for regional analysis.

We will incorporate relevant predictors into the regressor mix, as the related literature has highlighted their relevance. These include digital elevation maps, slopes, and climatological variables. Furthermore, current advances in ML techniques, such as oblique decision trees~\citep{carreira2021counterfactual}, are interesting to study in the light of the NFIP dataset.

\section*{Data Availability Statement}
The data and code generated or used during the study are available upon request.

\section*{Acknowledgements}

This work was funded in part by Liberty Mutual (029024-00020), ONR (N00014-19-1-2273), and the two MIT Climate Grand Challenge projects, namely, ``Preparing for a New World of Weather and Climate Extremes" and ``Reinventing Climate Change Adaptation with the Climate Resilience Early Warning System (CREWSnet)." SIP-IPN partly supported Joaquin Salas under grant 20220583 and by SECTEI CDMX under grant 910C21. The article’s content is solely the responsibility of the authors and does not necessarily represent the official views of the sponsors and funding sources.
The authors thank Dagoberto Pulido for providing routines for inferring the missing values and Goran Zivanovic for data analysis assistance.



\bibliographystyle{model2-names}
\bibliography{references}

\begin{thebibliography}{65}
\expandafter\ifx\csname natexlab\endcsname\relax\def\natexlab#1{#1}\fi
\providecommand{\url}[1]{\texttt{#1}}
\providecommand{\href}[2]{#2}
\providecommand{\path}[1]{#1}
\providecommand{\DOIprefix}{doi:}
\providecommand{\ArXivprefix}{arXiv:}
\providecommand{\URLprefix}{URL: }
\providecommand{\Pubmedprefix}{pmid:}
\providecommand{\doi}[1]{\href{http://dx.doi.org/#1}{\path{#1}}}
\providecommand{\Pubmed}[1]{\href{pmid:#1}{\path{#1}}}
\providecommand{\bibinfo}[2]{#2}
\ifx\xfnm\relax \def\xfnm[#1]{\unskip,\space#1}\fi
\bibitem[{Basha et~al.(2008)Basha, Ravela and Rus}]{basha10}
\bibinfo{author}{Basha, E.A.}, \bibinfo{author}{Ravela, S.},
  \bibinfo{author}{Rus, D.}, \bibinfo{year}{2008}.
\newblock \bibinfo{title}{Model-based monitoring for early warning flood
  detection}, in: \bibinfo{booktitle}{Proceedings of the 6th ACM Conference on
  Embedded Network Sensor Systems}, \bibinfo{publisher}{Association for
  Computing Machinery}, \bibinfo{address}{New York, NY, USA}. p.
  \bibinfo{pages}{295–308}.
\bibitem[{Basnayake et~al.(2021)Basnayake, Ulubasoglu, Rahman, Premalal,
  Chandrapala, Shrestha, Jayasinghe and Gupta}]{basnayake2021assessing}
\bibinfo{author}{Basnayake, S.}, \bibinfo{author}{Ulubasoglu, M.},
  \bibinfo{author}{Rahman, M.H.}, \bibinfo{author}{Premalal, S.},
  \bibinfo{author}{Chandrapala, L.}, \bibinfo{author}{Shrestha, M.L.},
  \bibinfo{author}{Jayasinghe, S.}, \bibinfo{author}{Gupta, N.},
  \bibinfo{year}{2021}.
\newblock \bibinfo{title}{{Assessing potential loss and damage for flood hazard
  using an econometric modelling technique}}.
\newblock \bibinfo{journal}{APN Science Bulletin} .
\bibitem[{Blasch et~al.(2018)Blasch, Ravela and Aved}]{dddasbook}
\bibinfo{author}{Blasch, E.}, \bibinfo{author}{Ravela, S.},
  \bibinfo{author}{Aved, A.}, \bibinfo{year}{2018}.
\newblock \bibinfo{title}{Handbook of dynamic data driven applications
  systems}.
\newblock \bibinfo{journal}{Springer} , \bibinfo{pages}{750}.
\bibitem[{Carreira-Perpi{\~n}{\'a}n and
  Hada(2021)}]{carreira2021counterfactual}
\bibinfo{author}{Carreira-Perpi{\~n}{\'a}n, M.{\'A}.}, \bibinfo{author}{Hada,
  S.S.}, \bibinfo{year}{2021}.
\newblock \bibinfo{title}{{Counterfactual explanations for oblique decision
  trees: Exact, efficient algorithms}}, in: \bibinfo{booktitle}{AAAI Conference
  on Artificial Intelligence}, pp. \bibinfo{pages}{6903--6911}.
\bibitem[{Chen et~al.(2021a)Chen, Peng and Chen}]{chen2021applying}
\bibinfo{author}{Chen, P.H.}, \bibinfo{author}{Peng, T.T.},
  \bibinfo{author}{Chen, C.J.}, \bibinfo{year}{2021}a.
\newblock \bibinfo{title}{{Applying value at risk and riskiness models to
  analyze the flood loss of transportation construction projects in Taiwan}}.
\newblock \bibinfo{journal}{Journal of the Chinese Institute of Engineers}
  \bibinfo{volume}{44}, \bibinfo{pages}{237--242}.
\bibitem[{Chen and Guestrin(2016)}]{chen2016xgboost}
\bibinfo{author}{Chen, T.}, \bibinfo{author}{Guestrin, C.},
  \bibinfo{year}{2016}.
\newblock \bibinfo{title}{Xgboost: A scalable tree boosting system}, in:
  \bibinfo{booktitle}{ACM SIGKDD International Conference on Knowledge
  Discovery and Data Mining}, pp. \bibinfo{pages}{785--794}.
\bibitem[{Chen et~al.(2021b)Chen, Li and Chen}]{chen2021does}
\bibinfo{author}{Chen, Y.}, \bibinfo{author}{Li, J.}, \bibinfo{author}{Chen,
  A.}, \bibinfo{year}{2021}b.
\newblock \bibinfo{title}{Does high risk mean high loss: evidence from flood
  disaster in southern china}.
\newblock \bibinfo{journal}{Science of the total environment}
  \bibinfo{volume}{785}, \bibinfo{pages}{147127}.
\bibitem[{Chicco et~al.(2021)Chicco, Warrens and
  Jurman}]{chicco2021coefficient}
\bibinfo{author}{Chicco, D.}, \bibinfo{author}{Warrens, M.},
  \bibinfo{author}{Jurman, G.}, \bibinfo{year}{2021}.
\newblock \bibinfo{title}{{The coefficient of determination R-squared is more
  informative than SMAPE, MAE, MAPE, MSE and RMSE in regression analysis
  evaluation}}.
\newblock \bibinfo{journal}{PeerJ Computer Science} \bibinfo{volume}{7},
  \bibinfo{pages}{e623}.
\bibitem[{Choi(2021)}]{choi2021development}
\bibinfo{author}{Choi, H.I.}, \bibinfo{year}{2021}.
\newblock \bibinfo{title}{Development of flood damage regression models by
  rainfall identification reflecting landscape features in gangwon province,
  the republic of korea}.
\newblock \bibinfo{journal}{Land} \bibinfo{volume}{10}, \bibinfo{pages}{123}.
\bibitem[{Costache et~al.(2021)Costache, Arabameri, Moayedi, Pham, Santosh,
  Nguyen, Pandey and Pham}]{costache2021flash}
\bibinfo{author}{Costache, R.}, \bibinfo{author}{Arabameri, A.},
  \bibinfo{author}{Moayedi, H.}, \bibinfo{author}{Pham, Q.B.},
  \bibinfo{author}{Santosh, M.}, \bibinfo{author}{Nguyen, H.},
  \bibinfo{author}{Pandey, M.}, \bibinfo{author}{Pham, B.T.},
  \bibinfo{year}{2021}.
\newblock \bibinfo{title}{{Flash-flood potential index estimation using fuzzy
  logic combined with deep learning neural network, naive Bayes, XGBoost and
  classification and regression tree}}.
\newblock \bibinfo{journal}{Geocarto International} , \bibinfo{pages}{1--28}.
\bibitem[{{CRED}(2018)}]{CRED2018}
\bibinfo{author}{{CRED}}, \bibinfo{year}{2018}.
\newblock \bibinfo{title}{{Economic Losses, Poverty and Disasters, 1998-2017}}.
\newblock \URLprefix \url{https://www.unisdr.org}.
\bibitem[{Dombrowski et~al.(2021)Dombrowski, Ratnadiwakara and
  Slawson~Jr}]{dombrowski2021fima}
\bibinfo{author}{Dombrowski, T.}, \bibinfo{author}{Ratnadiwakara, D.},
  \bibinfo{author}{Slawson~Jr, V.C.}, \bibinfo{year}{2021}.
\newblock \bibinfo{title}{{The FIMA NFIP’s Redacted Policies and Redacted
  Claims Datasets}}.
\newblock \bibinfo{journal}{Journal of Real Estate Literature}
  \bibinfo{volume}{28}, \bibinfo{pages}{190--212}.
\bibitem[{Elliott(2021)}]{elliott2021underwater}
\bibinfo{author}{Elliott, R.}, \bibinfo{year}{2021}.
\newblock \bibinfo{title}{{Underwater: Loss, flood insurance, and the moral
  economy of climate change in the United States}}.
\newblock \bibinfo{publisher}{Columbia University Press}.
\bibitem[{Emanuel(2005)}]{emanuel2005divine}
\bibinfo{author}{Emanuel, K.}, \bibinfo{year}{2005}.
\newblock \bibinfo{title}{{Divine wind: the history and science of
  hurricanes}}.
\newblock \bibinfo{publisher}{Oxford University press}.
\bibitem[{Emanuel and Ravela(2012)}]{emanuel_ravela_2012}
\bibinfo{author}{Emanuel, K.}, \bibinfo{author}{Ravela, S.},
  \bibinfo{year}{2012}.
\newblock \bibinfo{title}{{Synthetic Storm Simulation for Wind Risk
  Assessment}}. chapter~\bibinfo{chapter}{3}.
\newblock pp. \bibinfo{pages}{15--36}.
\bibitem[{Emanuel et~al.(2006)Emanuel, Ravela, Vivant and
  Risi}]{emanuel_ravela_2006}
\bibinfo{author}{Emanuel, K.}, \bibinfo{author}{Ravela, S.},
  \bibinfo{author}{Vivant, E.}, \bibinfo{author}{Risi, C.},
  \bibinfo{year}{2006}.
\newblock \bibinfo{title}{{A Statistical Deterministic Approach to Hurricane
  Risk Assessment}}.
\newblock \bibinfo{journal}{Bulletin of the American Meteorological Society}
  \bibinfo{volume}{87}, \bibinfo{pages}{299--314}.
\bibitem[{{FEMA}(2019)}]{fema_nfips}
\bibinfo{author}{{FEMA}}, \bibinfo{year}{2019}.
\newblock \bibinfo{title}{{Federal insurance and mitigation administration
  national flood insurance program (FIMA NFIP) redacted claims dataset}}.
\newblock \URLprefix
  \url{https://www.fema.gov/about/reports-and-data/openfema}.
\bibitem[{FEMA(2022)}]{hazus}
\bibinfo{author}{FEMA}, \bibinfo{year}{2022}.
\newblock \bibinfo{title}{Hazus flood technical manual}.
\newblock \bibinfo{journal}{Hazus} \bibinfo{volume}{5.1}, \bibinfo{pages}{110}.
\bibitem[{Frame et~al.(2020)}]{frame2020climate}
\bibinfo{author}{Frame, D.}, et~al., \bibinfo{year}{2020}.
\newblock \bibinfo{title}{{Climate change attribution and the economic costs of
  extreme weather events: a study on damages from extreme rainfall and
  drought}}.
\newblock \bibinfo{journal}{Climatic Change} \bibinfo{volume}{162},
  \bibinfo{pages}{781--797}.
\bibitem[{Friedman(2001)}]{friedman2001greedy}
\bibinfo{author}{Friedman, J.}, \bibinfo{year}{2001}.
\newblock \bibinfo{title}{{Greedy function approximation: a gradient boosting
  machine}}.
\newblock \bibinfo{journal}{Annals of statistics} ,
  \bibinfo{pages}{1189--1232}.
\bibitem[{Ghojogh et~al.(2021)Ghojogh, Ghodsi, Karray and
  Crowley}]{ghojogh2021reproducing}
\bibinfo{author}{Ghojogh, B.}, \bibinfo{author}{Ghodsi, A.},
  \bibinfo{author}{Karray, F.}, \bibinfo{author}{Crowley, M.},
  \bibinfo{year}{2021}.
\newblock \bibinfo{title}{Reproducing kernel hilbert space, mercer's theorem,
  eigenfunctions, nystr$\backslash$" om method, and use of kernels in machine
  learning: Tutorial and survey}.
\newblock \bibinfo{journal}{arXiv preprint arXiv:2106.08443} .
\bibitem[{Gil et~al.(2018)Gil, Pierce, Babaie, Banerjee, Borne, Bust, Cheatham,
  Ebert-Uphoff, Gomes, Hill, Horel, Hsu, Kinter, Knoblock, Krum, Kumar,
  Lermusiaux, Liu, North, Pankratius, Peters, Plale, Pope, Ravela, Restrepo,
  Ridley, Samet, Shekhar, Skinner, Smyth, Tikoff, Yarmey and
  Zhang}]{gil_pierce_19}
\bibinfo{author}{Gil, Y.}, \bibinfo{author}{Pierce, S.A.},
  \bibinfo{author}{Babaie, H.}, \bibinfo{author}{Banerjee, A.},
  \bibinfo{author}{Borne, K.}, \bibinfo{author}{Bust, G.},
  \bibinfo{author}{Cheatham, M.}, \bibinfo{author}{Ebert-Uphoff, I.},
  \bibinfo{author}{Gomes, C.}, \bibinfo{author}{Hill, M.},
  \bibinfo{author}{Horel, J.}, \bibinfo{author}{Hsu, L.},
  \bibinfo{author}{Kinter, J.}, \bibinfo{author}{Knoblock, C.},
  \bibinfo{author}{Krum, D.}, \bibinfo{author}{Kumar, V.},
  \bibinfo{author}{Lermusiaux, P.}, \bibinfo{author}{Liu, Y.},
  \bibinfo{author}{North, C.}, \bibinfo{author}{Pankratius, V.},
  \bibinfo{author}{Peters, S.}, \bibinfo{author}{Plale, B.},
  \bibinfo{author}{Pope, A.}, \bibinfo{author}{Ravela, S.},
  \bibinfo{author}{Restrepo, J.}, \bibinfo{author}{Ridley, A.},
  \bibinfo{author}{Samet, H.}, \bibinfo{author}{Shekhar, S.},
  \bibinfo{author}{Skinner, K.}, \bibinfo{author}{Smyth, P.},
  \bibinfo{author}{Tikoff, B.}, \bibinfo{author}{Yarmey, L.},
  \bibinfo{author}{Zhang, J.}, \bibinfo{year}{2018}.
\newblock \bibinfo{title}{Intelligent systems for geosciences: An essential
  research agenda}.
\newblock \bibinfo{journal}{Commun. ACM} \bibinfo{volume}{62},
  \bibinfo{pages}{76–84}.
\bibitem[{Goodfellow et~al.(2020)Goodfellow, Pouget-Abadie, Mirza, Xu,
  Warde-Farley, Ozair, Courville and Bengio}]{goodfellow2020generative}
\bibinfo{author}{Goodfellow, I.}, \bibinfo{author}{Pouget-Abadie, J.},
  \bibinfo{author}{Mirza, M.}, \bibinfo{author}{Xu, B.},
  \bibinfo{author}{Warde-Farley, D.}, \bibinfo{author}{Ozair, S.},
  \bibinfo{author}{Courville, A.}, \bibinfo{author}{Bengio, Y.},
  \bibinfo{year}{2020}.
\newblock \bibinfo{title}{{Generative adversarial networks}}.
\newblock \bibinfo{journal}{Communications of the ACM} \bibinfo{volume}{63},
  \bibinfo{pages}{139--144}.
\bibitem[{Guttman(1946)}]{guttman1946enlargement}
\bibinfo{author}{Guttman, L.}, \bibinfo{year}{1946}.
\newblock \bibinfo{title}{{Enlargement methods for computing the inverse
  matrix}}.
\newblock \bibinfo{journal}{The annals of mathematical statistics} ,
  \bibinfo{pages}{336--343}.
\bibitem[{Hallegatte et~al.(2016)Hallegatte, Vogt-Schilb, Bangalore and
  Rozenberg}]{hallegatte2016unbreakable}
\bibinfo{author}{Hallegatte, S.}, \bibinfo{author}{Vogt-Schilb, A.},
  \bibinfo{author}{Bangalore, M.}, \bibinfo{author}{Rozenberg, J.},
  \bibinfo{year}{2016}.
\newblock \bibinfo{title}{{Unbreakable: building the resilience of the poor in
  the face of natural disasters}}.
\newblock \bibinfo{publisher}{World Bank Publications}.
\bibitem[{Hu et~al.(2021)Hu, Wang, Liu, Gong and Kantz}]{hu2021using}
\bibinfo{author}{Hu, X.}, \bibinfo{author}{Wang, M.}, \bibinfo{author}{Liu,
  K.}, \bibinfo{author}{Gong, D.}, \bibinfo{author}{Kantz, H.},
  \bibinfo{year}{2021}.
\newblock \bibinfo{title}{Using climate factors to estimate flood economic loss
  risk}.
\newblock \bibinfo{journal}{International Journal of Disaster Risk Science}
  \bibinfo{volume}{12}, \bibinfo{pages}{731--744}.
\bibitem[{IPCC(2021)}]{IPCC2021climate}
\bibinfo{author}{IPCC}, \bibinfo{year}{2021}.
\newblock \bibinfo{title}{{Climate change 2021: The Physical Science Basis}}.
\newblock \bibinfo{publisher}{United Nations}.
\bibitem[{Janizadeh et~al.(2021)Janizadeh, Vafakhah, Kapelan and
  Dinan}]{janizadeh2021novel}
\bibinfo{author}{Janizadeh, S.}, \bibinfo{author}{Vafakhah, M.},
  \bibinfo{author}{Kapelan, Z.}, \bibinfo{author}{Dinan, N.M.},
  \bibinfo{year}{2021}.
\newblock \bibinfo{title}{{Novel Bayesian Additive Regression Tree Methodology
  for Flood Susceptibility Modeling}}.
\newblock \bibinfo{journal}{Water Resources Management} \bibinfo{volume}{35},
  \bibinfo{pages}{4621--4646}.
\bibitem[{Karpatne et~al.(2019)Karpatne, Ebert-Uphoff, Ravela, Babaie and
  Kumar}]{karpatne19}
\bibinfo{author}{Karpatne, A.}, \bibinfo{author}{Ebert-Uphoff, I.},
  \bibinfo{author}{Ravela, S.}, \bibinfo{author}{Babaie, H.A.},
  \bibinfo{author}{Kumar, V.}, \bibinfo{year}{2019}.
\newblock \bibinfo{title}{Machine learning for the geosciences: Challenges and
  opportunities}.
\newblock \bibinfo{journal}{IEEE Transactions on Knowledge and Data
  Engineering} \bibinfo{volume}{31}, \bibinfo{pages}{1544--1554}.
\newblock \DOIprefix\doi{10.1109/TKDE.2018.2861006}.
\bibitem[{Lea and Pralle(2022)}]{lea2022appeal}
\bibinfo{author}{Lea, D.}, \bibinfo{author}{Pralle, S.}, \bibinfo{year}{2022}.
\newblock \bibinfo{title}{{To appeal and amend: Changes to recently updated
  Flood Insurance Rate Maps}}.
\newblock \bibinfo{journal}{Risk, Hazards \& Crisis in Public Policy}
  \bibinfo{volume}{13}, \bibinfo{pages}{28--47}.
\bibitem[{Lee and Kim(2021)}]{lee2021scenario}
\bibinfo{author}{Lee, J.}, \bibinfo{author}{Kim, B.}, \bibinfo{year}{2021}.
\newblock \bibinfo{title}{Scenario-based real-time flood prediction with
  logistic regression}.
\newblock \bibinfo{journal}{Water} \bibinfo{volume}{13}, \bibinfo{pages}{1191}.
\bibitem[{Lin and Cha(2021)}]{lin2021hurricane}
\bibinfo{author}{Lin, C.Y.}, \bibinfo{author}{Cha, E.J.}, \bibinfo{year}{2021}.
\newblock \bibinfo{title}{{Hurricane Freshwater Flood Risk Assessment Model for
  Residential Buildings in Southeast US Coastal States Considering Climate
  Change}}.
\newblock \bibinfo{journal}{Natural Hazards Review} \bibinfo{volume}{22},
  \bibinfo{pages}{04020061}.
\bibitem[{Lin and Billa(2021)}]{lin2021spatial}
\bibinfo{author}{Lin, J.M.}, \bibinfo{author}{Billa, L.}, \bibinfo{year}{2021}.
\newblock \bibinfo{title}{{Spatial prediction of flood-prone areas using
  geographically weighted regression}}.
\newblock \bibinfo{journal}{Environmental Advances} \bibinfo{volume}{6},
  \bibinfo{pages}{100118}.
\bibitem[{Liu et~al.(2021)Liu, Gao and Wu}]{liu2021half}
\bibinfo{author}{Liu, L.}, \bibinfo{author}{Gao, J.}, \bibinfo{author}{Wu, S.},
  \bibinfo{year}{2021}.
\newblock \bibinfo{title}{{Half a degree warming might cause doubled economic
  loss and intensified affected population of flood in China}}.
\newblock \bibinfo{journal}{Natural Hazards and Earth System Sciences
  Discussions} , \bibinfo{pages}{1--19}.
\bibitem[{Lv et~al.(2021)Lv, Meng, Wu, Guan and Liu}]{lv2021construction}
\bibinfo{author}{Lv, H.}, \bibinfo{author}{Meng, Y.}, \bibinfo{author}{Wu, Z.},
  \bibinfo{author}{Guan, X.}, \bibinfo{author}{Liu, Y.}, \bibinfo{year}{2021}.
\newblock \bibinfo{title}{{Construction of flood loss function for cities
  lacking disaster data based on three-dimensional (object-function-array) data
  processing}}.
\newblock \bibinfo{journal}{Science of the total environment}
  \bibinfo{volume}{773}, \bibinfo{pages}{145649}.
\bibitem[{Maiwald et~al.(2021)Maiwald, Kaufmann, Langhammer and
  Schwarz}]{maiwald2021new}
\bibinfo{author}{Maiwald, H.}, \bibinfo{author}{Kaufmann, C.},
  \bibinfo{author}{Langhammer, T.}, \bibinfo{author}{Schwarz, J.},
  \bibinfo{year}{2021}.
\newblock \bibinfo{title}{{A new model for consideration of flow velocity in
  flood damage and loss prognosis}}, in: \bibinfo{booktitle}{European
  Conference on Flood Risk Management}, \bibinfo{organization}{Budapest
  University of Technology and Economics}. pp. \bibinfo{pages}{1--12}.
\bibitem[{Malan et~al.(2020)Malan, Smuts, Baumgartner and
  Ricci}]{malan2020missing}
\bibinfo{author}{Malan, L.}, \bibinfo{author}{Smuts, C.M.},
  \bibinfo{author}{Baumgartner, J.}, \bibinfo{author}{Ricci, C.},
  \bibinfo{year}{2020}.
\newblock \bibinfo{title}{Missing data imputation via the
  expectation-maximization algorithm can improve principal component analysis
  aimed at deriving biomarker profiles and dietary patterns}.
\newblock \bibinfo{journal}{Nutrition Research} \bibinfo{volume}{75},
  \bibinfo{pages}{67--76}.
\bibitem[{Mobley et~al.(2021)Mobley, Sebastian, Blessing, Highfield, Stearns
  and Brody}]{mobley2021quantification}
\bibinfo{author}{Mobley, W.}, \bibinfo{author}{Sebastian, A.},
  \bibinfo{author}{Blessing, R.}, \bibinfo{author}{Highfield, W.E.},
  \bibinfo{author}{Stearns, L.}, \bibinfo{author}{Brody, S.D.},
  \bibinfo{year}{2021}.
\newblock \bibinfo{title}{{Quantification of continuous flood hazard using
  random forest classification and flood insurance claims at large spatial
  scales: a pilot study in southeast Texas}}.
\newblock \bibinfo{journal}{Natural Hazards and Earth System Sciences}
  \bibinfo{volume}{21}, \bibinfo{pages}{807--822}.
\bibitem[{Mohammadi et~al.(2021)Mohammadi, Darabi, Mirchooli, Bakhshaee and
  Haghighi}]{mohammadi2021flood}
\bibinfo{author}{Mohammadi, M.}, \bibinfo{author}{Darabi, H.},
  \bibinfo{author}{Mirchooli, F.}, \bibinfo{author}{Bakhshaee, A.},
  \bibinfo{author}{Haghighi, A.T.}, \bibinfo{year}{2021}.
\newblock \bibinfo{title}{{Flood risk mapping and crop-water loss modeling
  using water footprint analysis in agricultural watershed, northern Iran}}.
\newblock \bibinfo{journal}{Natural Hazards} \bibinfo{volume}{105},
  \bibinfo{pages}{2007--2025}.
\bibitem[{Mohor et~al.(2021)Mohor, Thieken and Korup}]{mohor2021residential}
\bibinfo{author}{Mohor, G.S.}, \bibinfo{author}{Thieken, A.H.},
  \bibinfo{author}{Korup, O.}, \bibinfo{year}{2021}.
\newblock \bibinfo{title}{{Residential flood loss estimated from Bayesian
  multilevel models}}.
\newblock \bibinfo{journal}{Natural Hazards and Earth System Sciences}
  \bibinfo{volume}{21}, \bibinfo{pages}{1599--1614}.
\bibitem[{Moon(1996)}]{moon1996expectation}
\bibinfo{author}{Moon, T.K.}, \bibinfo{year}{1996}.
\newblock \bibinfo{title}{The expectation-maximization algorithm}.
\newblock \bibinfo{journal}{IEEE Signal processing magazine}
  \bibinfo{volume}{13}, \bibinfo{pages}{47--60}.
\bibitem[{Neumann et~al.(2015a)Neumann, Emanuel, Ravela, Ludwig, Kirshen, Bosma
  and Martinich}]{neumann2015joint}
\bibinfo{author}{Neumann, J.E.}, \bibinfo{author}{Emanuel, K.},
  \bibinfo{author}{Ravela, S.}, \bibinfo{author}{Ludwig, L.},
  \bibinfo{author}{Kirshen, P.}, \bibinfo{author}{Bosma, K.},
  \bibinfo{author}{Martinich, J.}, \bibinfo{year}{2015}a.
\newblock \bibinfo{title}{Joint effects of storm surge and sea-level rise on us
  coasts: new economic estimates of impacts, adaptation, and benefits of
  mitigation policy}.
\newblock \bibinfo{journal}{Climatic Change} \bibinfo{volume}{129},
  \bibinfo{pages}{337--349}.
\bibitem[{Neumann et~al.(2015b)Neumann, Emanuel, Ravela, Ludwig and
  Verly}]{neumann2015risks}
\bibinfo{author}{Neumann, J.E.}, \bibinfo{author}{Emanuel, K.A.},
  \bibinfo{author}{Ravela, S.}, \bibinfo{author}{Ludwig, L.C.},
  \bibinfo{author}{Verly, C.}, \bibinfo{year}{2015}b.
\newblock \bibinfo{title}{Risks of coastal storm surge and the effect of sea
  level rise in the red river delta, vietnam}.
\newblock \bibinfo{journal}{Sustainability} \bibinfo{volume}{7},
  \bibinfo{pages}{6553--6572}.
\bibitem[{Nofal et~al.(2021)Nofal, van~de Lindt, Cutler, Shields and
  Crofton}]{nofal2021modeling}
\bibinfo{author}{Nofal, O.M.}, \bibinfo{author}{van~de Lindt, J.W.},
  \bibinfo{author}{Cutler, H.}, \bibinfo{author}{Shields, M.},
  \bibinfo{author}{Crofton, K.}, \bibinfo{year}{2021}.
\newblock \bibinfo{title}{{Modeling the Impact of Building-Level Flood
  Mitigation Measures Made Possible by Early Flood Warnings on Community-Level
  Flood Loss Reduction}}.
\newblock \bibinfo{journal}{Buildings} \bibinfo{volume}{11},
  \bibinfo{pages}{475}.
\bibitem[{Parizi et~al.(2021)Parizi, Bagheri-Gavkosh, Hosseini and
  Geravand}]{parizi2021linkage}
\bibinfo{author}{Parizi, E.}, \bibinfo{author}{Bagheri-Gavkosh, M.},
  \bibinfo{author}{Hosseini, S.M.}, \bibinfo{author}{Geravand, F.},
  \bibinfo{year}{2021}.
\newblock \bibinfo{title}{{Linkage of geographically weighted regression with
  spatial cluster analyses for regionalization of flood peak discharges
  drivers: Case studies across Iran}}.
\newblock \bibinfo{journal}{Journal of Cleaner Production}
  \bibinfo{volume}{310}, \bibinfo{pages}{127526}.
\bibitem[{Patro et~al.(2009)Patro, Chatterjee, Mohanty, Singh and
  Raghuwanshi}]{patro2009flood}
\bibinfo{author}{Patro, S.}, \bibinfo{author}{Chatterjee, C.},
  \bibinfo{author}{Mohanty, S.}, \bibinfo{author}{Singh, R.},
  \bibinfo{author}{Raghuwanshi, N.}, \bibinfo{year}{2009}.
\newblock \bibinfo{title}{{Flood inundation modeling using MIKE FLOOD and
  remote sensing data}}.
\newblock \bibinfo{journal}{Journal of the Indian Society of Remote Sensing}
  \bibinfo{volume}{37}, \bibinfo{pages}{107--118}.
\bibitem[{Perkins and Enos(1968)}]{perkins1968hurricane}
\bibinfo{author}{Perkins, R.}, \bibinfo{author}{Enos, P.},
  \bibinfo{year}{1968}.
\newblock \bibinfo{title}{{Hurricane Betsy in the Florida-Bahama area: geologic
  effects and comparison with Hurricane Donna}}.
\newblock \bibinfo{journal}{The Journal of Geology} \bibinfo{volume}{76},
  \bibinfo{pages}{710--717}.
\bibitem[{Persiano et~al.(2021)Persiano, Salinas, Stedinger, Farmer, Lun,
  Viglione, Bl{\"o}schl and Castellarin}]{persiano2021comparison}
\bibinfo{author}{Persiano, S.}, \bibinfo{author}{Salinas, J.L.},
  \bibinfo{author}{Stedinger, J.R.}, \bibinfo{author}{Farmer, W.H.},
  \bibinfo{author}{Lun, D.}, \bibinfo{author}{Viglione, A.},
  \bibinfo{author}{Bl{\"o}schl, G.}, \bibinfo{author}{Castellarin, A.},
  \bibinfo{year}{2021}.
\newblock \bibinfo{title}{A comparison between generalized least squares
  regression and top-kriging for homogeneous cross-correlated flood regions}.
\newblock \bibinfo{journal}{Hydrological Sciences Journal}
  \bibinfo{volume}{66}, \bibinfo{pages}{565--579}.
\bibitem[{Prince(2012)}]{prince2012computer}
\bibinfo{author}{Prince, S.}, \bibinfo{year}{2012}.
\newblock \bibinfo{title}{{Computer vision: models, learning, and inference}}.
\newblock \bibinfo{publisher}{Cambridge University Press}.
\bibitem[{Ramasamy et~al.(2022)Ramasamy, Nagan and Kumar}]{ramasamy2022case}
\bibinfo{author}{Ramasamy, M.}, \bibinfo{author}{Nagan, S.},
  \bibinfo{author}{Kumar, P.S.}, \bibinfo{year}{2022}.
\newblock \bibinfo{title}{A case study of flood frequency analysis by
  intercomparison of graphical linear log-regression method and gumbel's
  analytical method in the vaigai river basin of tamil nadu, india}.
\newblock \bibinfo{journal}{Chemosphere} \bibinfo{volume}{286},
  \bibinfo{pages}{131571}.
\bibitem[{Ravela and Emanuel(2010)}]{ravela_emanuel_2010}
\bibinfo{author}{Ravela, S.}, \bibinfo{author}{Emanuel, K.},
  \bibinfo{year}{2010}.
\newblock \bibinfo{title}{Statistical-deterministic approach to natural
  disaster prediction}.
\newblock \bibinfo{journal}{United States Patent} .
\bibitem[{Rayner et~al.(2009)Rayner, Thas and Best}]{rayner2009smooth}
\bibinfo{author}{Rayner, J.C.}, \bibinfo{author}{Thas, O.},
  \bibinfo{author}{Best, D.J.}, \bibinfo{year}{2009}.
\newblock \bibinfo{title}{Smooth tests of goodness of fit: using R}.
\newblock \bibinfo{publisher}{John Wiley \& Sons}.
\bibitem[{Saha et~al.(2021)Saha, Pal, Arabameri, Blaschke, Panahi, Chowdhuri,
  Chakrabortty, Costache and Arora}]{saha2021flood}
\bibinfo{author}{Saha, A.}, \bibinfo{author}{Pal, S.C.},
  \bibinfo{author}{Arabameri, A.}, \bibinfo{author}{Blaschke, T.},
  \bibinfo{author}{Panahi, S.}, \bibinfo{author}{Chowdhuri, I.},
  \bibinfo{author}{Chakrabortty, R.}, \bibinfo{author}{Costache, R.},
  \bibinfo{author}{Arora, A.}, \bibinfo{year}{2021}.
\newblock \bibinfo{title}{Flood susceptibility assessment using novel ensemble
  of hyperpipes and support vector regression algorithms}.
\newblock \bibinfo{journal}{Water} \bibinfo{volume}{13}, \bibinfo{pages}{241}.
\bibitem[{Saha and Ravela(2022)}]{saha2022downscaling}
\bibinfo{author}{Saha, A.}, \bibinfo{author}{Ravela, S.}, \bibinfo{year}{2022}.
\newblock \bibinfo{title}{{Downscaling Extreme Rainfall Using
  Physical-Statistical Generative Adversarial Learning}}.
\newblock \bibinfo{journal}{arXiv preprint arXiv:2212.01446} .
\bibitem[{Scawthorn et~al.(2006)Scawthorn, Flores, Blais, Seligson, Tate,
  Chang, Mifflin, Thomas, Murphy, Jones et~al.}]{scawthorn2006hazus}
\bibinfo{author}{Scawthorn, C.}, \bibinfo{author}{Flores, P.},
  \bibinfo{author}{Blais, N.}, \bibinfo{author}{Seligson, H.},
  \bibinfo{author}{Tate, E.}, \bibinfo{author}{Chang, S.},
  \bibinfo{author}{Mifflin, E.}, \bibinfo{author}{Thomas, W.},
  \bibinfo{author}{Murphy, J.}, \bibinfo{author}{Jones, C.}, et~al.,
  \bibinfo{year}{2006}.
\newblock \bibinfo{title}{Hazus-mh flood loss estimation methodology. ii.
  damage and loss assessment}.
\newblock \bibinfo{journal}{Natural Hazards Review} \bibinfo{volume}{7},
  \bibinfo{pages}{72--81}.
\bibitem[{Schoppa et~al.(2021)Schoppa, Kreibich, Sieg, Vogel and
  Z{\"o}ller}]{schoppa2021developing}
\bibinfo{author}{Schoppa, L.}, \bibinfo{author}{Kreibich, H.},
  \bibinfo{author}{Sieg, T.}, \bibinfo{author}{Vogel, K.},
  \bibinfo{author}{Z{\"o}ller, G.}, \bibinfo{year}{2021}.
\newblock \bibinfo{title}{{Developing multivariable probabilistic flood loss
  models for companies}}, in: \bibinfo{booktitle}{European Conference on Flood
  Risk Management}, \bibinfo{organization}{Budapest University of Technology
  and Economics}. pp. \bibinfo{pages}{1--9}.
\bibitem[{Siam et~al.(2021a)Siam, Hasan, Anik, Noor, Adnan and
  Rahman}]{siam2021study}
\bibinfo{author}{Siam, Z.S.}, \bibinfo{author}{Hasan, R.T.},
  \bibinfo{author}{Anik, S.S.}, \bibinfo{author}{Noor, F.},
  \bibinfo{author}{Adnan, M.S.G.}, \bibinfo{author}{Rahman, R.M.},
  \bibinfo{year}{2021}a.
\newblock \bibinfo{title}{{Study of Hybridized Support Vector Regression Based
  Flood Susceptibility Mapping for Bangladesh}}, in:
  \bibinfo{booktitle}{International Conference on Industrial, Engineering and
  Other Applications of Applied Intelligent Systems},
  \bibinfo{organization}{Springer}. pp. \bibinfo{pages}{59--71}.
\bibitem[{Siam et~al.(2021b)Siam, Hasan and Rahman}]{siam2021effects}
\bibinfo{author}{Siam, Z.S.}, \bibinfo{author}{Hasan, R.T.},
  \bibinfo{author}{Rahman, R.M.}, \bibinfo{year}{2021}b.
\newblock \bibinfo{title}{{Effects of Label Noise on Regression Performances
  and Model Complexities for Hybridized Machine Learning Based Spatial Flood
  Susceptibility Modelling}}.
\newblock \bibinfo{journal}{Cybernetics and Systems} , \bibinfo{pages}{1--18}.
\bibitem[{Sun et~al.(2020)Sun, Bocchini and Davison}]{sun2020applications}
\bibinfo{author}{Sun, W.}, \bibinfo{author}{Bocchini, P.},
  \bibinfo{author}{Davison, B.D.}, \bibinfo{year}{2020}.
\newblock \bibinfo{title}{Applications of artificial intelligence for disaster
  management}.
\newblock \bibinfo{journal}{Natural Hazards} , \bibinfo{pages}{1--59}.
\bibitem[{Taghavi-Shahri et~al.(2020)Taghavi-Shahri, Fass{\`o}, Mahaki and
  Amini}]{taghavi2020concurrent}
\bibinfo{author}{Taghavi-Shahri, S.M.}, \bibinfo{author}{Fass{\`o}, A.},
  \bibinfo{author}{Mahaki, B.}, \bibinfo{author}{Amini, H.},
  \bibinfo{year}{2020}.
\newblock \bibinfo{title}{Concurrent spatiotemporal daily land use regression
  modeling and missing data imputation of fine particulate matter using
  distributed space-time expectation maximization}.
\newblock \bibinfo{journal}{Atmospheric Environment} \bibinfo{volume}{224},
  \bibinfo{pages}{117202}.
\bibitem[{Thornton et~al.(2020)Thornton, Shrestha, Wei, Thornton, Kao and
  Wilson}]{daymet2020}
\bibinfo{author}{Thornton, M.}, \bibinfo{author}{Shrestha, R.},
  \bibinfo{author}{Wei, Y.}, \bibinfo{author}{Thornton, P.},
  \bibinfo{author}{Kao, S.}, \bibinfo{author}{Wilson, B.},
  \bibinfo{year}{2020}.
\newblock \bibinfo{title}{Daymet: Daily surface weather data on a 1-km grid for
  north america, version 4}.
\newblock \URLprefix \url{https://daymet.ornl.gov},
  \DOIprefix\doi{10.3334/ORNLDAAC/1840}.
\bibitem[{{U.S. Bureau of Labor Statistics}(2021)}]{CPI2021}
\bibinfo{author}{{U.S. Bureau of Labor Statistics}}, \bibinfo{year}{2021}.
\newblock \bibinfo{title}{{Consumer Price Index}}.
\newblock \URLprefix \url{https://www.bls.gov/cpi/}.
\bibitem[{Wang et~al.(2021)Wang, Fang, Hong, Costache and Tang}]{wang2021flood}
\bibinfo{author}{Wang, Y.}, \bibinfo{author}{Fang, Z.}, \bibinfo{author}{Hong,
  H.}, \bibinfo{author}{Costache, R.}, \bibinfo{author}{Tang, X.},
  \bibinfo{year}{2021}.
\newblock \bibinfo{title}{{Flood susceptibility mapping by integrating
  frequency ratio and index of entropy with multilayer perceptron and
  classification and regression tree}}.
\newblock \bibinfo{journal}{Journal of Environmental Management}
  \bibinfo{volume}{289}, \bibinfo{pages}{112449}.
\bibitem[{Yang et~al.(2021)Yang, Shen, Yang, Anagnostou, He, Mo, Seyyedi,
  Kettner and Zhang}]{yang2021predicting}
\bibinfo{author}{Yang, Q.}, \bibinfo{author}{Shen, X.}, \bibinfo{author}{Yang,
  F.}, \bibinfo{author}{Anagnostou, E.N.}, \bibinfo{author}{He, K.},
  \bibinfo{author}{Mo, C.}, \bibinfo{author}{Seyyedi, H.},
  \bibinfo{author}{Kettner, A.J.}, \bibinfo{author}{Zhang, Q.},
  \bibinfo{year}{2021}.
\newblock \bibinfo{title}{{Predicting Flood Property Insurance Claims over
  CONUS, Fusing Big Earth Observation Data}}.
\newblock \bibinfo{journal}{Bulletin of the American Meteorological Society} .
\bibitem[{Zarekarizi et~al.(2021)Zarekarizi, Roop-Eckart, Sharma and
  Keller}]{zarekarizi2021flood}
\bibinfo{author}{Zarekarizi, M.}, \bibinfo{author}{Roop-Eckart, K.J.},
  \bibinfo{author}{Sharma, S.}, \bibinfo{author}{Keller, K.},
  \bibinfo{year}{2021}.
\newblock \bibinfo{title}{{The Flood Probability Interpolation Tool (FLOPIT): A
  Simple Tool to Improve Spatial Flood Probability Quantification and
  Communication}}.
\newblock \bibinfo{journal}{Water} \bibinfo{volume}{13}, \bibinfo{pages}{666}.

\end{thebibliography}
\newpage

\appendix
\section{The NFIP Dataset}
\begin{figure}
    \centering
    \begin{tabular}{cccc}
    \includegraphics[width=1.5in]{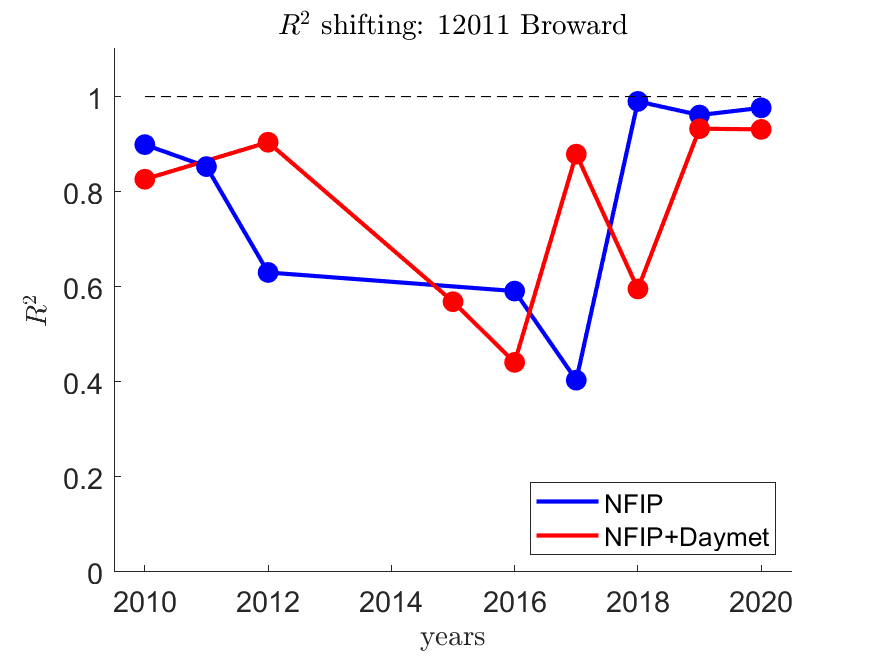}&
    \includegraphics[width=1.5in]{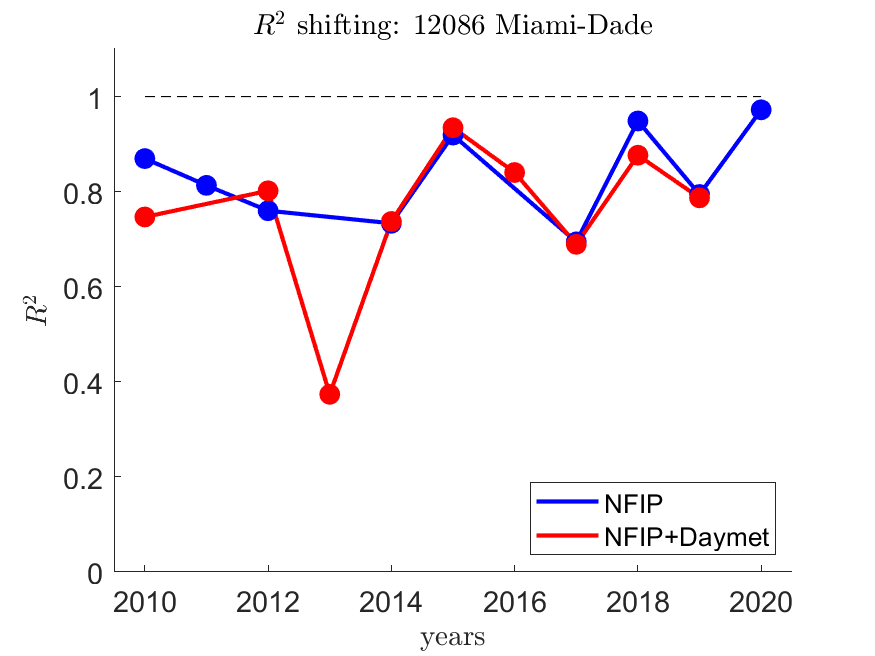} &
    \includegraphics[width=1.5in]{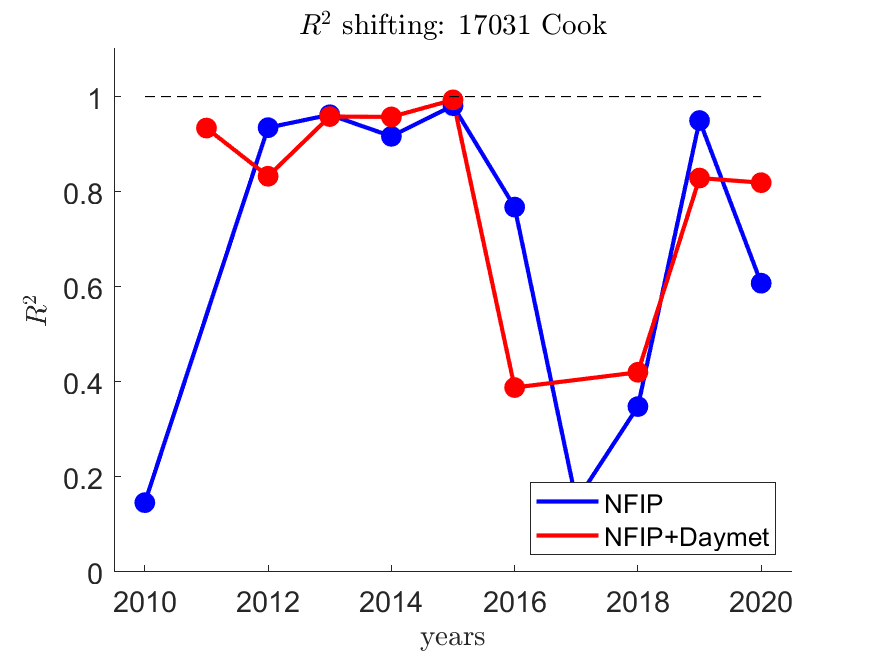}&
    \includegraphics[width=1.5in]{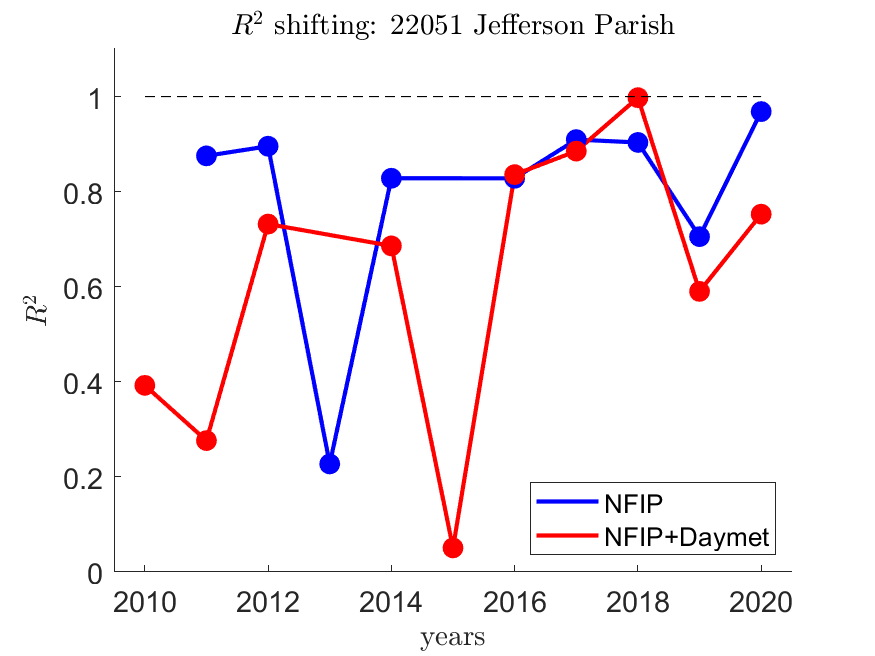} \\
(a.1) 12011& (b.1) 12086& (c.1) 17031 & (d.1) 22051\\
    \includegraphics[width=1.5in]{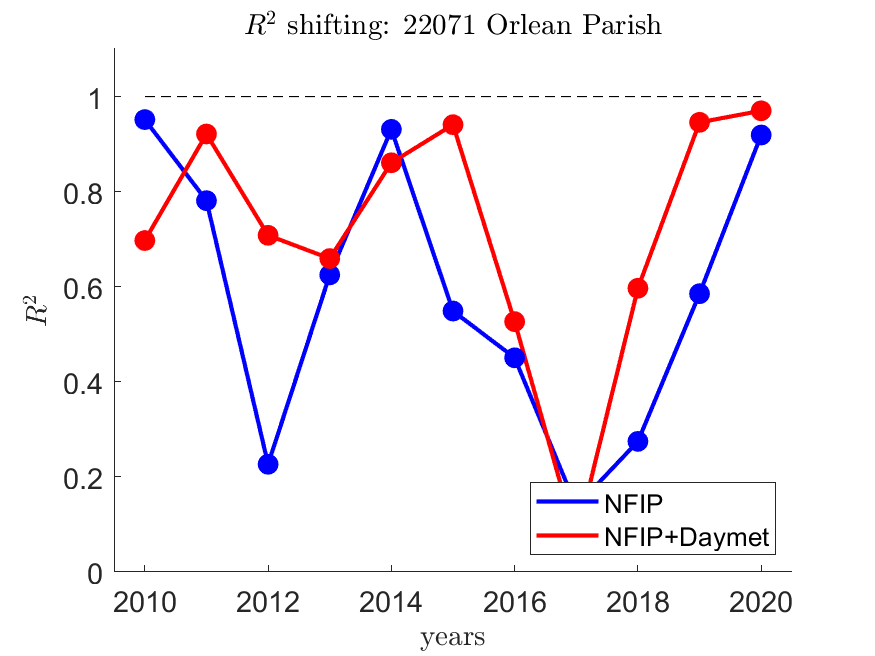} &
    \includegraphics[width=1.5in]{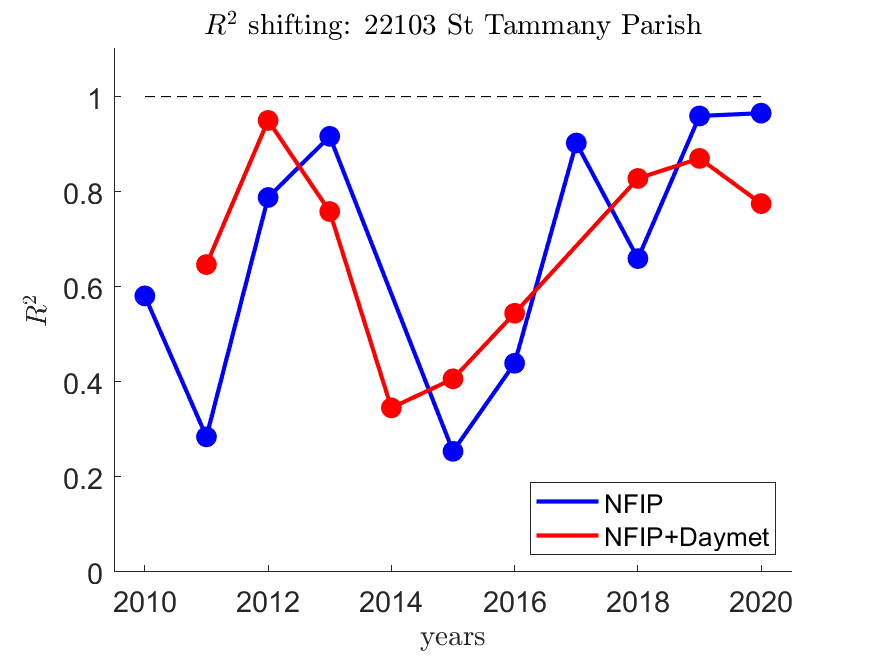} &
    \includegraphics[width=1.5in]{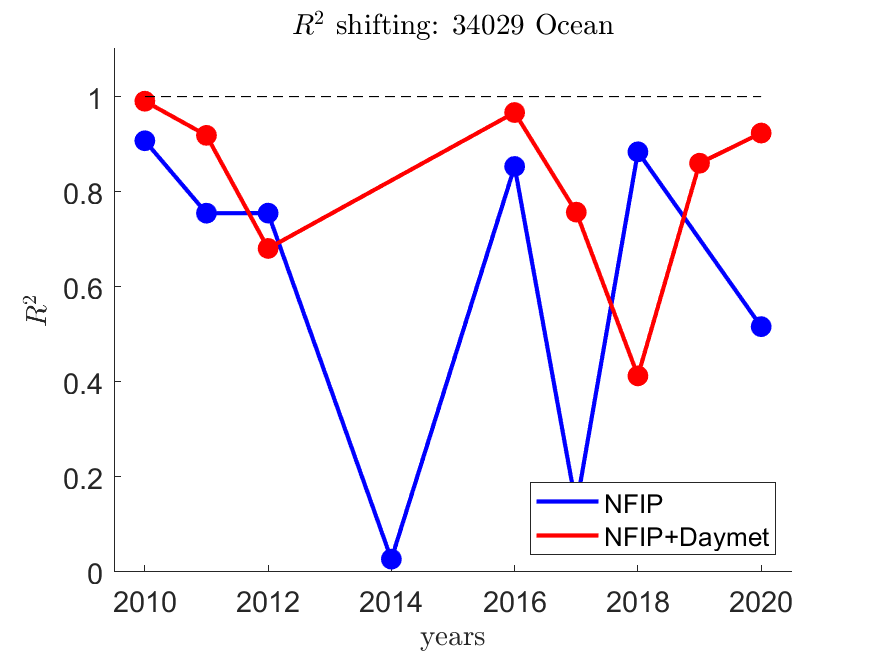} &
    \includegraphics[width=1.5in]{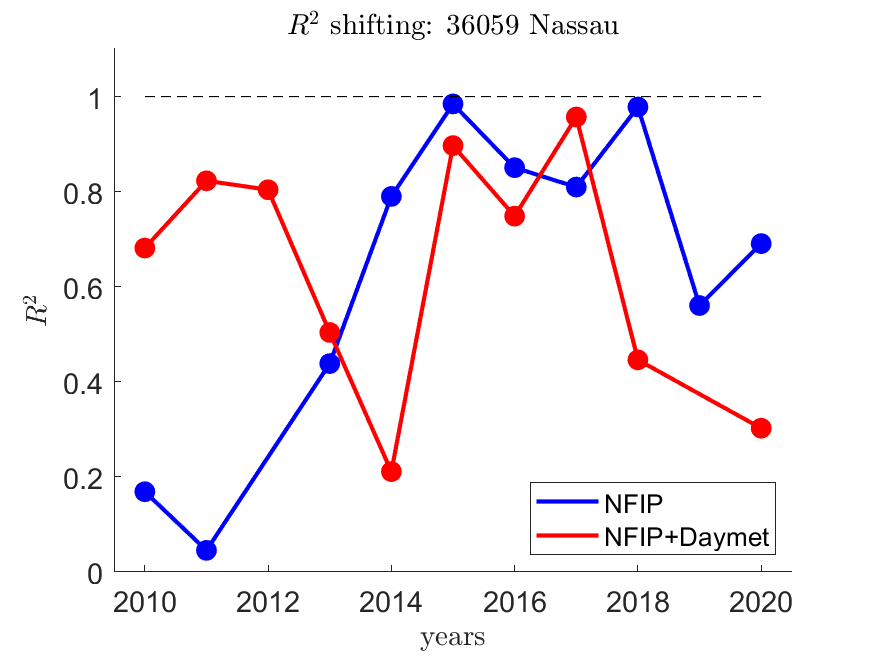} \\
(e.1) 22071& (f.1) 22103& (g.1) 34029& (h.1) 36059\\
    \includegraphics[width=1.5in]{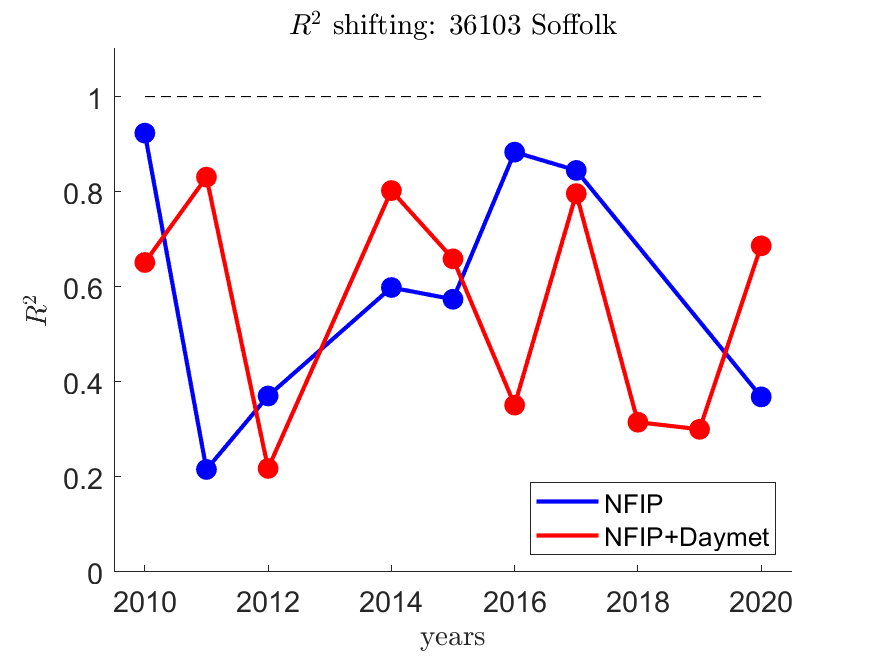} &
    \includegraphics[width=1.5in]{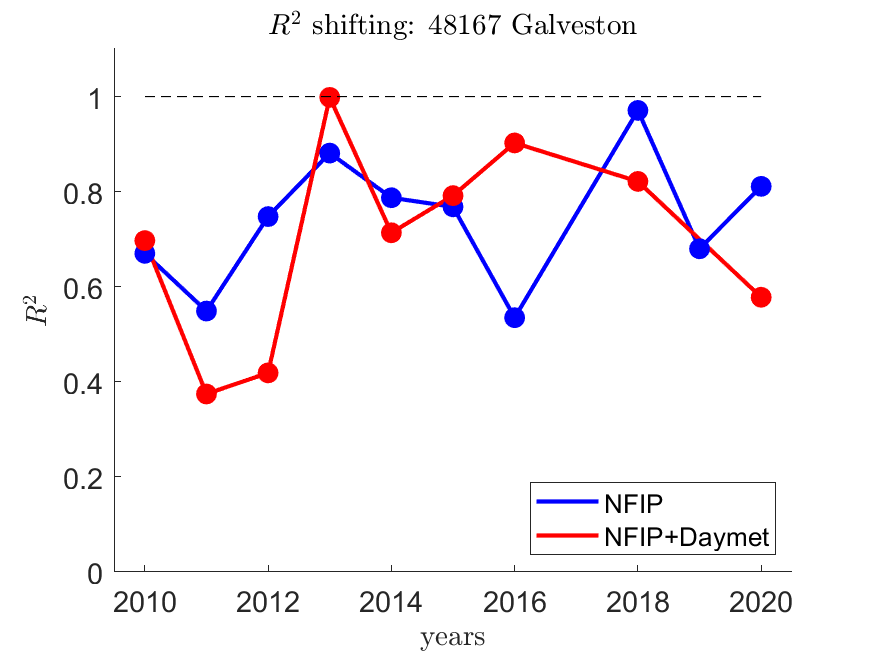} &
    \includegraphics[width=1.5in]{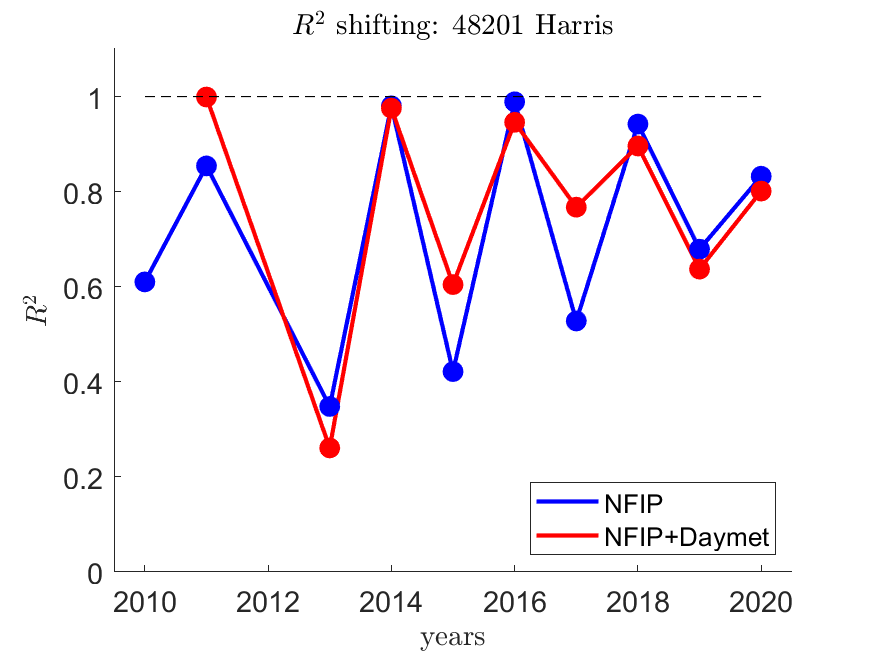} \\
(i.1) 36103& (j.1) 48167& (k.1) 48201\\
\end{tabular}
 \begin{tabular}{cccc}
    \includegraphics[width=1.5in]{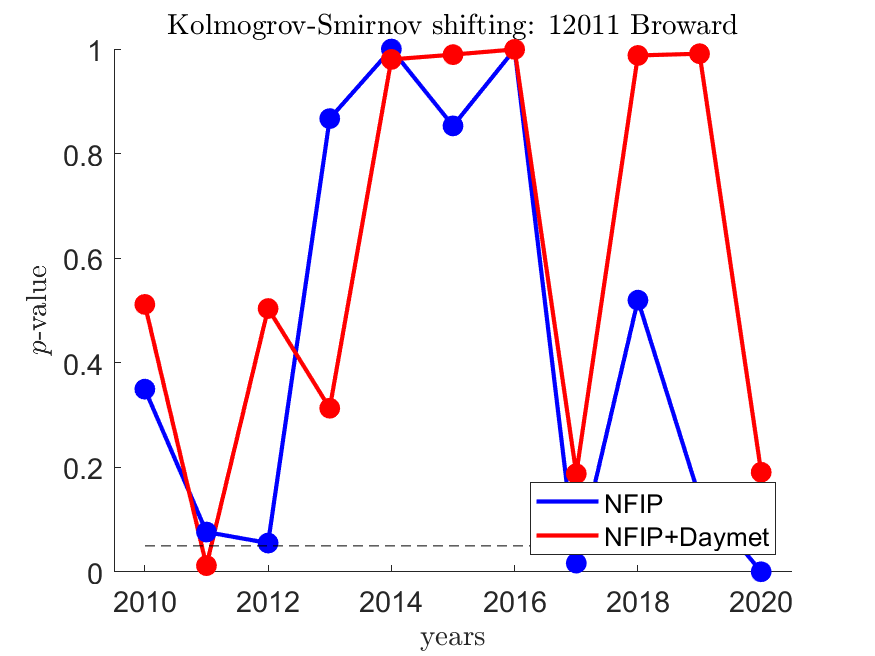} &
    \includegraphics[width=1.5in]{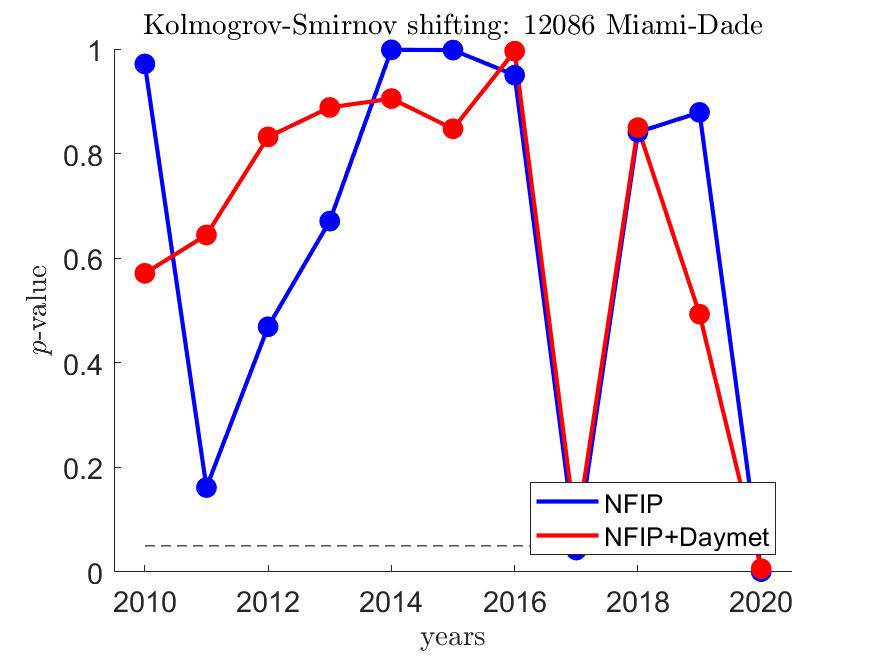} &
    \includegraphics[width=1.5in]{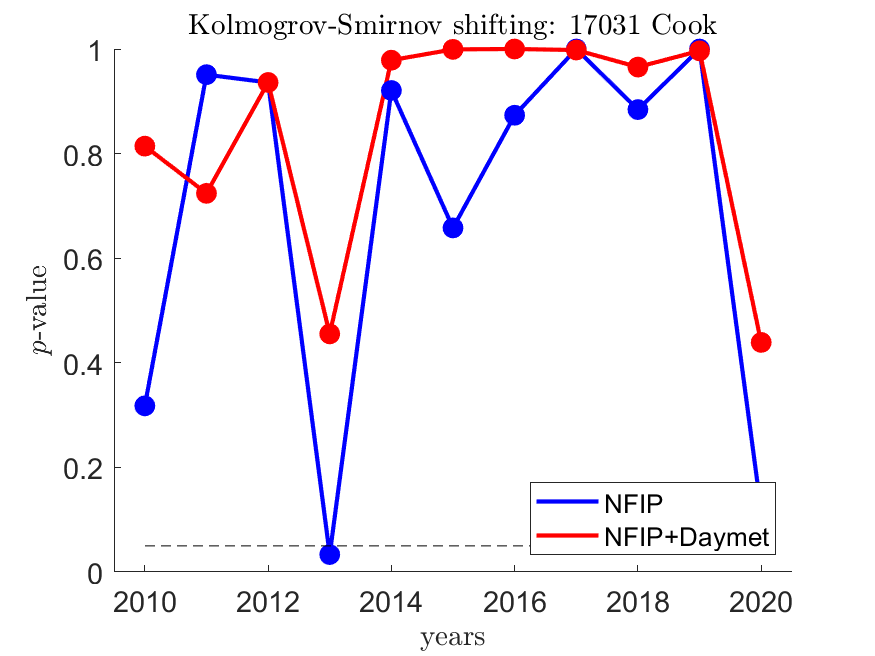} &
    \includegraphics[width=1.5in]{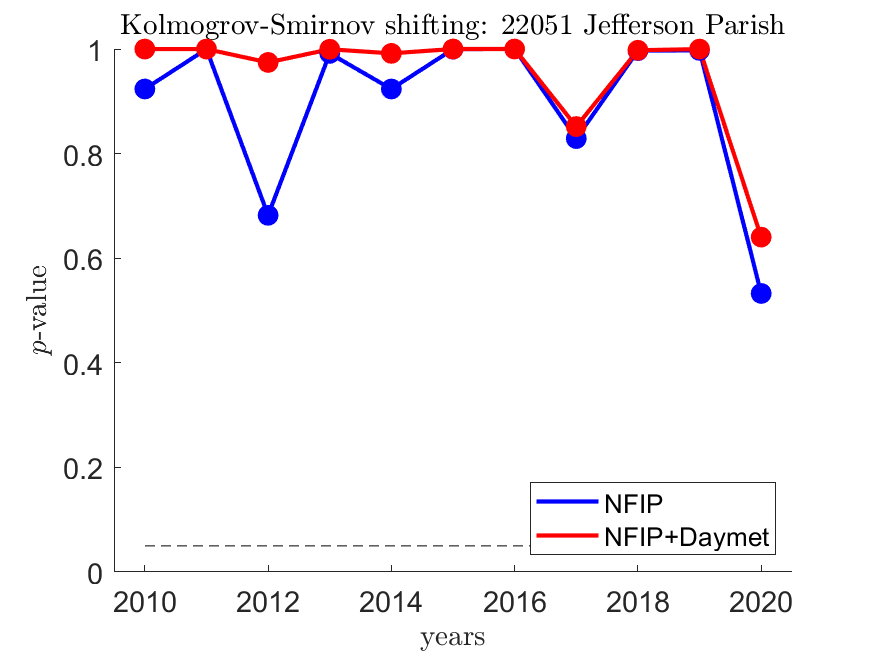} \\
(a.2) 12011& (b.2) 12086& (c.2) 17031 & (d.2) 22051\\
    \includegraphics[width=1.5in]{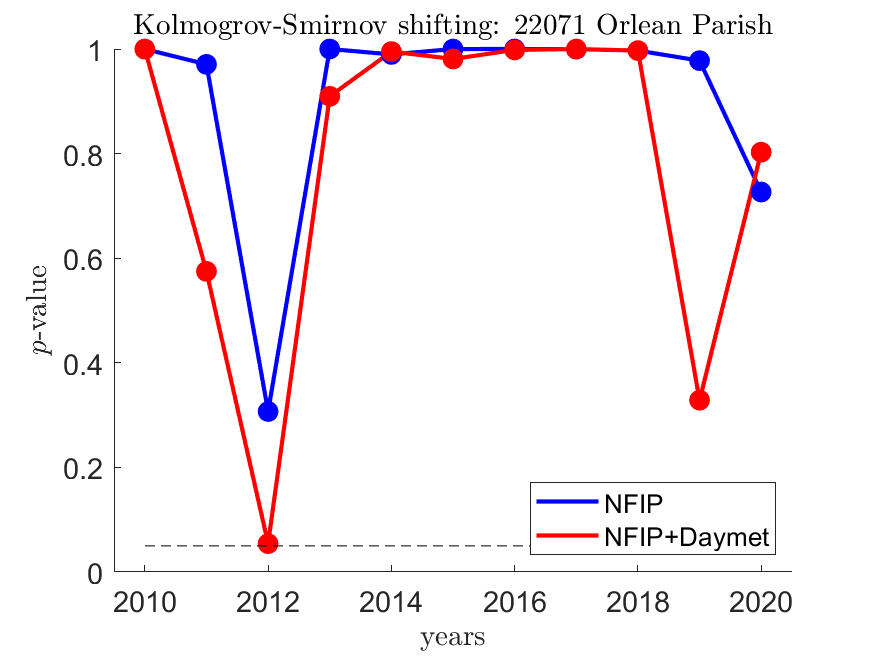} &
    \includegraphics[width=1.5in]{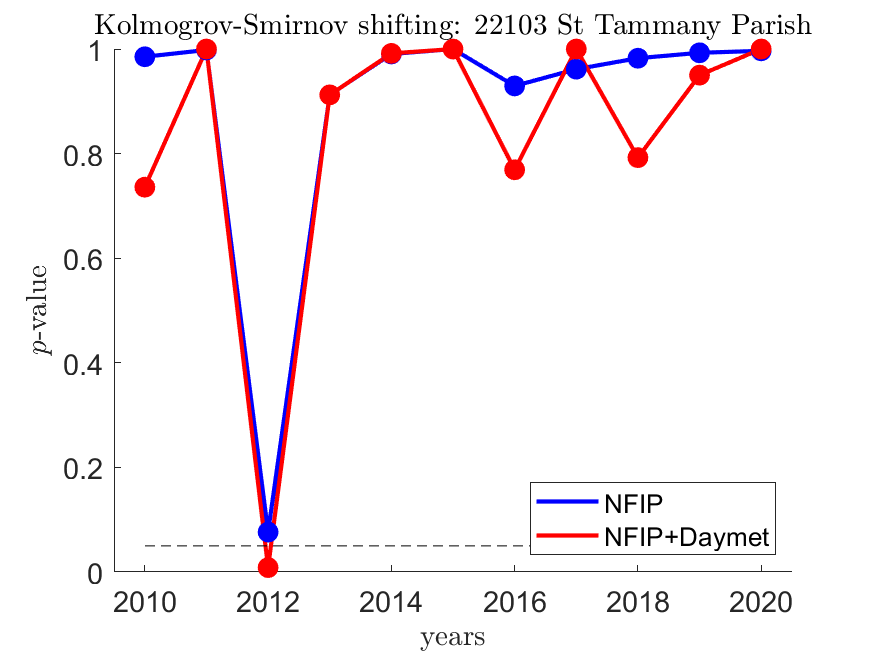} &
    \includegraphics[width=1.5in]{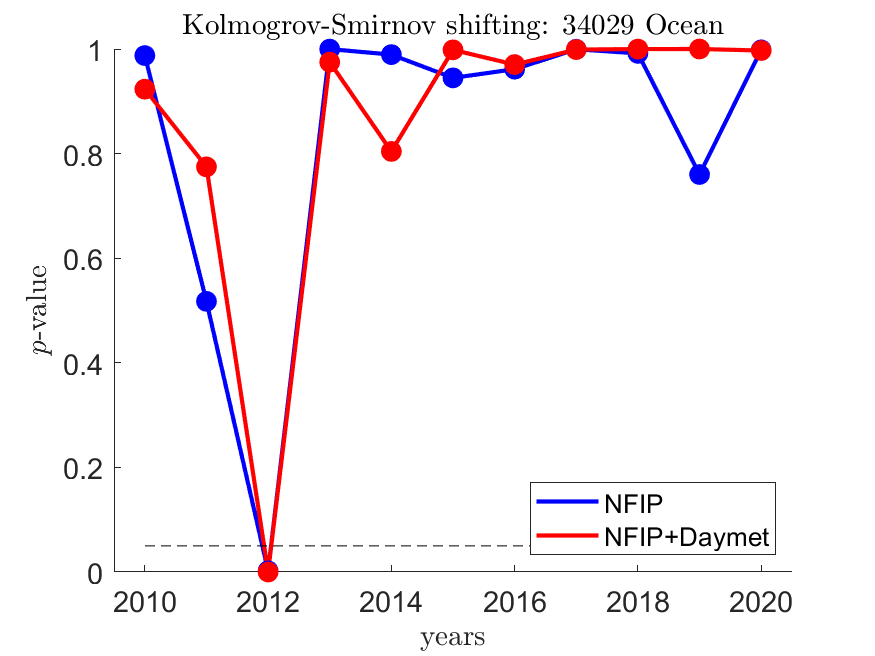} &
    \includegraphics[width=1.5in]{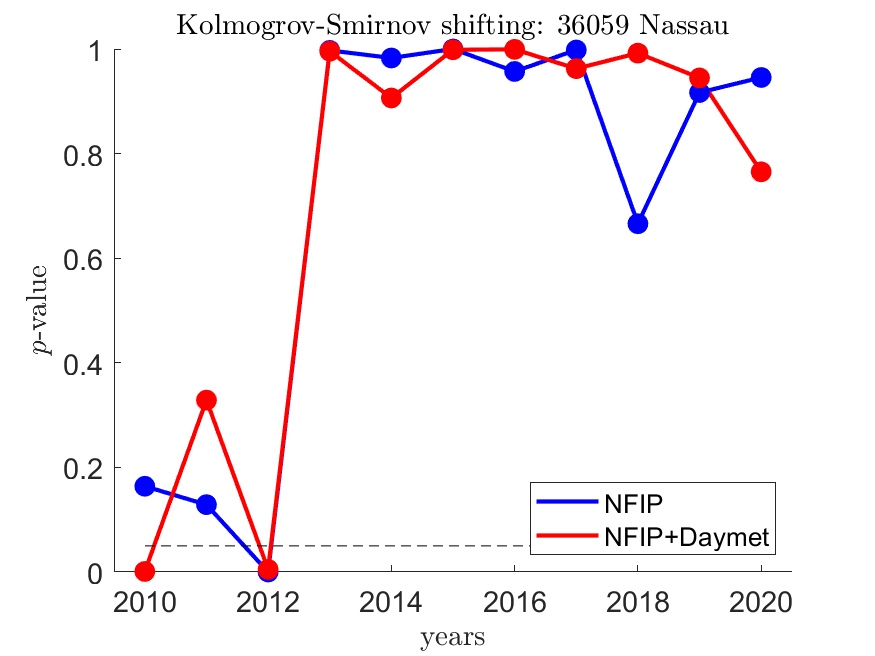} \\
(e.2) 22071& (f.2) 22103& (g.2) 34029& (h.2) 36059\\
    \includegraphics[width=1.5in]{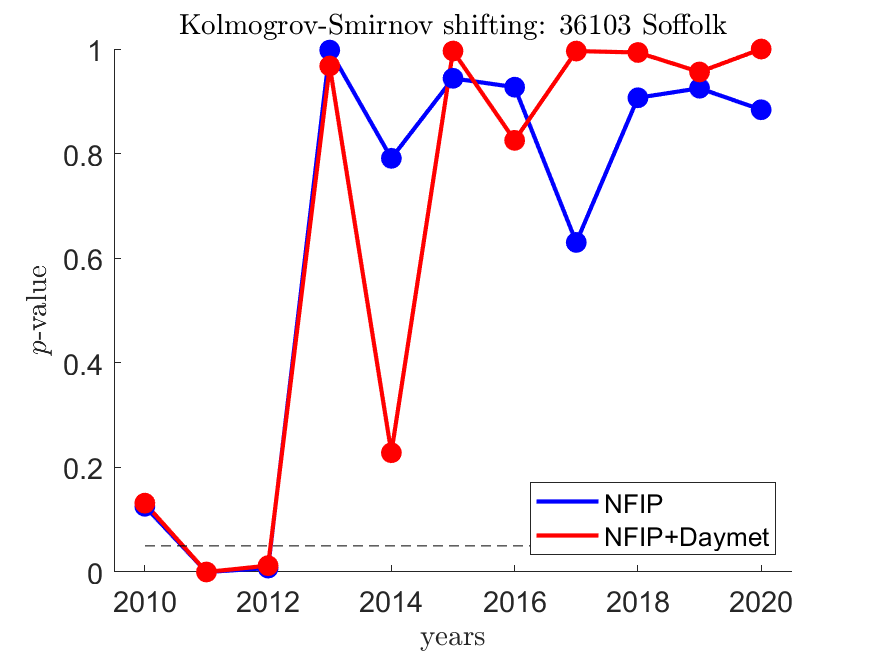} &
    \includegraphics[width=1.5in]{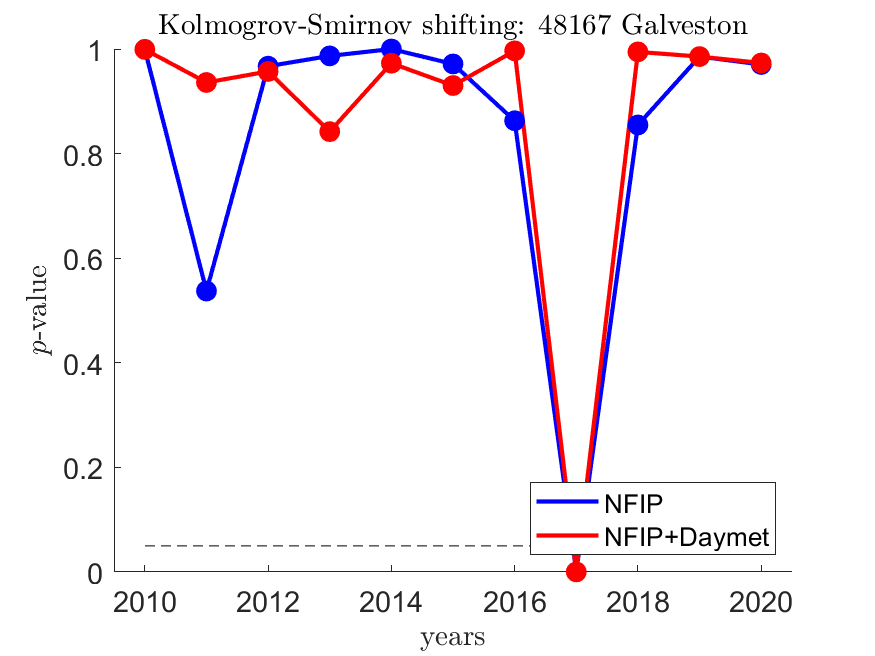} &
    \includegraphics[width=1.5in]{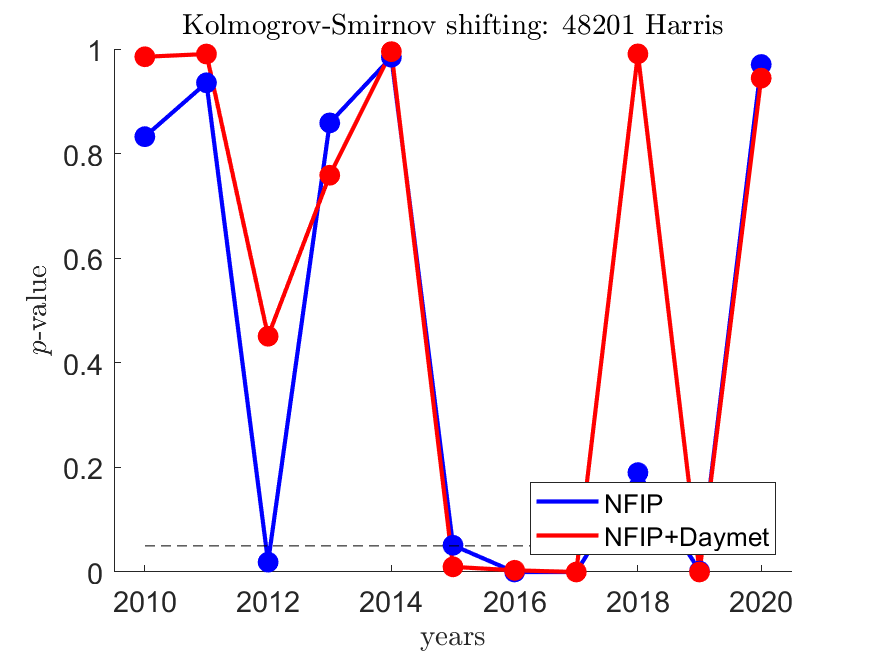} \\
(i.2) 36103& (j.2) 48167& (k.2) 48201\\

    \end{tabular}
    \caption{Shifting Window. Resulting in $R^2$ (*.1) and corresponding $p$-value (*.2) for the counties under analysis for the testing period between 2010 and 2020. For the $p$-value figures, the horizontal dotted line signals the reference value for $\alpha = 0.05$, rejecting values below the null hypothesis of suitable goodness of fit. The blue line shows the results for NFIP and the red line for the combination NFIP+Daymet.}
    \label{fig:shifting_window}
\end{figure}

\begin{figure}
    \centering
    \begin{tabular}{cccc}
    \includegraphics[width=1.5in]{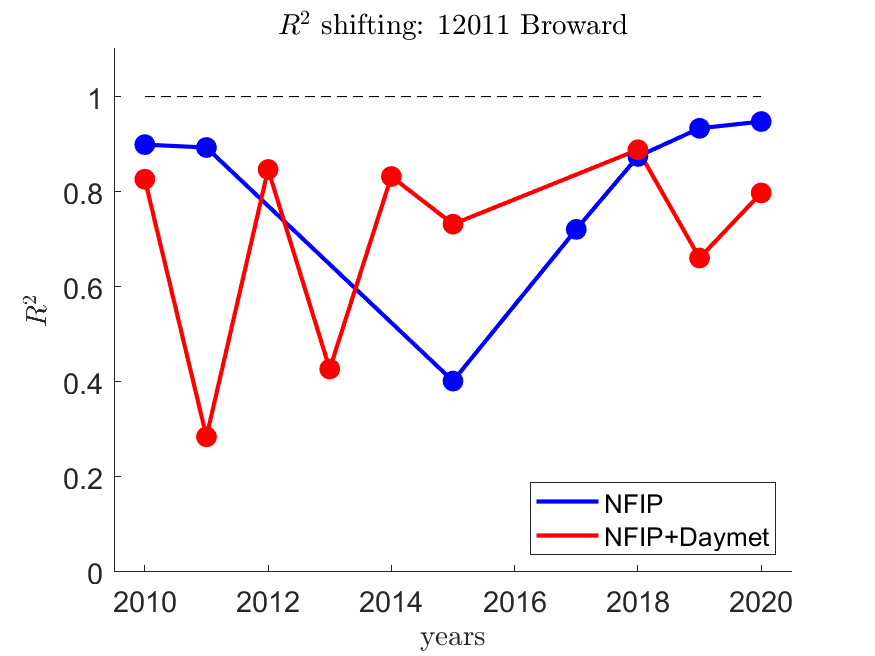} &
    \includegraphics[width=1.5in]{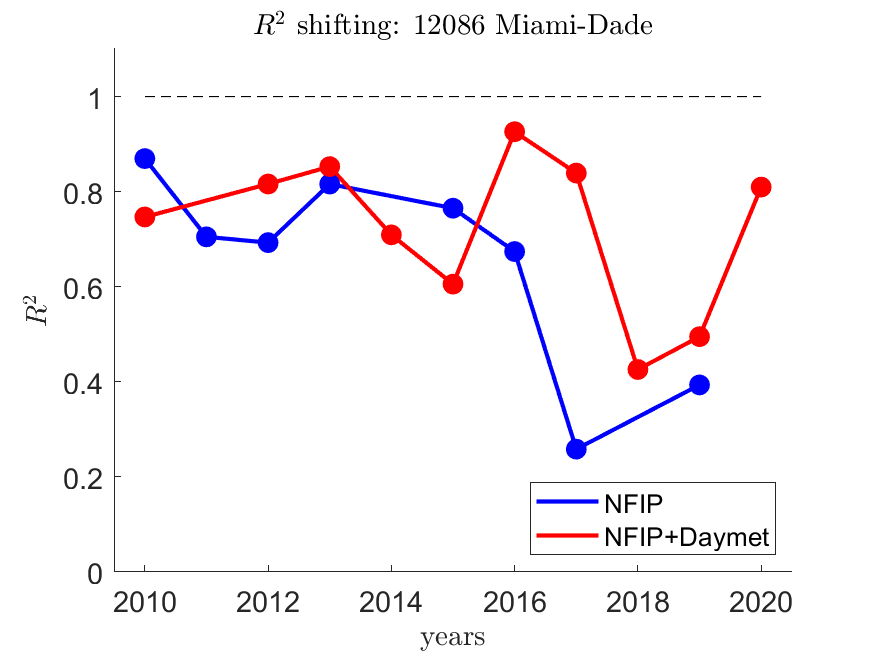} &
    \includegraphics[width=1.5in]{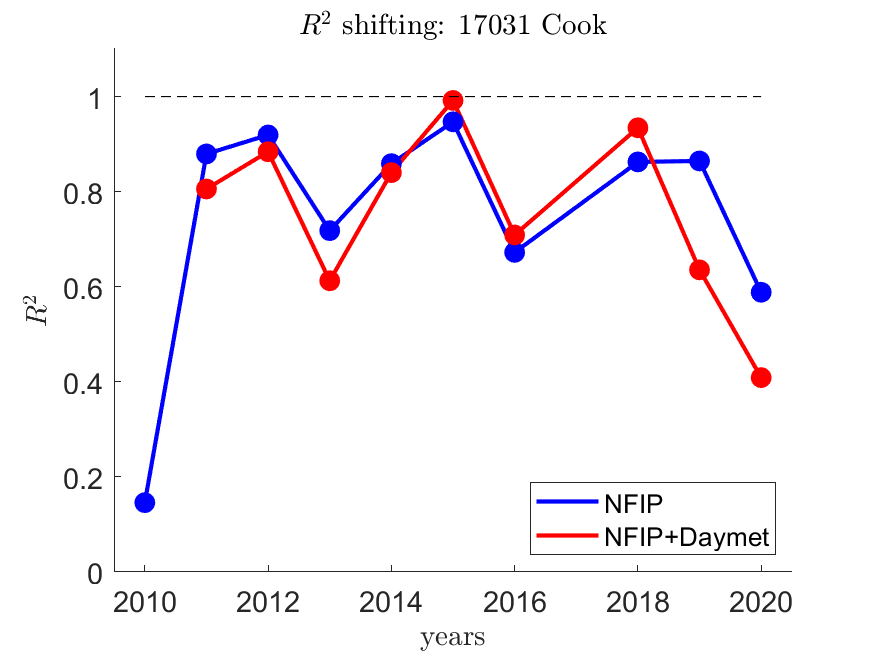} &
    \includegraphics[width=1.5in]{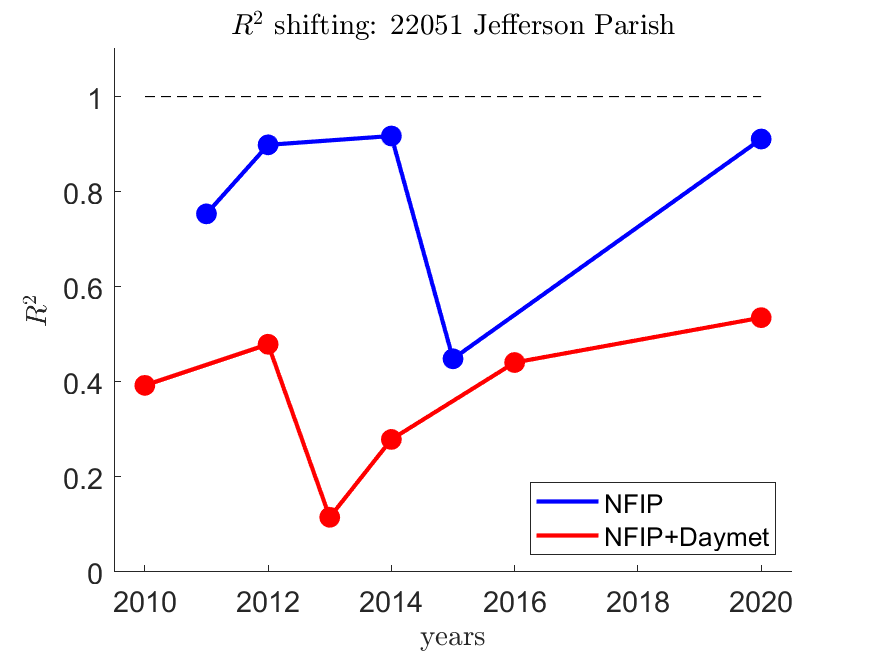} \\
(a.1) 12011& (b.1) 12086& (c.1) 17031 & (d.1) 22051\\
    \includegraphics[width=1.5in]{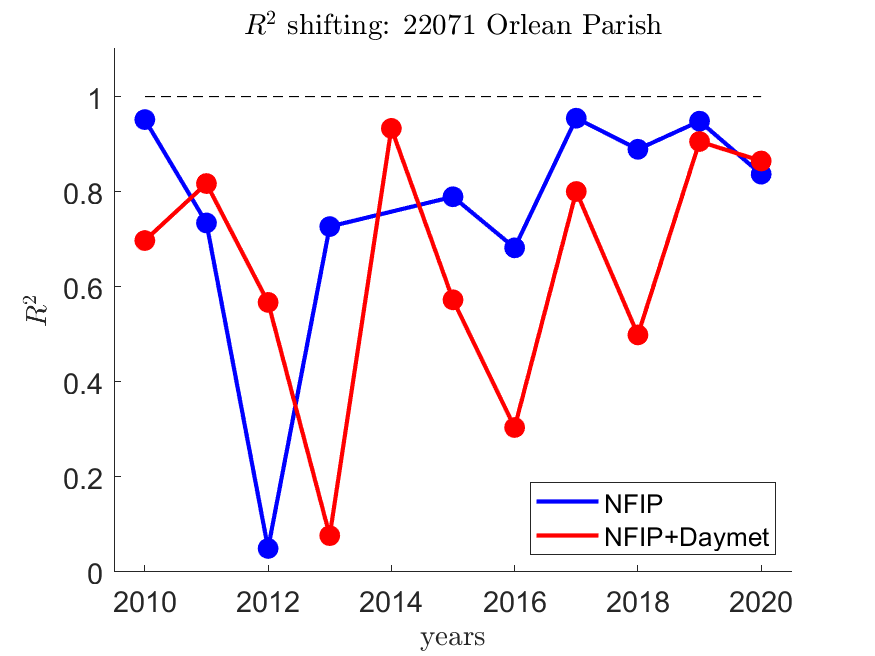} &
    \includegraphics[width=1.5in]{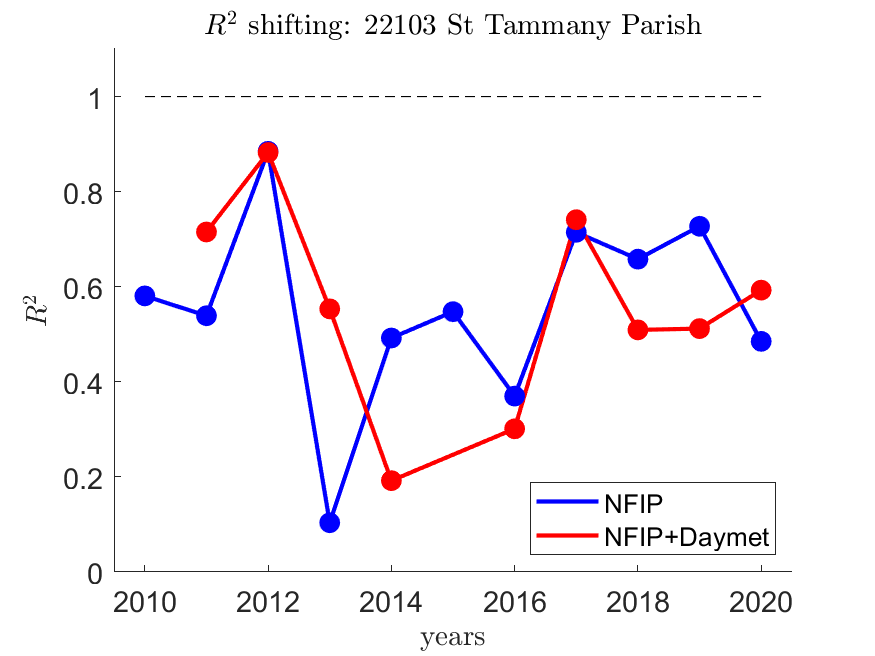} &
    \includegraphics[width=1.5in]{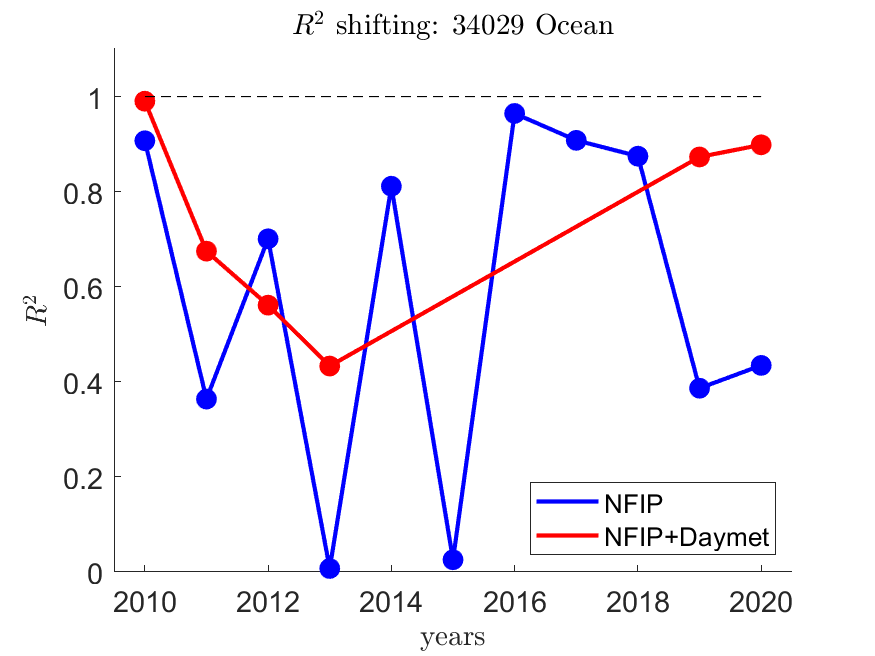} &
    \includegraphics[width=1.5in]{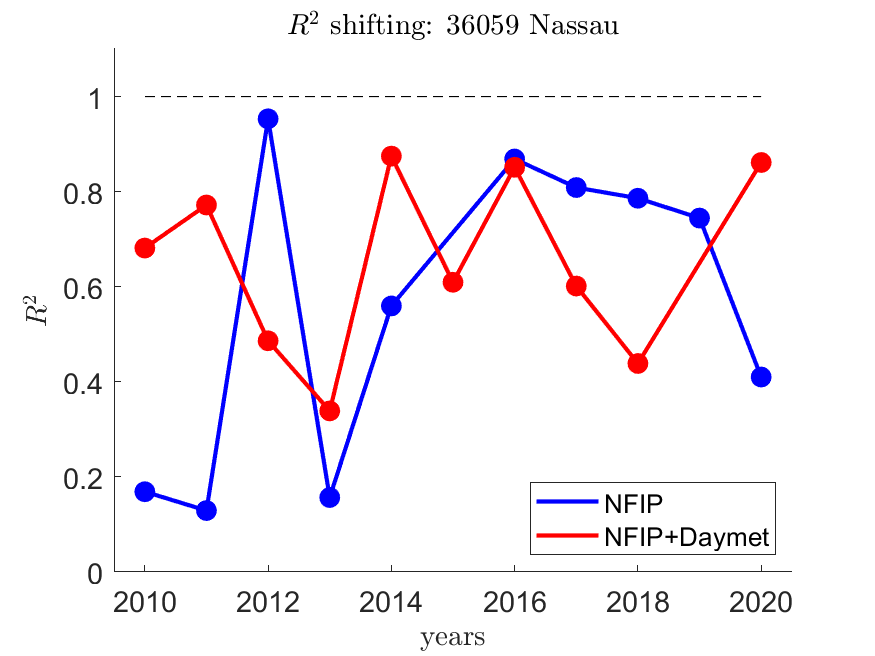} \\
(e.1) 22071& (f.1) 22103& (g.1) 34029& (h.1) 36059\\
    \includegraphics[width=1.5in]{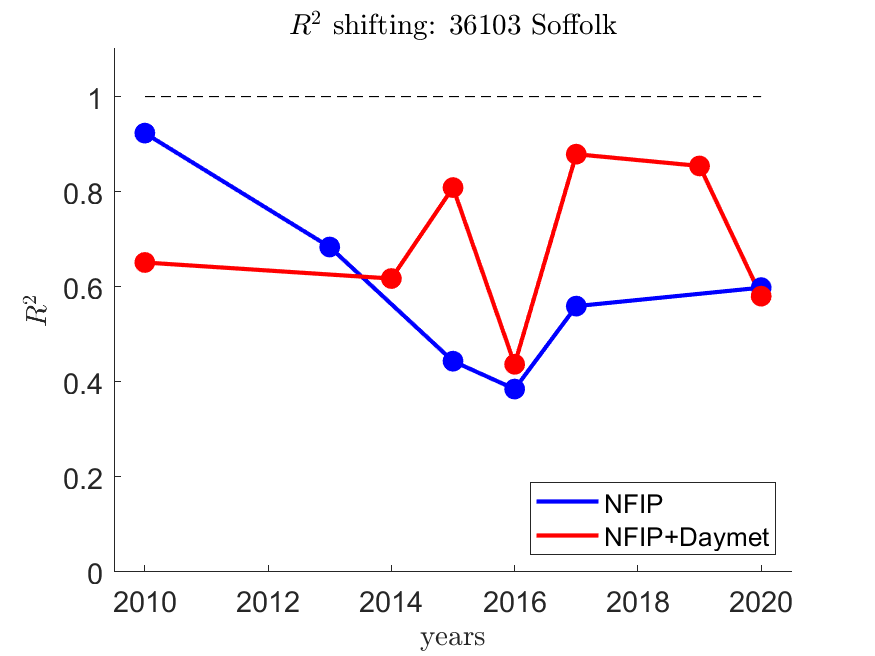} &
    \includegraphics[width=1.5in]{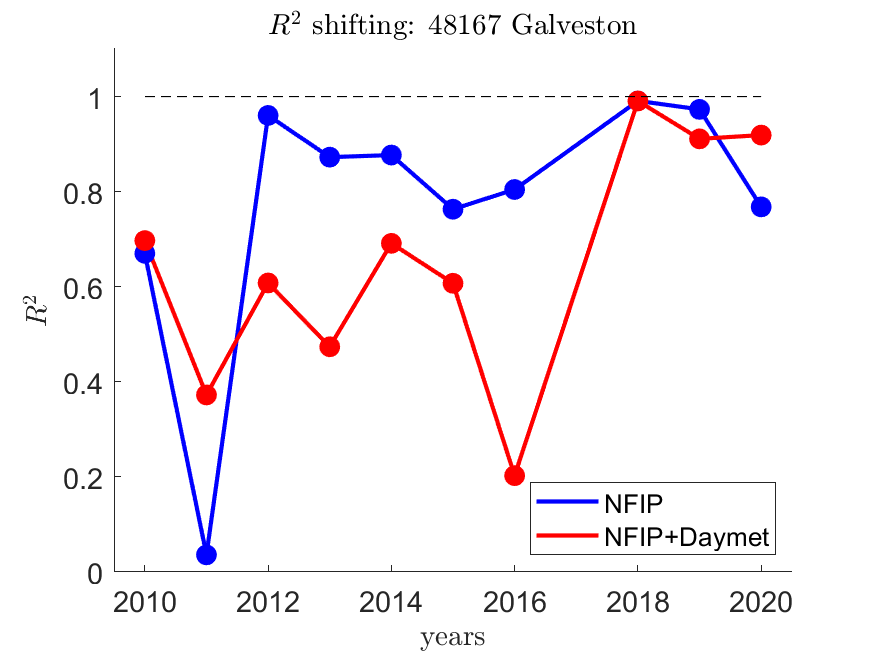} &
    \includegraphics[width=1.5in]{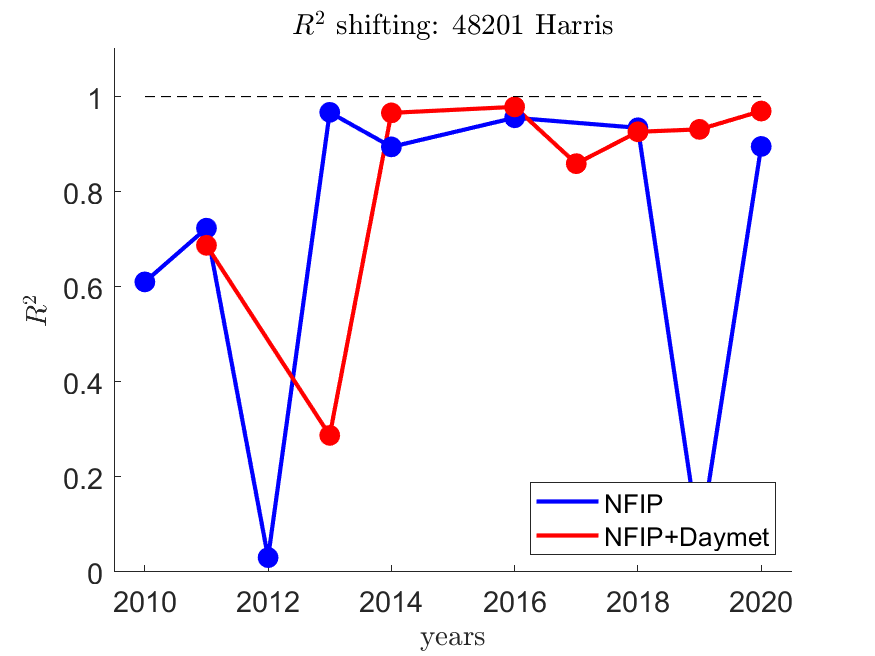} \\
(i.1) 36103& (j.1) 48167& (k.1) 48201\\
\end{tabular}
 \begin{tabular}{cccc}
    \includegraphics[width=1.5in]{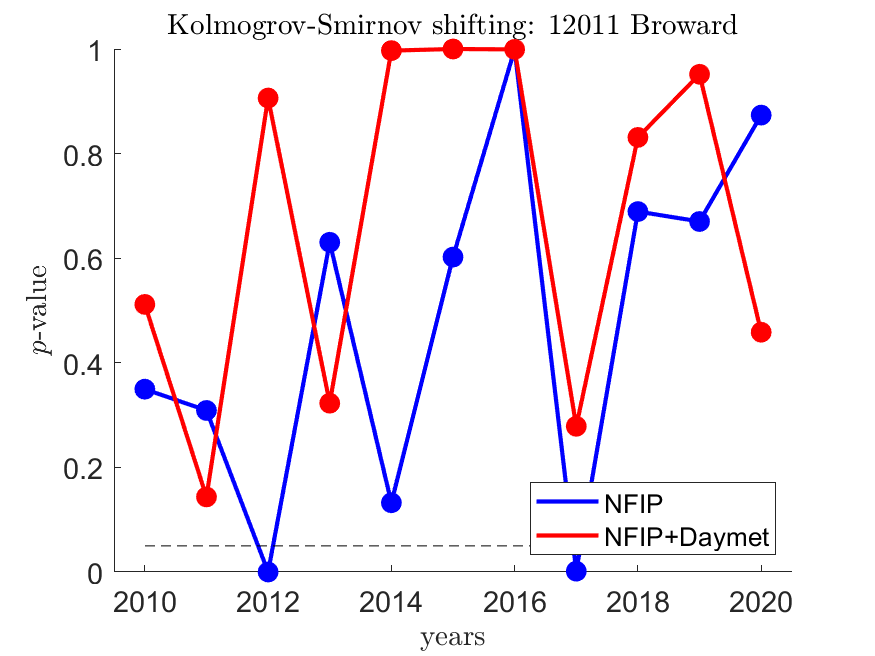} &
    \includegraphics[width=1.5in]{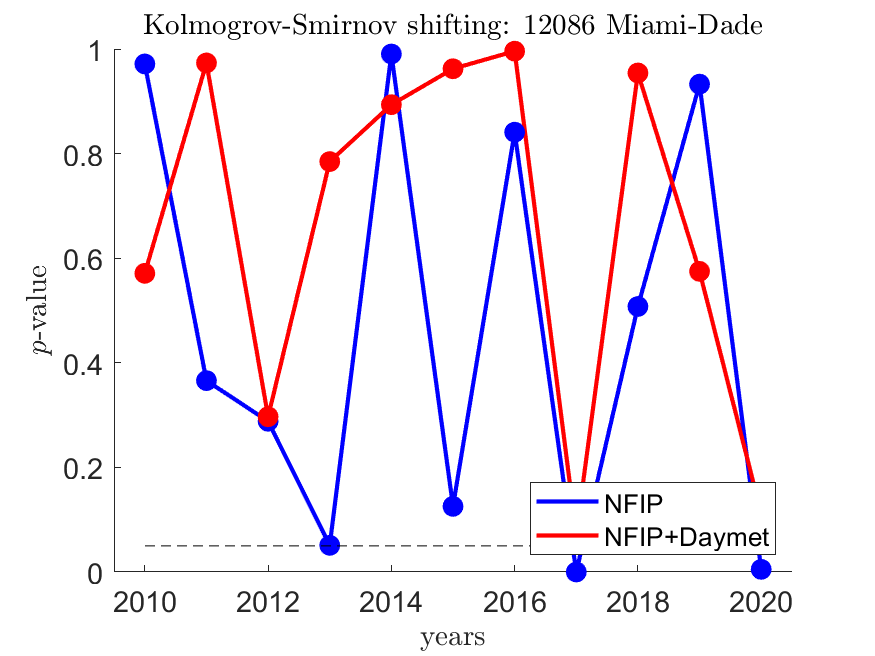} &
    \includegraphics[width=1.5in]{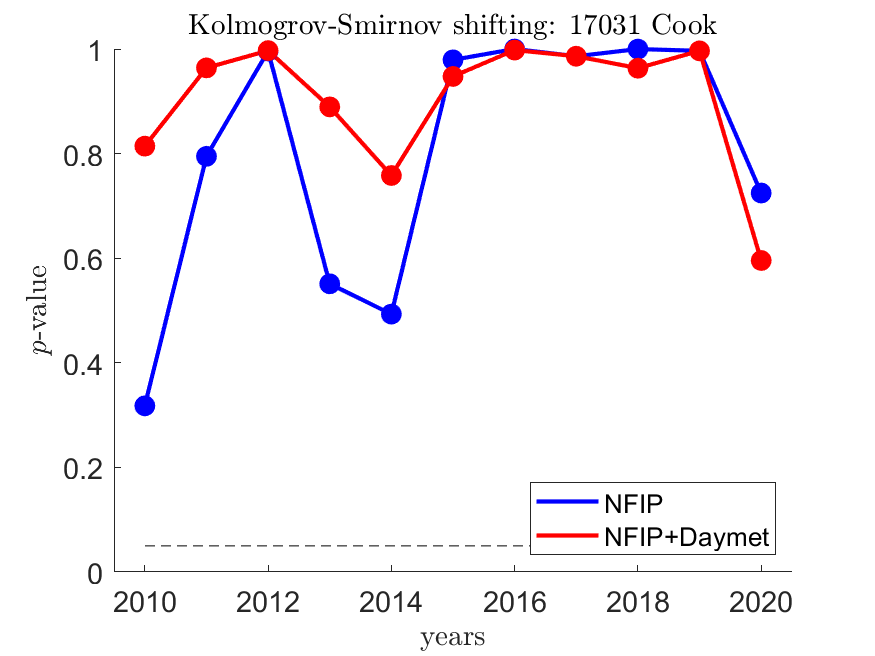} &
    \includegraphics[width=1.5in]{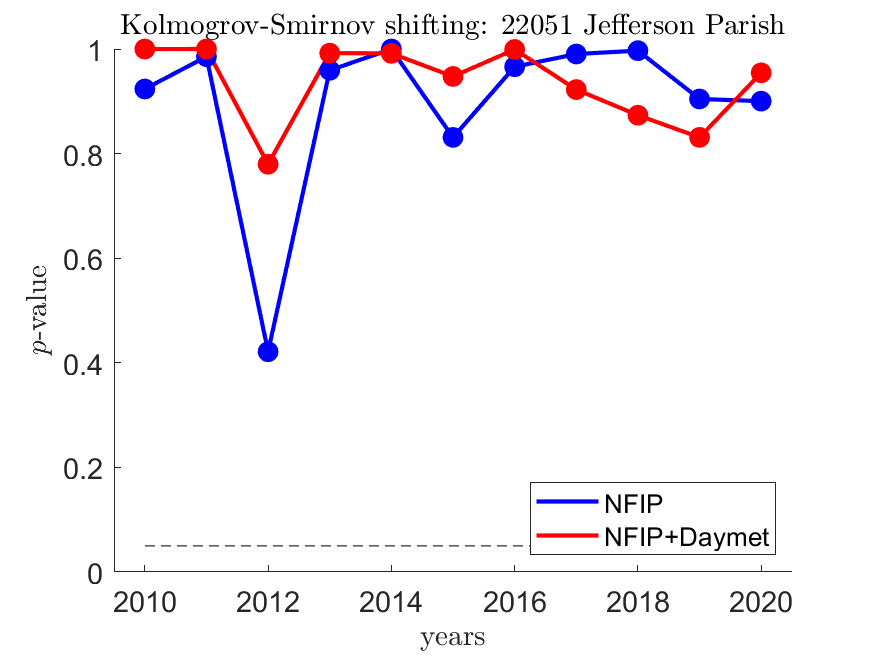} \\
(a.2) 12011& (b.2) 12086& (c.2) 17031 & (d.2) 22051\\
    \includegraphics[width=1.5in]{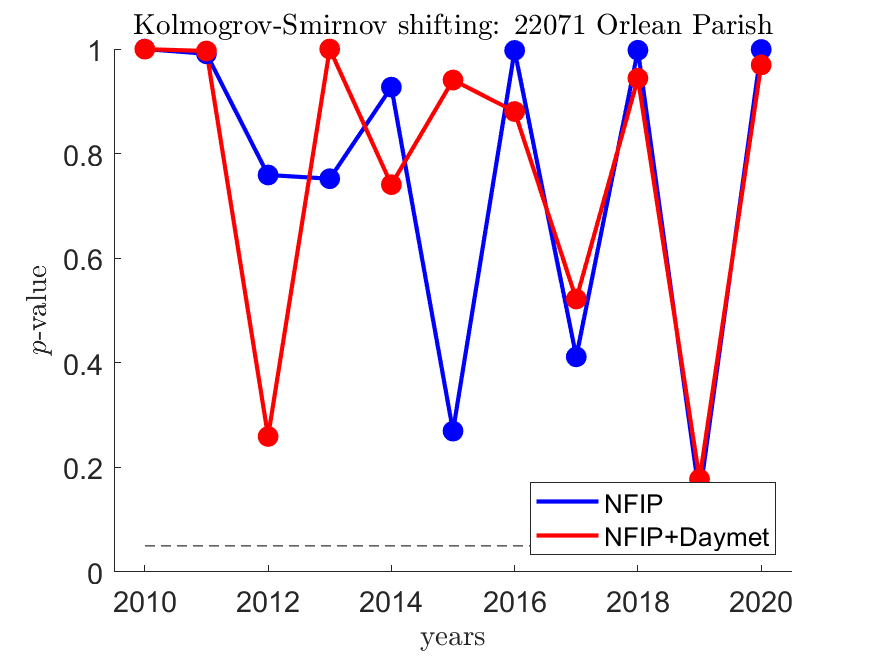} &
    \includegraphics[width=1.5in]{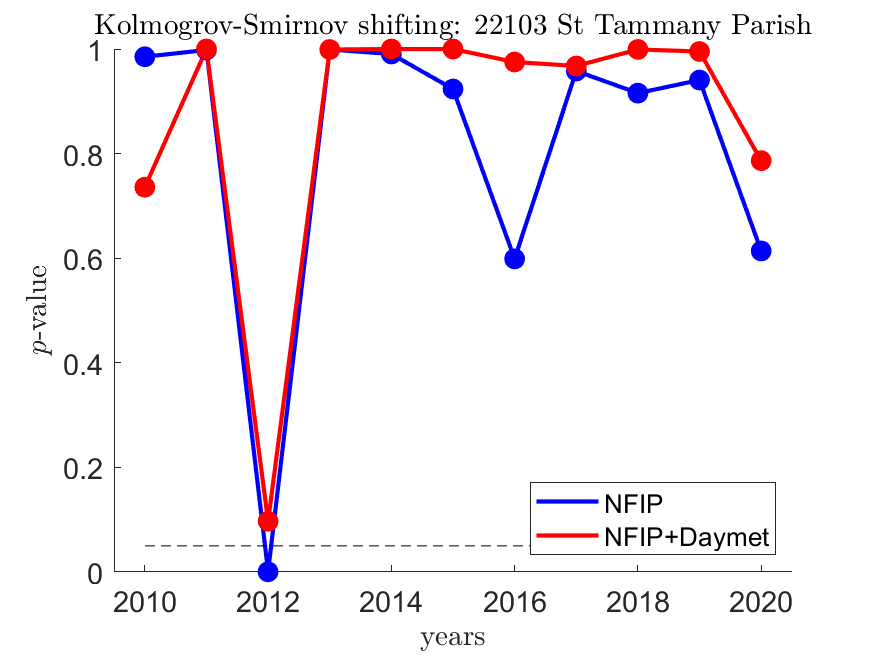} &
    \includegraphics[width=1.5in]{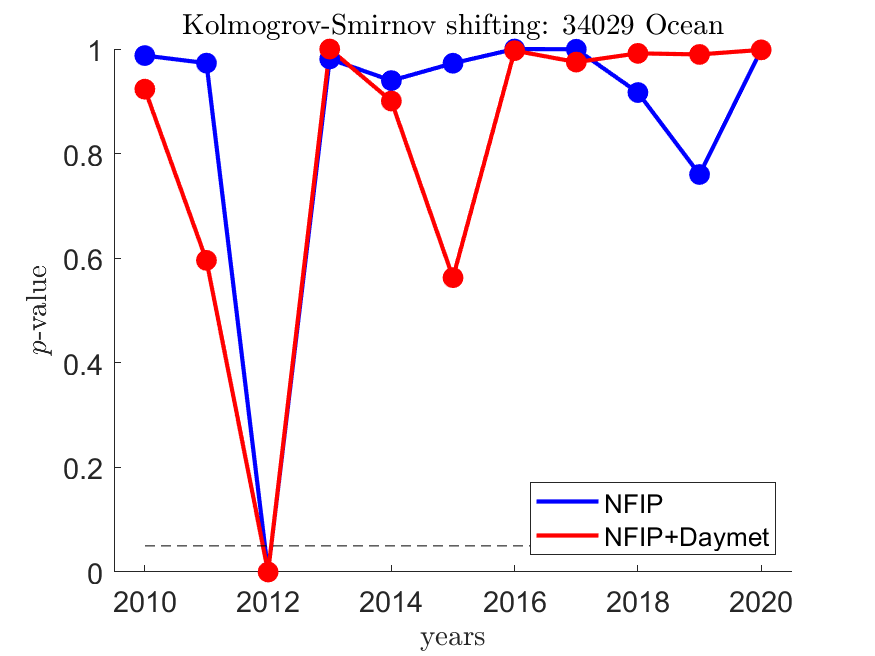} &
    \includegraphics[width=1.5in]{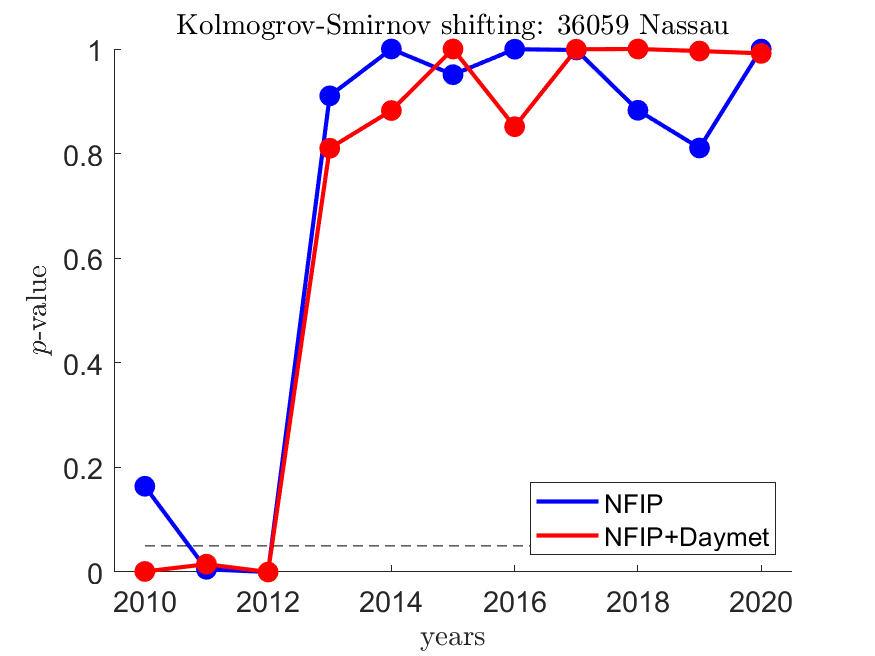} \\
(e.2) 22071& (f.2) 22103& (g.2) 34029& (h.2) 36059\\
    \includegraphics[width=1.5in]{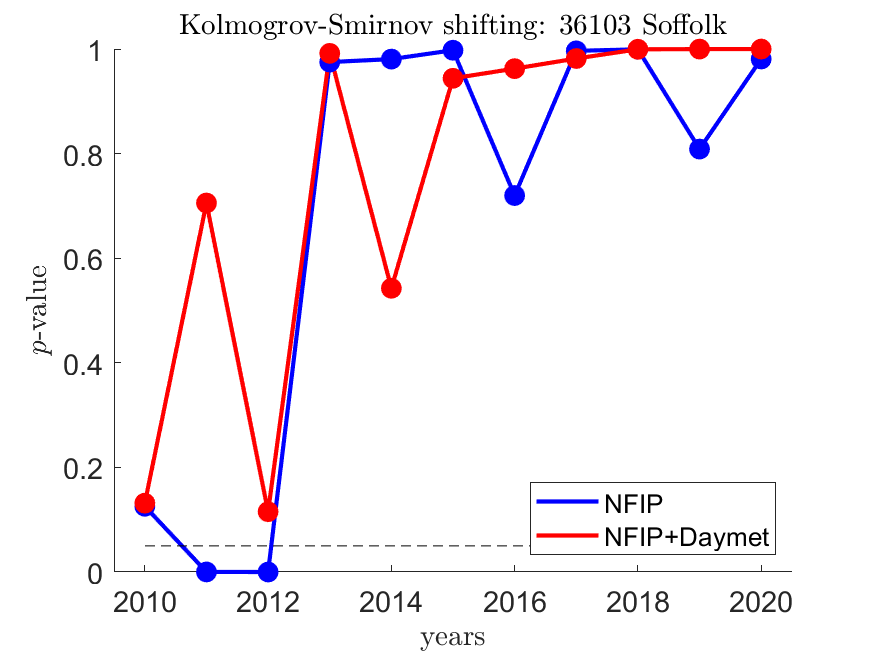} &
    \includegraphics[width=1.5in]{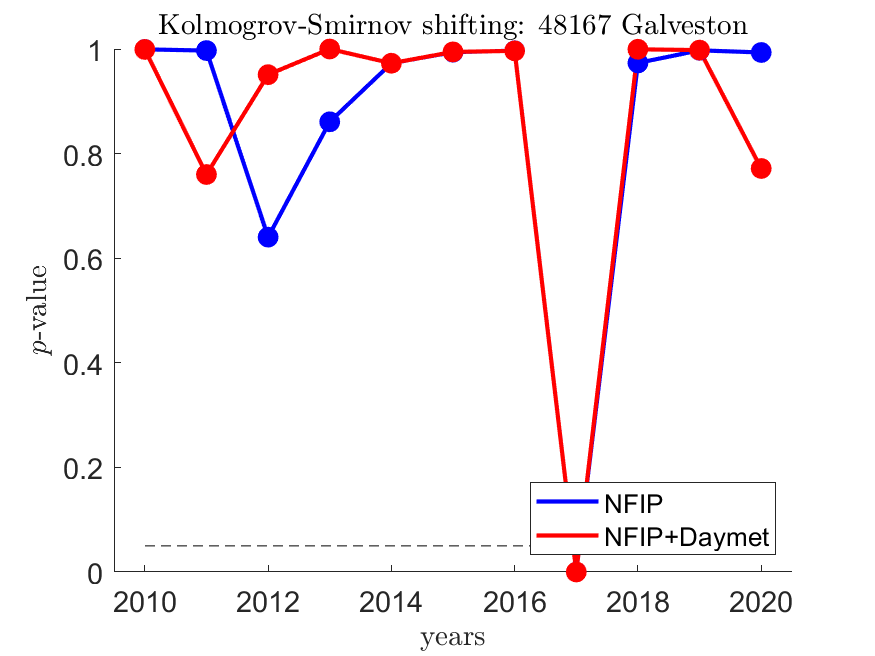} &
    \includegraphics[width=1.5in]{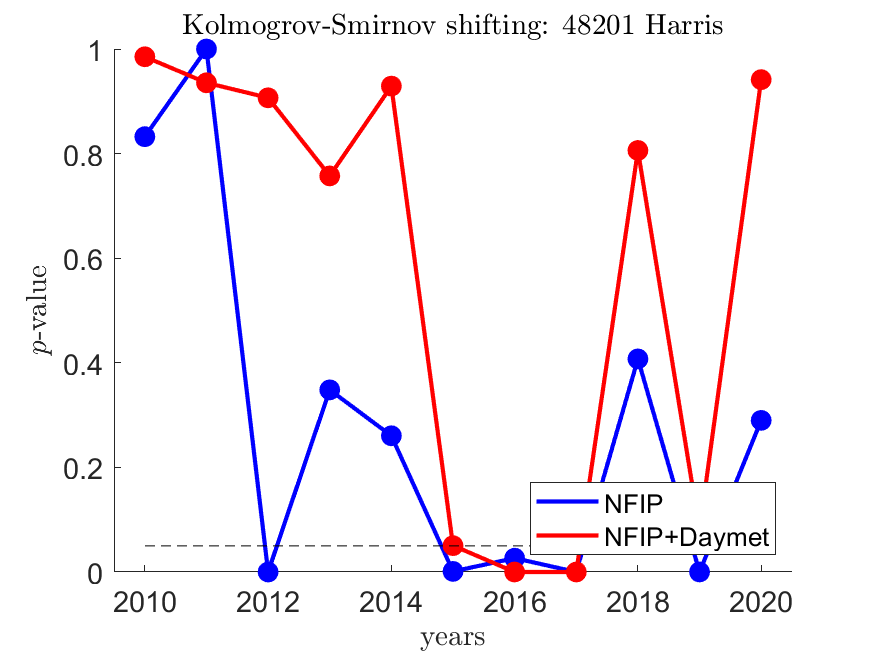} \\
(i.2) 36103& (j.2) 48167& (k.2) 48201\\

    \end{tabular}
    \caption{Expanding Window. Resulting in $R^2$ (*.1) and corresponding $p$-value (*.2) for the counties under analysis for the testing period between 2010 and 2020. For the $p$-value figures, the horizontal dotted line signals the reference value for $\alpha = 0.05$, rejecting values below the null hypothesis of suitable goodness of fit. The blue line shows the results for NFIP, and the red line for the combination NFIP+Daymet.}
    \label{fig:expanding_window}
\end{figure}

Table~\ref{tb:NFIPDataset} describes the structure of the NFIP dataset, while Figures~\ref{fig:shifting_window} and ~\ref{fig:expanding_window} provide further detail for the results of fitting regressors to each county.

\begin{table}[htb!]

    \centering
    \caption{NFIP dataset. Here, we describe the 22 discrete and nine continuous variables employed in this research and the range of values each takes in the dataset.
    }
     \label{tb:NFIPDataset}

    \begin{footnotesize}
    \begin{tabular}{lp{1.5in}p{1.75in}lp{1in}p{1.5in}}
    & \multicolumn{2}{c}{{\bf Discrete}} &  &   \multicolumn{2}{c}{{\bf Continuous}} \\
1 &        Agriculture Structure Indicator  &  Y, N &

23 &
Amount Paid On Building Claim & ${\mathbb{R}}\geq 0$
\\

2&        Basement Enclosure Crawlspace Type & $0, \dots, 4$ &
24 &
Base Flood Elevation & elevation feet for a 1\% chance/year of flooding
\\

3 & Community Rating System Discount & 1,\dots, 10
&
25 & Elevation Difference & ${\mathbb{R}}$ \\

4 & Condominium Indicator &
N, U, A, H, L, T;  &
26 &Latitude  & $-90^\circ \leq {\mathbb{R}} \leq 90^\circ $
\\

5 & Date Of Loss & 1970-08-31, \dots, 2021-09-08 &
27 & Longitude  & $-180^\circ \leq {\mathbb{R}} \leq 180^\circ $
\\

6 &Elevated Building Indicator & Y, N &
28 &Lowest Adjacent Grade &  ${\mathbb{R}}$\\

7 &Elevation Certificate Indicator & Y, N &
 \\


8 &Flood Zone & A, A01,$\dots$, A30, A99,
A; AE, A1 $\dots$ A30; A99  AH, AHB, AO,AOB,X, B,X, C ; D; V; VE, V1-V30; AE, VE, X, V1-V30,  B,C; AR; AHB, AOB, ARE, ARH, ARO, and ARA
&
29 & Lowest Floor Elevation & ${\mathbb{R}}$\\

9 &House of Worship & Y, N &
30 & Policy Count &  $1,\dots, {\cal{Z}}$\\

10 &Location Of Contents & 1, $\dots$, 7 &
31 & Total Building Insurance Coverage & ${\mathbb{R}}\geq 0$\\

11 &Lowest Adjacent Grade & $1,\dots, 6$ &
32 & Total Contents Insurance Coverage & ${\mathbb{R}}\geq 0$\\

12 &Number of Floors in the Insured Building & Y, N\\

13 &Non Profit Indicator & Y, N\\

14 & Obstruction Type & $10, \dots, 98$\\

15 & Occupancy Type  & $1, \dots, 4$\\

16 &Original Construction Date & 1950.01.01 to 2048.07.25\\

17 &Post-FIRM Construction Indicator & Y, N\\

18 &Rate Method & 1,\dots, 9, A, B, E, F, G, P,$\dots$, T, W\\

19 &Primary Residence& Y, N
\\

20 & Small Business Indicator Building & Y, N
\\

21 &Year of Loss & 1970, $\dots$, 2021
 \\

22 & Zip Code & 5-digit Postal Zip Code \\

    \end{tabular}
   \end{footnotesize}

\end{table}


\newpage

\end{singlespace}

\end{document}